\documentclass{article} % For LaTeX2e
\usepackage{macros}
\usepackage{xr}
\title{Algorithms for CVaR Optimization in MDPs}

\author{
Yinlam Chow\\
Institute of Computational \& Mathematical Engineering, Stanford University \\
Mohammad Ghavamzadeh\thanks{Mohammad Ghavamzadeh is at Adobe Research, on leave of absence from INRIA Lille - Team SequeL.}\\
INRIA Lille - Team SequeL \& Adobe Research}%\\
\newcommand{\BEAS}{\begin{eqnarray*}}
\newcommand{\EEAS}{\end{eqnarray*}}
\newcommand{\BEQ}{\begin{equation}}
\newcommand{\EEQ}{\end{equation}}
\newcommand{\BIT}{\begin{itemize}}
\newcommand{\EIT}{\end{itemize}}

\newcommand{\expec}{\mathbb{E}}

\usepackage{color}
\usepackage{amsmath}
\usepackage{amssymb}
\usepackage{graphicx}
\usepackage{comment,xspace}
\usepackage{fancybox}
\newcommand{\real}{{\mathbb{R}}}
\newcommand{\reals}{\real}

\newtheorem{proposition}[theorem]{Proposition}

\newtheorem{assumption}[theorem]{Assumption}

\newcommand{\argmin}{\operatornamewithlimits{argmin}}

\def\A{\mathcal{A}}
\def\R{R}

\def\P{P}

\def\V{V}
\def\Q{Q}
\def\X{\mathcal{X}}

\begin{document}

\maketitle

%%%%%%%%%%%%%%%%%%%%%%%%%%%%%%%%%%%%%%%%%%%%%%%%%%%%%%%%%%%%%%
%%%%%%%%%%%%%%%%%%%%%%%%%%%%%%%%%%%%%%%%%%%%%%%%%%%%%%%%%%%%%%
%%%%%%%%%%%%%%%%%%%%%%%%%%%%%%%%%%%%%%%%%%%%%%%%%%%%%%%%%%%%%%
%%%%%%%%%%%%%%%%%%%%%%%%%%%%%%%%%%%%%%%%%%%%%%%%%%%%%%%%%%%%%%
%%%%%%%%%%%%%%%%%%%%%%%%%%%%%%%%%%%%%%%%%%%%%%%%%%%%%%%%%%%%%%

\vspace{-0.325in}
\begin{abstract} 
In many sequential decision-making problems we may want to manage risk by minimizing some measure of variability in costs in addition to minimizing a standard criterion. Conditional value-at-risk (CVaR) is a relatively new risk measure that addresses some of the shortcomings of the well-known variance-related risk measures, and because of its computational efficiencies has gained popularity in finance and operations research. In this paper, we consider the mean-CVaR optimization problem in MDPs. We first derive a formula for computing the gradient of this risk-sensitive objective function. We then devise policy gradient and actor-critic algorithms that each uses a specific method to estimate this gradient and updates the policy parameters in the descent direction. We establish the convergence of our algorithms to locally risk-sensitive optimal policies. Finally, we demonstrate the usefulness of our algorithms in an optimal stopping problem.
\end{abstract} 

%%%%%%%%%%%%%%%%%%%%%%%%%%%%%%%%%%%%%%%%%%%%%%%%%%%%%%%%%%%%%%
%%%%%%%%%%%%%%%%%%%%%%%%%%%%%%%%%%%%%%%%%%%%%%%%%%%%%%%%%%%%%%
%%%%%%%%%%%%%%%%%%%%%%%%%%%%%%%%%%%%%%%%%%%%%%%%%%%%%%%%%%%%%%
%%%%%%%%%%%%%%%%%%%%%%%%%%%%%%%%%%%%%%%%%%%%%%%%%%%%%%%%%%%%%%
%%%%%%%%%%%%%%%%%%%%%%%%%%%%%%%%%%%%%%%%%%%%%%%%%%%%%%%%%%%%%%

\vspace{-0.2in}
\section{Introduction}
\label{sec:introduction}
\vspace{-0.075in}

A standard optimization criterion for an infinite horizon Markov decision process (MDP) is the {\em expected sum of (discounted) costs} (i.e.,~finding a policy that minimizes the value function of the initial state of the system). However in many applications, we may prefer to minimize some measure of {\em risk} in addition to this standard optimization criterion. In such cases, we would like to use a criterion that incorporates a penalty for the {\em variability} (due to the stochastic nature of the system) induced by a given policy. In {\em risk-sensitive} MDPs~\cite{Howard72RS}, the objective is to minimize a risk-sensitive criterion such as the expected exponential utility~\citep{Howard72RS}, a variance-related measure~\citep{Sobel82VD,filar1989variance}, or the percentile performance~\citep{Filar95PP}. The issue of how to construct such criteria in a manner that will be both conceptually meaningful and mathematically tractable is still an open question.

Although most losses (returns) are not normally distributed, the typical Markiowitz mean-variance optimization~\cite{Markowitz59PS}, that relies on the first two moments of the loss (return) distribution, has dominated the risk management for over $50$ years. Numerous alternatives to mean-variance optimization have emerged in the literature, but there is no clear leader amongst these alternative risk-sensitive objective functions. {\em Value-at-risk} (VaR) and {\em conditional value-at-risk} (CVaR) are two promising such alternatives that quantify the losses that might be encountered in the tail of the loss distribution, and thus, have received high status in risk management. For (continuous) loss distributions, while VaR measures risk as the maximum loss that might be incurred w.r.t.~a given confidence level $\alpha$, CVaR measures it as the expected loss given that the loss is greater or equal to VaR$_\alpha$. Although VaR is a popular risk measure, CVaR's computational advantages over VaR has boosted the development of CVaR optimization techniques. We provide the exact definitions of these two risk measures and briefly discuss some of the VaR's shortcomings in Section~\ref{sec:preliminaries}. CVaR minimization was first developed by Rockafellar and Uryasev~\cite{Rockafellar00OC} and its numerical effectiveness was demonstrated in portfolio optimization and option hedging problems. Their work was then extended to objective functions consist of different combinations of the expected loss and the CVaR, such as %the minimization of CVaR subject to a constraint on expected loss and 
the minimization of the expected loss subject to a constraint on CVaR. This is the objective function that we study in this paper, although we believe that our proposed algorithms can be easily extended to several other CVaR-related objective functions. Boda and Filar~\cite{Boda06TC} and B{\"a}uerle and Ott~\cite{Ott10MD,Bauerle11MD} extended the results of~\cite{Rockafellar00OC} to MDPs (sequential decision-making). While the former proposed to use dynamic programming (DP) to optimize CVaR, an approach that is limited to small problems, the latter showed that in both finite and infinite horizon MDPs, there exists a {\em deterministic history-dependent} optimal policy for CVaR optimization (see Section~\ref{sec:CVaR-Opt} for more details). 

Most of the work in risk-sensitive sequential decision-making has been in the context of MDPs (when the model is known) and much less work has been done within the reinforcement learning (RL) framework. In risk-sensitive RL, we can mention the work by Borkar~\citep{Borkar01SF,Borkar02QR} who considered the expected exponential utility and those by Tamar et al.~\cite{tamar2012policy} and Prashanth and Ghavamzadeh~\cite{Prashanth13AC} on several variance-related risk measures. CVaR optimization in RL is a rather novel subject. Morimura et al.~\cite{Morimura10NR} estimate the return distribution while exploring using a CVaR-based risk-sensitive policy. Their algorithm does not scale to large problems. Petrik and Subramanian~\cite{Petrik12AS} propose a method based on stochastic dual DP to optimize CVaR in large-scale MDPs. However, their method is limited to linearly controllable problems. Borkar and Jain~\cite{Borkar14RC} consider a finite-horizon MDP with CVaR constraint and sketch a stochastic approximation algorithm to solve it. Finally, Tamar et al.~\cite{Tamar14PG} have recently proposed a policy gradient algorithm for CVaR optimization. 

In this paper, we develop policy gradient (PG) and actor-critic (AC) algorithms for mean-CVaR optimization in MDPs. We first derive a formula for computing the gradient of this risk-sensitive objective function. We then propose several methods to estimate this gradient both incrementally and using system trajectories (update at each time-step vs.~update after observing one or more trajectories). We then use these gradient estimations to devise PG and AC algorithms that update the policy parameters in the descent direction. Using the ordinary differential equations (ODE) approach, we establish the asymptotic convergence of our algorithms to locally risk-sensitive optimal policies. Finally, we demonstrate the usefulness of our algorithms in an optimal stopping problem. In comparison to~\cite{Tamar14PG}, while they develop a PG algorithm for CVaR optimization in stochastic shortest path problems that only considers continuous loss distributions, uses a biased estimator for VaR, is not incremental, and has no convergence proof, here we study mean-CVaR optimization, consider both discrete and continuous loss distributions, devise both PG and (several) AC algorithms (trajectory-based and incremental -- plus AC helps in reducing the variance of PG algorithms), and establish convergence proof for our algorithms.

\vspace{-0.1in}
\section{Preliminaries} 
\label{sec:preliminaries}
\vspace{-0.05in}

We consider problems in which the agent's interaction with the environment is modeled as a MDP. A MDP is a tuple $\mathcal{M}=(\X,\A,C,P,P_0)$, where $\X=\{1,\ldots,n\}$ and $\A=\{1,\ldots,m\}$ are the state and action spaces; $C(x,a)\in[-C_{\max},C_{\max}]$ is the bounded cost random variable whose expectation is denoted by $c(x,a)=\E\big[C(x,a)\big]$; $P(\cdot|x,a)$ is the transition probability distribution; and $P_0(\cdot)$ is the initial state distribution. For simplicity, we assume that the system has a single initial state $x^0$, i.e.,~$P_0(x)=\mathbf{1}\{x=x^0\}$. All the results of the paper can be easily extended to the case that the system has more than one initial state. 
%where $x_R$ is the sink state and $x_T$ is the target state; $C(x,a)\in[-C_{\max},C_{\max}]$ is the bounded reward function; $P(\cdot|x,a)$ is the transition probability distribution; and $P_0(\cdot)$ is the initial state distribution. For simplicity, in this paper we assume $P_0=\mathbf 1\{x=x^0\}$ for some given $x^0\in\{1,\ldots,n\}$. 
We also need to specify the rule according to which the agent selects actions at each state. A {\em stationary policy} $\mu(\cdot|x)$ is a probability distribution over actions, conditioned on the current state. In policy gradient and actor-critic methods, we define a class of parameterized stochastic policies $\big\{\mu(\cdot|x;\theta),x\in\X,\theta\in\Theta\subseteq\R^{\kappa_1}\big\}$, estimate the gradient of a performance measure w.r.t.~the policy parameters $\theta$ from the observed system trajectories, and then improve the policy by adjusting its parameters in the direction of the gradient. Since in this setting a policy $\mu$ is represented by its $\kappa_1$-dimensional parameter vector $\theta$, policy dependent functions can be written as a function of $\theta$ in place of $\mu$. So, we use $\mu$ and $\theta$ interchangeably in the paper. We denote by $d_\gamma^\mu(x|x^0)=(1-\gamma)\sum_{k=0}^\infty\gamma^k \mathbb P(x_k=x|x_0=x^0;\mu)$ and $\pi_\gamma^\mu(x,a|x^0)=d_\gamma^\mu(x|x^0)\mu(a|x)$ the $\gamma$-discounted visiting distribution of state $x$ and state-action pair $(x,a)$ under policy $\mu$, respectively. 

%We have the following assumptions on transient policies and stopping times.
%\begin{definition}
%Define $\X^\prime=\X\cap\{x_R\}^c=\{1,\ldots,n, x_T\}$ as the space for transient states. A stationary policy $\mu$ is said to be transient if,
%\begin{enumerate}
%\item $\sum_{k=0}^\infty \mathbb P(x_n=x|x_0=x^0,\mu)< \infty$ for every $x\in \X^\prime$, and
%\item  $P(x_R|x_T,a)=1$ and $P(x_R|x_R,a)=1$ for every admissible control action $a\in\A$.
%\end{enumerate}
%\end{definition}
%Let $T$ be the first visit time  to the target state. Since the state trajectory is non-deterministic, $T$ is a random variable. By the above assumption, we also know that for any $t>T$, $x_k=x_R$, where $x_R$ is a sink absorbing state. Furthermore,  we have the following assumption on the stage-wise cost function.
%\begin{assumption}
%When $x\in\{x_R,x_T\}$, the immediate cost function is zero, i.e., $C(x,a)=0$ for any $a\in\A$.
%\end{assumption}

Let $Z$ be a bounded-mean random variable, i.e.,~$\E[|Z|]<\infty$, with the cumulative distribution function $F(z)=\mathbb{P}(Z\leq z)$ (e.g.,~one may think of $Z$ as the loss of an investment strategy $\mu$). We define the {\em value-at-risk} at the confidence level $\alpha\in (0,1)$ as VaR$_\alpha(Z) = \min\big\{z\mid F(z)\geq\alpha\big\}$. 
%
%\vspace{-0.1in}
%\begin{small}
%\begin{equation}
%\label{eq:VaR}
%\text{VaR}_\alpha(Z) = \min\big\{z\mid F(z)\geq\alpha\big\}.
%\end{equation}
%\end{small}
%\vspace{-0.15in}
%
%The minimum in~\eqref{eq:VaR} is attained because $F$ is non-decreasing and right-continuous in $z$. When $F$ is continuous and strictly increasing, VaR$_\alpha(Z)$ is the unique $z$ satisfying $F(z)=\alpha$, otherwise, \eqref{eq:VaR} can have no solution or a whole range of solutions. Although VaR is a popular risk measure, it suffers from being unstable and difficult to work with numerically when $Z$ is not normally distributed, which is often the case as loss distributions tend to exhibit fat tails or empirical discreteness. Moreover, VaR is not a {\em coherent} risk measure~\cite{Artzner99CM} and more importantly does not quantify the losses that might be suffered beyond its value at the $\alpha$-tail of the distribution~\cite{Rockafellar02CV}. An alternative measure that addresses most of the VaR's shortcomings is {\em conditional value-at-risk}, CVAR$_\alpha(Z)$, which is the mean of the $\alpha$-tail distribution of $Z$. If there is no probability atom at VaR$_\alpha(Z)$, CVaR$_\alpha(Z)$ has a unique value that is defined as 
%
Here the minimum is attained because $F$ is non-decreasing and right-continuous in $z$. When $F$ is continuous and strictly increasing, VaR$_\alpha(Z)$ is the unique $z$ satisfying $F(z)=\alpha$, otherwise, the VaR equation can have no solution or a whole range of solutions. Although VaR is a popular risk measure, it suffers from being unstable and difficult to work with numerically when $Z$ is not normally distributed, which is often the case as loss distributions tend to exhibit fat tails or empirical discreteness. Moreover, VaR is not a {\em coherent} risk measure~\cite{Artzner99CM} and more importantly does not quantify the losses that might be suffered beyond its value at the $\alpha$-tail of the distribution~\cite{Rockafellar02CV}. An alternative measure that addresses most of the VaR's shortcomings is {\em conditional value-at-risk}, CVAR$_\alpha(Z)$, which is the mean of the $\alpha$-tail distribution of $Z$. If there is no probability atom at VaR$_\alpha(Z)$, CVaR$_\alpha(Z)$ has a unique value that is defined as CVaR$_\alpha(Z) = \E\big[Z\mid Z\geq \text{VaR}_\alpha(Z)\big]$. 
%
%\vspace{-0.1in}
%\begin{small}
%\begin{equation}
%\label{eq:CVaR1}
%\text{CVaR}_\alpha(Z) = \E\big[Z\mid Z\geq \text{VaR}_\alpha(Z)\big].
%\end{equation}
%\end{small}
%\vspace{-0.15in}
%
Rockafellar and Uryasev~\cite{Rockafellar00OC} showed that 

\vspace{-0.15in}
\begin{small}
\begin{equation}
\label{eq:CVaR2}
\text{CVaR}_\alpha(Z) = \min_{\nu\in\reals} H_\alpha(Z,\nu) \stackrel{\triangle}{=} \min_{\nu\in\reals}\Big\{\nu + \frac{1}{1-\alpha}\E\big[(Z-\nu)^+\big]\Big\}.
\end{equation}
\end{small}
\vspace{-0.15in}

Note that as a function of $\nu$, $H_\alpha(\cdot,\nu)$ is finite and convex (hence continuous). 

%%%%%%%%%%%%%%%%%%%%%%%%%%%%%%%%%%%%%%%%%%%%%%%%%%%%%%%%%%%%%%
%%%%%%%%%%%%%%%%%%%%%%%%%%%%%%%%%%%%%%%%%%%%%%%%%%%%%%%%%%%%%%
%%%%%%%%%%%%%%%%%%%%%%%%%%%%%%%%%%%%%%%%%%%%%%%%%%%%%%%%%%%%%%
%%%%%%%%%%%%%%%%%%%%%%%%%%%%%%%%%%%%%%%%%%%%%%%%%%%%%%%%%%%%%%
%%%%%%%%%%%%%%%%%%%%%%%%%%%%%%%%%%%%%%%%%%%%%%%%%%%%%%%%%%%%%%

\vspace{-0.1in}
\section{CVaR Optimization in MDPs} 
\label{sec:CVaR-Opt}
\vspace{-0.075in}

For a policy $\mu$, we define the loss of a state $x$ (state-action pair $(x,a)$) as the sum of (discounted) costs encountered by the agent when it starts at state $x$ (state-action pair $(x,a)$) and then follows policy \begin{small}$\mu$, i.e.,~$\;D^\theta(x)=\sum_{k=0}^\infty\gamma^kC(x_k,a_k)\mid x_0=x,\;\mu\;$\end{small} and \begin{small}$\;D^\theta(x,a)=\sum_{k=0}^\infty\gamma^kC(x_k,a_k)\mid x_0=x,\;a_0=a,\;\mu$\end{small}. 
%
%\vspace{-0.1in}
%\begin{small}
%\begin{equation*}
%D^\theta(x)=\sum_{k=0}^\infty\gamma^kC(x_k,a_k)\mid x_0=x,\;\mu, \quad\quad\quad\quad D^\theta(x,a)=\sum_{k=0}^\infty\gamma^kC(x_k,a_k)\mid x_0=x,\;a_0=a,\;\mu.
%\end{equation*}
%\end{small}
%\vspace{-0.15in}
%
The expected value of these two random variables are the value and action-value functions of policy $\mu$, i.e.,~$V^\theta(x)=\E\big[D^\theta(x)\big]$ and $Q^\theta(x,a)=\E\big[D^\theta(x,a)\big]$. The goal in the standard discounted formulation is to find an optimal policy $\theta^*=\argmin_\theta V^\theta(x^0)$.%, where $x^0\in\X^\prime$ is the initial state of the system. This can be easily extended to the case that the system has more than one initial state $\theta^*=\argmax_\theta \sum\limits_{x\in \X} P_0(x) V^\theta(x)$.

%Measuring the {\em variability} in the stream of rewards is more difficult in discounted than average reward MDPs. The most common measure is the {\em variance of the return}

For CVaR optimization in MDPs, we consider the following optimization problem: For a given confidence level $\alpha\in (0,1)$ and loss tolerance $\beta\in\reals$, 

%Now, based on \cite{}, we define the Conditional value-at-risk (CVAR) as follows
% \[
% \text{CVaR}_{\alpha}(Z)=\inf_{ \nu\in\reals}\left\{ \nu+\frac{1}{1-\alpha}\expec\left[\left[Z- \nu\right]^+\right]\right\}.
% \]
% 
%We consider the following risk-sensitive constrained optimal control problem for discounted MDPs: for any given $\alpha\in(0,1)$ and $K\in\reals$,
\vspace{-0.125in}
\begin{small}
\begin{equation}
\label{eq:norm_reward_eqn1}
\min_\theta V^\theta(x^0)\quad\quad \text{subject to} \quad\quad \text{CVaR}_\alpha\big(D^\theta(x^0)\big)\leq\beta.
\end{equation}
\end{small}
\vspace{-0.15in}

By Theorem~16 in~\cite{Rockafellar02CV}, the optimization problem~\eqref{eq:norm_reward_eqn1} is equivalent to ($H_\alpha$ is defined by \eqref{eq:CVaR2})

\vspace{-0.125in}
\begin{small}
\begin{equation}
\label{eq:norm_reward_eqn2}
\min_{\theta,\nu} V^\theta(x^0)\quad\quad \text{subject to} \quad\quad H_\alpha\big(D^\theta(x^0),\nu\big)\leq\beta.
\end{equation}
\end{small}
\vspace{-0.15in}

To solve~\eqref{eq:norm_reward_eqn2}, we employ the Lagrangian relaxation procedure~\citep{bertsekas1999nonlinear} to convert it to the following unconstrained problem:  

\vspace{-0.125in}
\begin{small}
\begin{equation}
\label{eq:unconstrained-discounted-risk-measure}
\max_\lambda\min_{\theta,\nu}\bigg(L(\theta,\nu,\lambda)\stackrel{\triangle}{=} V^\theta(x^0)+\lambda\Big(H_\alpha\big(D^\theta(x^0),\nu\big)-\beta\Big)\bigg),
\end{equation}
\end{small}
\vspace{-0.15in}

where $\lambda$ is the Lagrange multiplier. The goal here is to find the saddle point of {\small $L(\theta,\nu,\lambda)$}, i.e.,~a point {\small $(\theta^*,\nu^*,\lambda^*)$} that satisfies {\small $L(\theta,\nu, \lambda^*) \ge L(\theta^*,\nu^*, \lambda^*) \ge L(\theta^*,\nu^*, \lambda),\forall\theta,\nu,\forall \lambda>0$}. This is achieved by descending in {\small $(\theta,\nu)$} and ascending in {\small $\lambda$} using the gradients of $L(\theta,\nu, \lambda)$ w.r.t.~$\theta$, $\nu$, and $\lambda$, i.e.,\footnote{The notation $\ni$ in \eqref{eq:grad-nu} means that the right-most term is a member of the sub-gradient set $\partial_\nu L(\theta,\nu,\lambda)$.}

\vspace{-0.15in}
\begin{small}
\begin{align}
\label{eq:grad-theta}
\nabla_\theta L(\theta,\nu,\lambda) &= \nabla_\theta V^\theta(x^0) + \frac{\lambda}{(1-\alpha)} \nabla_\theta\E\Big[\big(D^\theta(x^0)- \nu\big)^+\Big], \\
\label{eq:grad-nu}
\partial_\nu L(\theta,\nu,\lambda) &= \lambda\bigg(1 + \frac{1}{(1-\alpha)}\partial_\nu\E\Big[\big(D^\theta(x^0)- \nu\big)^+\Big]\bigg) \ni \lambda\bigg(1 - \frac{1}{(1-\alpha)}\mathbb{P}\big(D^\theta(x^0) \geq \nu\big)\bigg), \\
\label{eq:grad-lambda}
\nabla_\lambda L(\theta,\nu, \lambda) &= \nu + \frac{1}{(1-\alpha)}\E\Big[\big(D^\theta(x^0) - \nu\big)^+\Big] - \beta.
\end{align}
\end{small}
\vspace{-0.175in}

We assume that there exists a policy $\mu(\cdot|\cdot;\theta)$ such that CVaR$_\alpha\big(D^\theta(x^0)\big)\leq\beta$ (feasibility assumption). As discussed in Section~\ref{sec:introduction}, B{\"a}uerle and Ott~\cite{Ott10MD,Bauerle11MD} showed that there exists a {\em deterministic history-dependent} optimal policy for CVaR optimization. The important point is that this policy does not depend on the complete history, but only on the current time step $k$, current state of the system $x_k$, and accumulated discounted cost $\sum_{i=0}^k\gamma^i c(x_i,a_i)$.  

In the following, we present a policy gradient (PG) algorithm (Sec.~\ref{sec:PG-alg}) and several actor-critic (AC) algorithms (Sec.~\ref{sec:AC-alg}) to optimize~\eqref{eq:unconstrained-discounted-risk-measure}. While the PG algorithm updates its parameters after observing several trajectories, the AC algorithms are incremental and update their parameters at each time-step.

%%%%%%%%%%%%%%%%%%%%%%%%%%%%%%%%%%%%%%%%%%%%%%%%%%%%%%%%%%%%%%
%%%%%%%%%%%%%%%%%%%%%%%%%%%%%%%%%%%%%%%%%%%%%%%%%%%%%%%%%%%%%%
%%%%%%%%%%%%%%%%%%%%%%%%%%%%%%%%%%%%%%%%%%%%%%%%%%%%%%%%%%%%%%
%%%%%%%%%%%%%%%%%%%%%%%%%%%%%%%%%%%%%%%%%%%%%%%%%%%%%%%%%%%%%%
%%%%%%%%%%%%%%%%%%%%%%%%%%%%%%%%%%%%%%%%%%%%%%%%%%%%%%%%%%%%%%

\vspace{-0.1in}
\section{A Trajectory-based Policy Gradient Algorithm}
\label{sec:PG-alg}
\vspace{-0.075in}

In this section, we present a policy gradient algorithm to solve the optimization problem~\eqref{eq:unconstrained-discounted-risk-measure}. The unit of observation in this algorithm is a system trajectory generated by following the current policy. At each iteration, the algorithm generates $N$ trajectories by following the current policy, use them to estimate the gradients in \eqref{eq:grad-theta}-\eqref{eq:grad-lambda}, and then use these estimates to update the parameters $\theta,\nu,\lambda$.

Let $\xi=\{x_0,a_0,c_0,x_1,a_1,c_1,\ldots,x_{T-1},a_{T-1},c_{T-1},x_T\}$ be a trajectory generated by following the policy $\theta$, where $x_0=x^0$ and $x_T$ is usually a terminal state of the system.  After $x_k$ visits the terminal state, it enters a recurring sink state $x_R$ at the next time step, incurring zero cost, i.e., $C(x_R,a)=0$, $\forall a\in\A$. Time index $T$ is referred as the stopping time of the MDP. Since the transition is stochastic, $T$ is a non-deterministic quantity. Here we assume that the policy $\mu$ is proper, i.e., $\sum_{k=0}^\infty \mathbb P(x_k=x|x_0=x^0,\mu)< \infty$ for every $x\not\in \{x_S,x_T\}$. This further means that with probability $1$, the MDP exits the transient states and hits $x_T$ (and stays in $x_S$) in finite time $T$.
For simplicity, we assume that the agent incurs zero cost in the terminal state. Analogous results for the general case with a non-zero terminal cost can be derived using identical arguments. The loss and probability of $\xi$ are defined as $D(\xi)=\sum_{k=0}^{T-1}\gamma^kc(x_k,a_k)$ and $\mathbb{P}_\theta(\xi)=P_0(x_0)\prod_{k=0}^{T-1}\mu(a_k|x_k;\theta)P(x_{k+1}|x_k,a_k)$, respectively. It can be easily shown that $\nabla_\theta\log\mathbb{P}_\theta(\xi)=\sum_{k=0}^{T-1}\nabla_\theta\log\mu(a_k|x_k;\theta)$.

Algorithm~\ref{alg_traj} contains the pseudo-code of our proposed policy gradient algorithm. What appears inside the parentheses on the right-hand-side of the update equations are the estimates of the gradients of $L(\theta,\nu,\lambda)$ w.r.t.~$\theta,\nu,\lambda$ (estimates of \eqref{eq:grad-theta}-\eqref{eq:grad-lambda}) (see Appendix~\ref{subsec:grad-comp}). %First define the projection operator $\Gamma_{\phi}$ on set $\Phi$ that projects a vector $\phi$ to the closest point (in Euclidean norm) in a compact and convex set $\Phi$. In other words, 
$\Gamma_\Theta$ is an operator that projects a vector $\theta\in\reals^{\kappa_1}$ to the closest point in a compact and convex set $\Theta\subset\reals^{\kappa_1}$, and $\Gamma_N$ and $\Gamma_\Lambda$ are projection operators to $[-\frac{C_{\max}}{1-\gamma},\frac{C_{\max}}{1-\gamma}]$ and $[0,\lambda_{\max}]$, respectively. These projection operators are necessary to ensure the convergence of the algorithm. The step-size schedules satisfy the standard conditions for stochastic approximation algorithms, and ensures that the VaR parameter $\nu$ update is on the fastest time-scale $\big\{\zeta_3(i)\big\}$, the policy parameter $\theta$ update is on the intermediate time-scale $\big\{\zeta_2(i)\big\}$, and the Lagrange multiplier $\lambda$ update is on the slowest time-scale $\big\{\zeta_1(i)\big\}$ (see Appendix~\ref{subsec:ass} for the conditions on the step-size schedules). This results in a three time-scale stochastic approximation algorithm. We prove that our policy gradient algorithm converges to a (local) saddle point of the risk-sensitive objective function $L(\theta,\nu,\lambda)$ (see Appendix~\ref{subsec:conv-proof}).

\vspace{-0.05in}
\begin{algorithm}
\begin{small}
\begin{algorithmic}
\STATE {\bf Input:} parameterized policy $\mu(\cdot|\cdot;\theta)$, confidence level $\alpha$, loss tolerance $\beta$ and Lagrangian threshold $\lambda_{\max}$
%Let $x^0\in\X$ be an initial state, $\alpha\in(0,1)$ be the given CVaR parameter and $K$ be the given threshold parameter. 
\STATE {\bf Initialization:} policy parameter $\theta=\theta_0$, VaR parameter $\nu=\nu_0$, and the Lagrangian parameter $\lambda=\lambda_0$
%\STATE {\bf Step Length Conditions:} Step lengths $\zeta_{1,i}$, $\zeta_{2,i}$, $\zeta_{3,i}$ are square summable but not summable. Furthermore, $\{\zeta_{1,i}\}$ is on the fastest time-scale, the update corresponds to $\{\zeta_{2,i}\}$ is on the intermediate time-scale, and the update corresponds to $\{\zeta_{3,i}\}$ is on the slowest time-scale. More details can be found in the Appendix.
\WHILE{1}
\FOR{$i = 0,1,2,\ldots$}
\FOR{$j = 1,2,\ldots$}
\STATE Generate $N$ trajectories $\{\xi_{j,i}\}_{j=1}^N$ by starting at $x_0=x^0$ and following the current policy $\theta_i$.
\ENDFOR  
%\STATE Simulate the state-action sample trajectory $h_k=\left\{x^0,a^0,x^1,a^1,\ldots,x^{T-1},a^{T-1} \right\}$ from the multi-variate probability mass function $f_{\hr,\theta}$ with $\theta=\theta_i$, where $x^{T}=x_T$ is the stopping state .
%\begin{small}
%\begin{footnotesize}
\begin{align*}
\textrm{{\bf $\nu$ Update:}}\quad& \nu_{i+1} = \Gamma_N\bigg[\nu_i - \zeta_3(i)\bigg(\lambda_i - \frac{\lambda_i}{(1-\alpha)N}\sum_{j=1}^N\mathbf{1}\big\{D(\xi_{j,i})\geq\nu_i\big\}\bigg)\bigg] \\
\textrm{\bf $\theta$ Update:} \quad &  \theta_{i+1} = \Gamma_\Theta\bigg[\theta_i -\zeta_2(i)\bigg(\frac{1}{N}\sum_{j=1}^N\nabla_\theta\log\mathbb{P}_\theta(\xi_{j,i})\vert_{\theta=\theta_i}D(\xi_{j,i}) \nonumber \\ 
&\hspace{0.725in}+ \frac{\lambda_i}{(1-\alpha)N}\sum_{j=1}^N\nabla_\theta\log\mathbb{P}_\theta(\xi_{j,i})\vert_{\theta=\theta_i}\big(D(\xi_{j,i})-\nu_i\big)\mathbf{1}\big\{D(\xi_{j,i})\geq\nu_i\big\}\bigg)\bigg] \\
\textrm{{\bf $\lambda$ Update:}}\quad&\lambda_{i+1} = \Gamma_\Lambda\bigg[\lambda_i + \zeta_1(i)\bigg(\nu_i - \beta + \frac{1}{(1-\alpha)N}\sum_{j=1}^N\big(D(\xi_{j,i})-\nu_i\big)\mathbf{1}\big\{D(\xi_{j,i})\geq\nu_i\big\}\bigg)\bigg]
\end{align*}
%\end{footnotesize}
%\begin{align}
%\textrm{\bf $\theta$ Update:} \quad &  \theta_{i+1}=\Pi^{(i)}_{\Theta}\left(\theta_{i}-\zeta_{1,i} \frac{\partial \log (f_{\hr,\theta}(h_k))}{\partial \theta}\bigg\vert_{\theta=\theta_i}\bigg(D(h_k) \right.\nonumber\\
%&\left.\quad\quad\quad\quad\quad\quad\quad+ \frac{\lambda_i}{1-\alpha} \left(D(h_k)- \nu_{i}\right)^+\bigg)\right),\,i\in\{1,\ldots,\kappa_1\}.\label{theta_up_h}\\
%\textrm{{\bf $\nu$ Update:}}\quad& \nu_{i+1}=\nu_{i}-\zeta_{2,i}\left(\lambda_i-\frac{\lambda_i}{1-\alpha}\mathbf 1\{D(h_k)\geq \nu_{i}\}\right). \label{s_up_h} \\
%\textrm{{\bf $\lambda$ Update:}}\quad&\lambda_{i+1}=\left(\lambda_i+\zeta_{3,i} \left(\nu_{i}+\frac{1}{1-\alpha} \left(D(h_k)- \nu_{i}\right)^+ -K\right)\right)^+. \label{lambda_up_h}
%\end{align}
%\end{small}
\ENDFOR  
\IF{$\{\lambda_i\}$ converges to $\lambda_{\max}$}
\STATE{Set $\lambda_{\max}\leftarrow 2\lambda_{\max}$.}
\ELSE
\STATE {\bf return} parameters $\nu,\theta,\lambda$ and {\bf break}
\ENDIF
\ENDWHILE
\end{algorithmic}
\end{small}
%\vspace{-0.1in}
\caption{Trajectory-based Policy Gradient Algorithm for CVaR Optimization}
\label{alg_traj}
%\vspace{-0.1in}
\end{algorithm}

\vspace{-0.15in}
\section{Incremental Actor-Critic Algorithms}
\label{sec:AC-alg}
\vspace{-0.075in}

As mentioned in Section~\ref{sec:PG-alg}, the unit of observation in our policy gradient algorithm (Algorithm~\ref{alg_traj}) is a system trajectory. This may result in high variance for the gradient estimates, especially when the length of the trajectories is long. To address this issue, in this section, we propose actor-critic algorithms that use linear approximation for some quantities in the gradient estimates and update the parameters incrementally (after each state-action transition). To develop our actor-critic algorithms, we should show how the gradients of \eqref{eq:grad-theta}-\eqref{eq:grad-lambda} are estimated in an incremental fashion. We show this in the next four subsections, followed by a subsection that contains the algorithms. 

%%%%%%%%%%%%%%%%%%%%%%%%%%%%%%%%%%%%%%%%%%%%%%%%%%%%%%%%%%%%%%
%%%%%%%%%%%%%%%%%%%%%%%%%%%%%%%%%%%%%%%%%%%%%%%%%%%%%%%%%%%%%%
%%%%%%%%%%%%%%%%%%%%%%%%%%%%%%%%%%%%%%%%%%%%%%%%%%%%%%%%%%%%%%

\vspace{-0.1in}
\subsection{Gradient w.r.t.~the Policy Parameters $\theta$}
\label{subsec:grad-theta}
\vspace{-0.05in}

The gradient of our objective function w.r.t.~the policy parameters $\theta$ in \eqref{eq:grad-theta} may be rewritten as

\vspace{-0.125in}
\begin{small}
\begin{equation}
\label{eq:grad-theta1}
\nabla_\theta L(\theta,\nu,\lambda) = \nabla_\theta\left(\E\big[D^\theta(x^0)\big] + \frac{\lambda}{(1-\alpha)}\E\Big[\big(D^\theta(x^0)- \nu\big)^+\Big]\right).
\end{equation}
\end{small}
\vspace{-0.15in}

Given the original MDP $\mathcal{M}=(\X,\A,C,P,P_0)$ and the parameter $\lambda$, we define the augmented MDP $\bar{\mathcal{M}}=(\bar{\X},\bar{\A},\bar{C},\bar{P},\bar{P}_0)$ as $\bar{\X}=\X\times\reals$, $\bar{\A}=\A$, $\bar{P}_0(x,s)=P_0(x)\mathbf{1}\{s=s^0\}$, and

\vspace{-0.15in}
\begin{small}
\begin{alignat*}{2}
\bar{C}(x,s,a)&=\left\{\begin{array}{cc}
\lambda(-s)^+/ (1-\alpha)& \text{if $x=x_T$}\\
C(x,a)& \text{otherwise}
\end{array}\right.\!\!
,&\, 
\bar{P}(x^\prime,s^\prime|x,s,a)=\left\{\begin{array}{ll}
\P(x^\prime|x,a)&\text{if $s^\prime=\big(s-C(x,a)\big)/\gamma$}\\
0&\text{otherwise}
\end{array}\right.
\end{alignat*}
\end{small}

where $x_T$ is any terminal state of the original MDP $\mathcal{M}$ and $s_T$ is the value of the $s$ part of the state when a policy $\theta$ reaches a terminal state $x_T$ after $T$ steps, i.e., $s_T=\frac{1}{\gamma^T}\big(s^0-\sum_{k=0}^{T-1}\gamma^kC(x_k,a_k)\big)$. We define a class of parameterized stochastic policies $\big\{\mu(\cdot|x,s;\theta),(x,s)\in\bar{\X},\theta\in\Theta\subseteq\R^{\kappa_1}\big\}$ for this augmented MDP. Thus, the total (discounted) loss of this trajectory can be written as

\vspace{-0.125in}
\begin{small}
\begin{equation}
\label{eq:aug-loss}
\sum_{k=0}^{T-1}\gamma^kC(x_k,a_k)+\gamma^T\bar{C}(x_T,s_T,a) = D^\theta(x^0) + \frac{\lambda}{(1-\alpha)}\big(D^\theta(x^0)-s^0\big)^+.
\end{equation}
\end{small}
\vspace{-0.15in}

From \eqref{eq:aug-loss}, it is clear that the quantity in the parenthesis of \eqref{eq:grad-theta1} is the value function of the policy $\theta$ at state $(x^0,s^0=\nu)$ in the augmented MDP $\bar{\mathcal{M}}$, i.e.,~$V^\theta(x^0,\nu)$. Thus, it is easy to show that (the proof of the second equality can be found in the literature, e.g.,~\cite{Peters05NA})

\vspace{-0.125in}
\begin{small}
\begin{equation}
\nabla_\theta L(\theta,\nu,\lambda) = \nabla_\theta V^\theta(x^0,\nu) = \frac{1}{1-\gamma}\sum_{x,s,a}\pi_\gamma^\theta(x,s,a|x^0,\nu)\;\nabla\log\mu(a|x,s;\theta)\;Q^\theta(x,s,a), 
\end{equation}
\end{small}
\vspace{-0.15in}

where $\pi_\gamma^\theta$ is the discounted visiting distribution (defined in Section~\ref{sec:preliminaries}) and $Q^\theta$ is the action-value function of policy $\theta$ in the augmented MDP $\bar{\mathcal{M}}$. We can show that $\frac{1}{1-\gamma}\nabla\log\mu(a_k|x_k,s_k;\theta)\cdot\delta_k$ is an unbiased estimate of $\nabla_\theta L(\theta,\nu,\lambda)$, where $\delta_k=\bar{C}(x_k,s_k,a_k)+\gamma\widehat{V}(x_{k+1},s_{k+1})-\widehat{V}(x_k,s_k)$ is the temporal-difference (TD) error in $\bar{\mathcal{M}}$, and $\widehat{V}$ is an unbiased estimator of $V^\theta$ (see e.g.~\cite{bhatnagar2009natural}). In our actor-critic algorithms, the critic uses linear approximation for the value function $V^\theta(x,s)\approx v^\top\phi(x,s)=\widetilde{V}^{\theta,v}(x,s)$, where the feature vector $\phi(\cdot)$ is from low-dimensional space $\reals^{\kappa_2}$.

%\begin{remark}
%As mentioned in Sec.~\ref{sec:CVaR-Opt}, there exists a history-dependent optimal policy for the CVaR optimization problem that does not depend on the complete history, just the accumulated discounted cost. Thus, we may define a parameterized class of policies over the augmented MDP $\bar{\mathcal{M}}$ and implement the policy gradient Alg.~\ref{alg_traj} for this MDP. This way we have a better chance that our class of policies contains an optimal policy. 
%\end{remark}

%%%%%%%%%%%%%%%%%%%%%%%%%%%%%%%%%%%%%%%%%%%%%%%%%%%%%%%%%%%%%%
%%%%%%%%%%%%%%%%%%%%%%%%%%%%%%%%%%%%%%%%%%%%%%%%%%%%%%%%%%%%%%
%%%%%%%%%%%%%%%%%%%%%%%%%%%%%%%%%%%%%%%%%%%%%%%%%%%%%%%%%%%%%%

\vspace{-0.1in}
\subsection{Gradient w.r.t.~the Lagrangian Parameter $\lambda$}
\label{subsec:grad-lambda}
\vspace{-0.05in}

We may rewrite the gradient of our objective function w.r.t.~the Lagrangian parameters $\lambda$ in \eqref{eq:grad-lambda} as

\vspace{-0.2in}
\begin{small}
\begin{equation}
\label{eq:grad-lambda1}
\nabla_\lambda L(\theta,\nu,\lambda) = \nu - \beta + \nabla_\lambda\left(\E\big[D^\theta(x^0)\big] + \frac{\lambda}{(1-\alpha)}\E\Big[\big(D^\theta(x^0)- \nu\big)^+\Big]\right)\stackrel{\text{(a)}}{=} \nu - \beta + \nabla_\lambda V^\theta(x^0,\nu).
\end{equation}
\end{small}
\vspace{-0.175in}

Similar to Section~\ref{subsec:grad-theta}, {\bf (a)} comes from the fact that the quantity in the parenthesis in \eqref{eq:grad-lambda1} is $V^\theta(x^0,\nu)$, the value function of the policy $\theta$ at state $(x^0,\nu)$ in the augmented MDP $\bar{\mathcal{M}}$. Note that the dependence of $V^\theta(x^0,\nu)$ on $\lambda$ comes from the definition of the cost function $\bar{C}$ in $\bar{\mathcal{M}}$. We now derive an expression for $\nabla_\lambda V^\theta(x^0,\nu)$, which in turn will give us an expression for $\nabla_\lambda L(\theta,\nu,\lambda)$.
\begin{lemma}
\label{lem:grad_lambda}
The gradient of $V^\theta(x^0,\nu)$ w.r.t.~the Lagrangian parameter $\lambda$ may be written as

\vspace{-0.125in}
\begin{small}
\begin{equation}
\label{eq:grad-lambda-V}
\nabla_\lambda V^\theta(x^0,\nu) = \frac{1}{1-\gamma}\sum_{x,s,a}\pi_\gamma^\theta(x,s,a|x^0,\nu)\frac{1}{(1-\alpha)}\mathbf{1}\{x=x_T\}(-s)^+.
\end{equation}
\end{small}
\vspace{-0.225in}
\end{lemma}
\begin{prooff}
See Appendix~\ref{subsec:grad-lambda-comp}.
\end{prooff}

From Lemma~\ref{lem:grad_lambda} and \eqref{eq:grad-lambda1}, it is easy to see that $\nu-\beta+\frac{1}{(1-\gamma)(1-\alpha)}\mathbf{1}\{x=x_T\}(-s)^+$ is an unbiased estimate of $\nabla_\lambda L(\theta,\nu,\lambda)$. An issue with this estimator is that its value is fixed to $\nu_k-\beta$ all along a system trajectory, and only changes at the end to $\nu_k-\beta+\frac{1}{(1-\gamma)(1-\alpha)}(-s_T)^+$. This may affect the incremental nature of our actor-critic algorithm. To address this issue, we propose a different approach to estimate the gradients w.r.t.~$\theta$ and $\lambda$ in Sec.~\ref{subsec:grad-alter} (of course this does not come for free).

Another important issue is that the above estimator is unbiased only if the samples are generated from the distribution $\pi_\gamma^\theta(\cdot|x^0,\nu)$. If we just follow the policy, then we may use $\nu_k-\beta+\frac{\gamma^k}{(1-\alpha)}\mathbf{1}\{x_k=x_T\}(-s_k)^+$ as an estimate for $\nabla_\lambda L(\theta,\nu,\lambda)$ (see \eqref{lambda_up_incre} and~\eqref{lambda_up_incre_semi} in Algorithm~\ref{alg:AC}). Note that this is an issue for all discounted actor-critic algorithms that their (likelihood ratio based) estimate for the gradient is unbiased only if the samples are generated from $\pi_\gamma^\theta$, and not just when we simply follow the policy. Although this issue was known in the community, there is a recent paper that investigates it in details~\cite{Thomas14BN}. Moreover, this might be a main reason that we have no convergence analysis (to the best of our knowledge) for (likelihood ratio based) discounted actor-critic algorithms.\footnote{Note that the discounted actor-critic algorithm with convergence proof in~\cite{Bhatnagar10AC} is based on SPSA.}

\vspace{-0.1in}
\subsection{Sub-Gradient w.r.t.~the VaR Parameter $\nu$}
\label{subsec:grad-nu}
\vspace{-0.05in}

We may rewrite the sub-gradient of our objective function w.r.t.~the VaR parameters $\nu$ in \eqref{eq:grad-nu} as

\vspace{-0.175in}
\begin{small}
\begin{equation}
\label{eq:grad-nu1}
\partial_\nu L(\theta,\nu,\lambda) \ni \lambda\bigg(1-\frac{1}{(1-\alpha)}\mathbb{P}\Big(\sum_{k=0}^\infty\gamma^kC(x_k,a_k)\geq\nu\mid x_0=x^0;\theta\Big)\bigg).
\end{equation}
\end{small}
\vspace{-0.15in}

From the definition of the augmented MDP $\bar{\mathcal{M}}$, the probability in \eqref{eq:grad-nu1} may be written as $\mathbb{P}(s_T\leq 0\mid x_0=x^0,s_0=\nu;\theta)$, where $s_T$ is the $s$ part of the state in $\bar{\mathcal{M}}$ when we reach a terminal state, i.e.,~$x=x_T$ (see Section~\ref{subsec:grad-theta}). Thus, we may rewrite \eqref{eq:grad-nu1} as 

\vspace{-0.15in}
\begin{small}
\begin{equation}
\label{eq:grad-nu2}
\partial_\nu L(\theta,\nu,\lambda) \ni \lambda\Big(1-\frac{1}{(1-\alpha)}\mathbb{P}\big(s_T\leq 0\mid x_0=x^0,s_0=\nu;\theta\big)\Big).
\end{equation}
\end{small}
\vspace{-0.15in}

From \eqref{eq:grad-nu2}, it is easy to see that $\lambda-\lambda\mathbf 1\{s_T\leq 0\}/(1-\alpha)$ is an unbiased estimate of the sub-gradient of $L(\theta,\nu,\lambda)$ w.r.t.~$\nu$. An issue with this (unbiased) estimator is that it can be only applied at the end of a system trajectory (i.e.,~when we reach the terminal state $x_T$), and thus, using it prevents us of having a fully incremental algorithm. In fact, this is the estimator that we use in our {\em semi trajectory-based} actor-critic algorithm (see \eqref{nu_up_incre_SPSA_semi} in Algorithm~\ref{alg:AC}). 

One approach to estimate this sub-gradient incrementally, hence having a fully incremental algorithm, is to use {\em simultaneous perturbation stochastic approximation} (SPSA) method~\cite{Bhatnagar13SR}. The idea of SPSA is to estimate the sub-gradient $g(\nu)\in\partial_{\nu}  L(\theta,\nu,\lambda)$ using two values of $g$ at $\nu^-=\nu-\Delta$ and $\nu^+=\nu+\Delta$, where $\Delta>0$ is a positive perturbation (see Sec.~\ref{sec:AC-alg} for the detailed description of $\Delta$).\footnote{SPSA-based gradient estimate was first proposed in~\cite{Spall92MS} and has been widely used in various settings, especially those involving high-dimensional parameter. The SPSA estimate described above is two-sided. It can also be implemented single-sided, where we use the values of the function at $\nu$ and $\nu^+$. We refer the readers to~\cite{Bhatnagar13SR} for more details on SPSA and to~\cite{Prashanth13AC} for its application in learning in risk-sensitive MDPs.} In order to see how SPSA can help us to estimate our sub-gradient incrementally, note that 

\vspace{-0.15in}
\begin{small}
\begin{equation}
\label{eq:grad-nu3}
\partial_\nu L(\theta,\nu,\lambda) = \lambda + \partial_\nu\left(\E\big[D^\theta(x^0)\big] + \frac{\lambda}{(1-\alpha)}\E\Big[\big(D^\theta(x^0)- \nu\big)^+\Big]\right) \stackrel{\text{(a)}}{=} \lambda + \partial_\nu V^\theta(x^0,\nu).
\end{equation}
\end{small}
\vspace{-0.15in}

Similar to Sections~\ref{subsec:grad-theta} and~\ref{subsec:grad-lambda}, {\bf (a)} comes from the fact that the quantity in the parenthesis in \eqref{eq:grad-nu3} is $V^\theta(x^0,\nu)$, the value function of the policy $\theta$ at state $(x^0,\nu)$ in the augmented MDP $\bar{\mathcal{M}}$. Since the critic uses a linear approximation for the value function, i.e.,~$V^\theta(x,s)\approx v^\top\phi(x,s)$, in our actor-critic algorithms (see Section~\ref{subsec:grad-theta} and Algorithm~\ref{alg:AC}), the SPSA estimate of the sub-gradient would be of the form $g(\nu)\approx\lambda+v^\top\big[\phi(x^0,\nu^+)-\phi(x^0,\nu^-)\big]/2\Delta$ (see \eqref{nu_up_incre_SPSA} in Algorithm~\ref{alg:AC}).

\vspace{-0.1in}
\subsection{An Alternative Approach to Compute the Gradients}
\label{subsec:grad-alter}
\vspace{-0.05in}

In this section, we present an alternative way to compute the gradients, especially those w.r.t.~$\theta$ and $\lambda$. This allows us to estimate the gradient w.r.t.~$\lambda$ in a (more) incremental fashion (compared to the method of Section~\ref{subsec:grad-lambda}), with the cost of the need to use two different linear function approximators (instead of one used in Algorithm~\ref{alg:AC}). In this approach, we define the augmented MDP slightly different than the one in Section~\ref{subsec:grad-lambda}. The only difference is in the definition of the cost function, which is defined here as (note that $C(x,a)$ has been replaced by $0$ and $\lambda$ has been removed)

\vspace{-0.15in}
\begin{small}
\begin{equation*}
\bar{C}(x,s,a)=\left\{\begin{array}{cc}
(-s)^+/ (1-\alpha)& \text{if $x=x_T$,}\\
0 & \text{otherwise,}
\end{array}\right.
\end{equation*}
\end{small}
\vspace{-0.15in}

where $x_T$ is any terminal state of the original MDP $\mathcal{M}$. It is easy to see that the term $\frac{1}{(1-\alpha)}\E\Big[\big(D^\theta(x^0)- \nu\big)^+\Big]$ appearing in the gradients of \eqref{eq:grad-theta}-\eqref{eq:grad-lambda} is the value function of the policy $\theta$ at state $(x^0,\nu)$ in this augmented MDP. As a result, we have 

\noindent
{\bf Gradient w.r.t.~$\theta$:} It is easy to see that now this gradient \eqref{eq:grad-theta} is the gradient of the value function of the original MDP, $\nabla_\theta V^\theta(x^0)$, plus $\lambda$ times the gradient of the value function of the augmented MDP, $\nabla_\theta V^\theta(x^0,\nu)$, both at the initial states of these MDPs (with abuse of notation, we use $V$ for the value function of both MDPs). Thus, using linear approximators $u^\top f(x,s)$ and $v^\top\phi(x,s)$ for the value functions of the original and augmented MDPs, $\nabla_\theta L(\theta,\nu,\lambda)$ can be estimated as $\nabla_\theta\log\mu(a_k|x_k,s_k;\theta)\cdot(\epsilon_k+\lambda\delta_k)$, where $\epsilon_k$ and $\delta_k$ are the TD-errors of these MDPs. 

\noindent 
{\bf Gradient w.r.t.~$\lambda$:} Similar to the case for $\theta$, it is easy to see that this gradient \eqref{eq:grad-lambda} is $\nu-\beta$ plus the value function of the augmented MDP, $V^\theta(x^0,\nu)$, and thus, can be estimated {\em incrementally} as $\nabla_\lambda L(\theta,\nu,\lambda)\approx\nu-\beta+v^\top\phi(x,s)$.

\noindent 
{\bf Sub-Gradient w.r.t.~$\nu$:} This sub-gradient \eqref{eq:grad-nu} is $\lambda$ times one plus the gradient w.r.t.~$\nu$ of the value function of the augmented MDP, $\nabla_\nu V^\theta(x^0,\nu)$, and thus using SPSA, can be estimated {\em incrementally} as $\lambda\big(1+\frac{v^\top\big[\phi(x^0,\nu^+)-\phi(x^0,\nu^-)\big]}{2\Delta}\big)
$. 

\noindent 
Algorithm~\ref{alg:AC-alt} in Appendix~\ref{subsec:alt-alg} contains the pseudo-code of the resulting algorithm.

%%%%%%%%%%%%%%%%%%%%%%%%%%%%%%%%%%%%%%%%%%%%%%%%%%%%%%%%%%%%%%
%%%%%%%%%%%%%%%%%%%%%%%%%%%%%%%%%%%%%%%%%%%%%%%%%%%%%%%%%%%%%%
%%%%%%%%%%%%%%%%%%%%%%%%%%%%%%%%%%%%%%%%%%%%%%%%%%%%%%%%%%%%%%

\vspace{-0.1in}
\subsection{Actor-Critic Algorithms}
\label{sec:AC-alg}
\vspace{-0.05in}

In this section, we present two actor-critic algorithms for optimizing the risk-sensitive measure~\eqref{eq:unconstrained-discounted-risk-measure}. These algorithms are based on the gradient estimates of Sections~\ref{subsec:grad-theta}-\ref{subsec:grad-nu}. While the first algorithm (SPSA-based) is fully incremental and updates all the parameters $\theta,\nu,\lambda$ at each time-step, the second one updates $\theta$ at each time-step and updates $\nu$ and $\lambda$ only at the end of each trajectory, thus given the name semi trajectory-based. Algorithm~\ref{alg:AC} contains the pseudo-code of these algorithms. The projection operators $\Gamma_\Theta$, $\Gamma_N$, and $\Gamma_\Lambda$ are defined as in Section~\ref{sec:PG-alg} and are necessary to ensure the convergence of the algorithms. The step-size schedules satisfy the standard conditions for stochastic approximation algorithms, and ensures that the critic update is on the fastest time-scale $\big\{\zeta_4(k)\big\}$, the policy and VaR parameter updates are on the intermediate time-scale, with $\nu$-update $\big\{\zeta_3(k)\big\}$ being faster than $\theta$-update $\big\{\zeta_2(k)\big\}$, and finally the Lagrange multiplier update is on the slowest time-scale $\big\{\zeta_1(k)\big\}$ (see Appendix~\ref{subsec:ass2} for the conditions on these step-size schedules). This results in four time-scale stochastic approximation algorithms. We prove that these actor-critic algorithms converge to a (local) saddle point of the risk-sensitive objective function $L(\theta,\nu,\lambda)$ (see Appendix~\ref{subsec:convergence-proof-AC}).

%Our proposed actor critic algorithm uses a 4-timescale stochastic approximation on the value functions and gradients of the objective function. Similar to the simulated trajectory based algorithm, the proof of asymptotic convergence to a local minimum point of the mean-CVaR risk sensitive 
%objective function $ L(\theta,\nu,\lambda)$ will be established using the gradient flow approach. Consider the four-timescale policy gradient algorithm in Algorithm \ref{alg}. Based on the arguments in stochastic approximation, one can prove that the solutions of Algorithm \ref{alg} converge to the saddle point of the Lagrangian function if the linear function approximation of $V^\theta$ such that $|\tilde L_{v^*}(\theta,\nu, \lambda)-\tilde L_v(\theta,\nu,\lambda)|$ is sufficiently small when $\theta\rightarrow\theta^*$ and $\lambda\rightarrow\lambda^*$. 

\vspace{-0.1in}
\begin{algorithm}
\begin{small}
\begin{algorithmic}
\STATE {\bf Input:} Parameterized policy $\mu(\cdot|\cdot;\theta)$ and value function feature vector $\phi(\cdot)$ (both over the augmented MDP $\bar{\mathcal{M}}$), confidence level $\alpha$, loss tolerance $\beta$ and Lagrangian threshold $\lambda_{\max}$
\STATE {\bf Initialization:} policy parameters $\theta=\theta_0$; VaR parameter $\nu=\nu_0$; Lagrangian parameter $\lambda=\lambda_0$; value function weight vector $v=v_0$ 
%\STATE {\bf Input:} Let $x^0\in\X$ be an initial state and $\alpha\in(0,1)$ be the given CVaR parameter. 
%\STATE 
\WHILE{1}
\STATE \textbf{// (1) SPSA-based Algorithm:}
\FOR{$k = 0,1,2,\ldots$}
%\STATE Simulate the augmented states and action $(x_k,s_k,a^t)$ with the current policy $\theta_k$ where $s_k=(s_{t-1}-C(x_{t-1},a_{t-1}))/\gamma$ and $s_0=\nu_k$. 
\STATE Draw action $\;a_k\sim\mu(\cdot|x_k,s_k;\theta_k)$; $\quad\quad\quad\quad\quad\quad\quad$ Observe cost $\;\bar{C}(x_k,s_k,a_k)$;
\STATE Observe next state $(x_{k+1},s_{k+1})\sim \bar{P}(\cdot|x_k,s_k,a_k)$; $\;$ \begin{footnotesize}{\em // note that $s_{k+1}=(s_k-C\big(x_k,a_k)\big)/\gamma\;$ (see Sec.~\ref{subsec:grad-theta})}\end{footnotesize}
\begin{align}
\textrm{\bf TD Error:} \quad & \delta_k(v_k) = \bar{C}(x_k,s_k,a_k) + \gamma v_k^\top\phi(x_{k+1},s_{k+1}) - v_k^\top\phi(x_k,s_k) \label{TD-calc} \\
\textrm{\bf Critic Update:} \quad & v_{k+1}=v_k+\zeta_4(k)\delta_k(v_k)\phi(x_k,s_k) \label{v_up_incre} \\
\textrm{{\bf Actor Updates:}}\quad &  \nu_{k+1} = \Gamma_N\left(\nu_k - \zeta_3(k)\Big(\lambda_k + \frac{v_k^\top\big[\phi\big(x^0,\nu_k+\Delta_k\big)- \phi(x^0,\nu_k-\Delta_k)\big]}{2\Delta_k}\Big)\right) \label{nu_up_incre_SPSA} \\
&\theta_{k+1} = \Gamma_\Theta\Big(\theta_k-\frac{\zeta_2(k)}{1-\gamma}\nabla_\theta\log\mu(a_k|x_k,s_k;\theta)\vert_{\theta=\theta_k}\cdot\delta_k(v_k)\Big) \label{theta_up_incre} \\
%\textrm{{\bf $\nu$ Update:}}\quad 
%&\text{{\bf TODO: Is this necessary here?}where $\Delta_k\geq 0$, $\Delta_k\rightarrow 0$ almost surely as $k\rightarrow\infty$, $\sum_{k=0}^\infty\mathbb E\left[\left({\zeta_3(k)}/{\Delta_k}\right)^2\right] <\infty$}\nonumber\\
%&\textbf{(SPSA iterates)}\nonumber\\
%\textrm{{\bf $\lambda$ Update:}}\quad
&\lambda_{k+1} = \Gamma_\Lambda\Big(\lambda_k + \zeta_1(k)\big(\nu_k - \beta + \frac{\gamma^k}{1-\alpha}\mathbf 1 \{x_k=x_T\}(-s_k)^+\big)\Big) \label{lambda_up_incre}
\end{align}
\ENDFOR

%\STATE
\STATE \textbf{// (2) Semi Trajectory-based Algorithm:}
\FOR{$i = 0,1,2,\ldots$}
\STATE Set $\;k=0$ and $\;(x_k,s_k)=(x^0,\nu_i)$
\WHILE{$x_k\neq x_T$} 
\STATE Draw action $\;a_k\sim\mu(\cdot|x_k,s_k;\theta_k)$; $\quad$ Observe $\;\bar{C}(x_k,s_k,a_k)\;$ and $\;(x_{k+1},s_{k+1})\sim \bar{P}(\cdot|x_k,s_k,a_k)$
%with the current policy $\theta_k$ where $s_k=(s_{t-1}-C(x_{t-1},a_{t-1}))/\gamma$ and $s_0=\nu_i$.
%\begin{small}
\STATE For fixed values of $\nu_i$ and $\lambda_i$, execute \eqref{TD-calc}-\eqref{v_up_incre} and \eqref{theta_up_incre} with $(\zeta_{4}(k),\zeta_{2}(k))$ replaced by $(\zeta_{4}(i),\zeta_{2}(i))$; $\quad\quad\quad\quad\quad\quad\quad\quad\quad\quad$ $k \leftarrow k + 1$;
%\STATE $k \leftarrow k + 1$ 
%\begin{align}
%\textrm{\bf TD(0) Critic Update and $\theta$ Update:} \quad & \text{Use equation \eqref{v_up_incre} and \eqref{theta_up_incre} respectively, with $\lambda=\lambda_i$}\nonumber
%\end{align}
\ENDWHILE {$\quad\quad\quad$ // \begin{footnotesize}{\em we reach a terminal state $(x_T,s_T)$ (end of the trajectory)}\end{footnotesize}}
\begin{align}
%\textrm{\bf Set}\quad &x^{T(i)}=x_T,\, s^{T(i)}=\left(s^{T(i)-1}-C(x^{T(i)-1},a^{T(i)-1})\right)/\gamma\nonumber\\
\textrm{{\bf $\nu$ Update:}}\quad&\nu_{i+1}=\Gamma_N\left(\nu_i-\zeta_2(i)\Big(\lambda_i-\frac{\lambda_i}{1-\alpha}\mathbf 1\big\{s_T\leq 0\big\}\Big)\right) \label{nu_up_incre_SPSA_semi} \\
\textrm{{\bf $\lambda$ Update:}}\quad&\lambda_{i+1}=\Gamma_\Lambda\Big(\lambda_i+\zeta_1(i) \big(\nu_i-\beta+\frac{\gamma^T}{(1-\alpha)}(-s_T)^+\big)\Big) \label{lambda_up_incre_semi}  
%\textrm{{\bf $\lambda$ Update:}}\quad&\lambda_{i+1}=\Gamma_\Lambda\Big(\lambda_i+\zeta_1(i) \big(\nu_i-\beta+\frac{1}{(1-\alpha)(1-\gamma)}(-s_T)^+\big)\Big) \label{lambda_up_incre_semi}  
\end{align}
\ENDFOR 
\IF{$\{\lambda_i\}$ converges to $\lambda_{\max}$}
\STATE{Set $\lambda_{\max}\leftarrow 2\lambda_{\max}$.}
\ELSE
\STATE {\bf return} policy and value function parameters $v,\nu,\theta,\lambda$ and {\bf break}
\ENDIF
\ENDWHILE
\end{algorithmic}
\end{small}
\caption{Actor-Critic Algorithm for CVaR Optimization}
\label{alg:AC}
\end{algorithm}

\section{Experimental Results}
%\subsection{An Optimal Stopping Problem}
\vspace{-0.075in}

We consider an optimal stopping problem in which the state at each time step $k \leq T$ consists of the cost $c_k$ and time $k$, i.e.,~$x=(c_k,k)$, where $T$ is the stopping time. The agent (buyer) should decide either to accept the present cost or wait. If she accepts or when $k=T$, the system reaches a terminal state and the cost $c_k$ is received, otherwise, she receives the cost $p_h$ and the new state is $(c_{k+1},k+1)$, where $c_{k+1}$ is $f_uc_k$ w.p.~$p$ and $f_dc_k$ w.p.~$1-p$ ($f_u>1$ and $f_d<1$ are constants). Moreover, there is a discounted factor $\gamma\in(0,1)$ to account for the increase in the buyer's affordability. The problem has been described in more details in Appendix~\ref{sec:appendix_experiment}. Note that if we change cost to reward and minimization to maximization, this is exactly the American option pricing problem, a standard testbed to evaluate risk-sensitive algorithms (e.g.,~\cite{tamar2012policy}). Since the state space is continuous, solving for an exact solution via DP is infeasible, and thus, it requires approximation and sampling techniques.

\begin{figure*}[t]
\centering
\includegraphics[width=1.5in,angle=0]{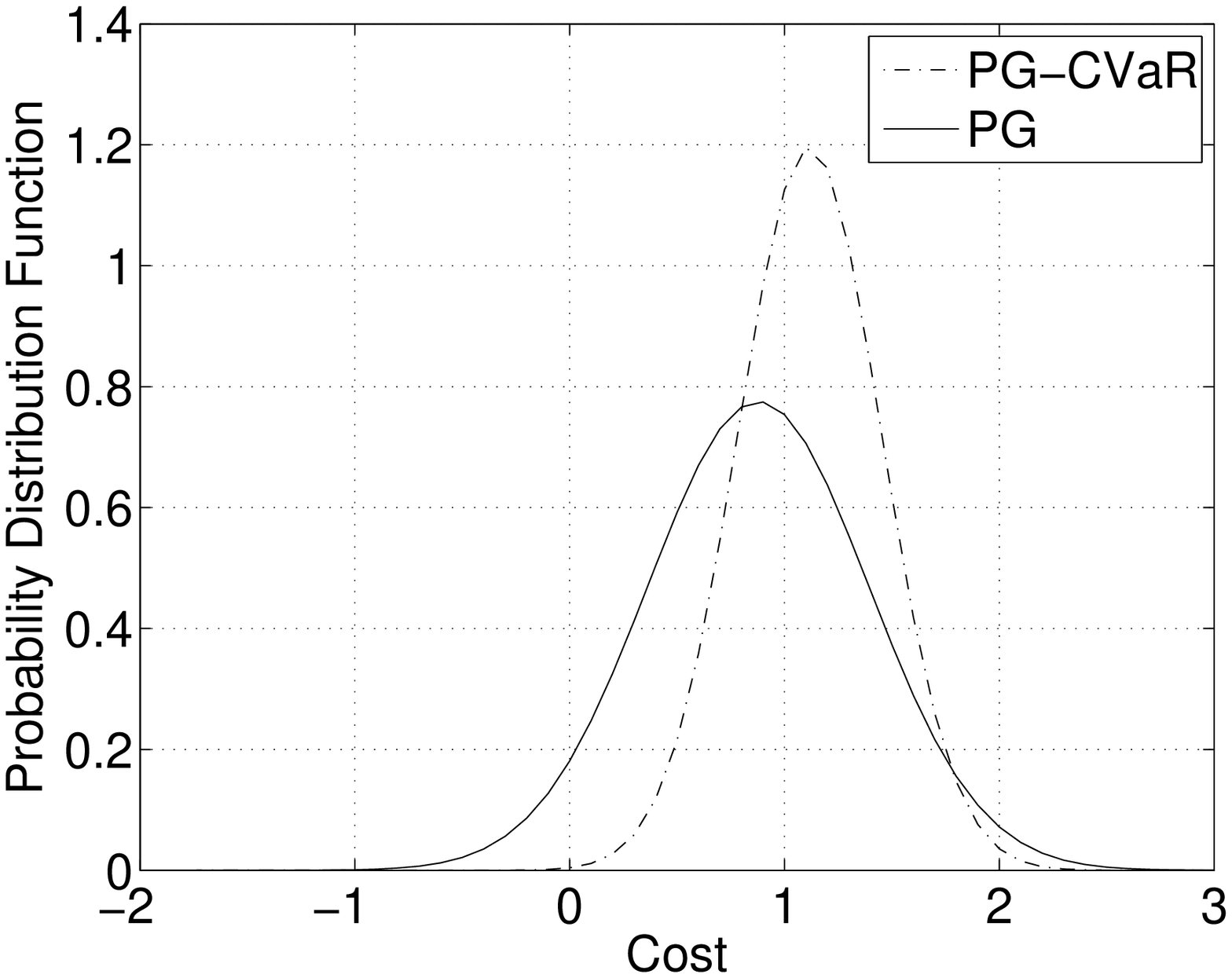}
\includegraphics[width=1.5in,angle=0]{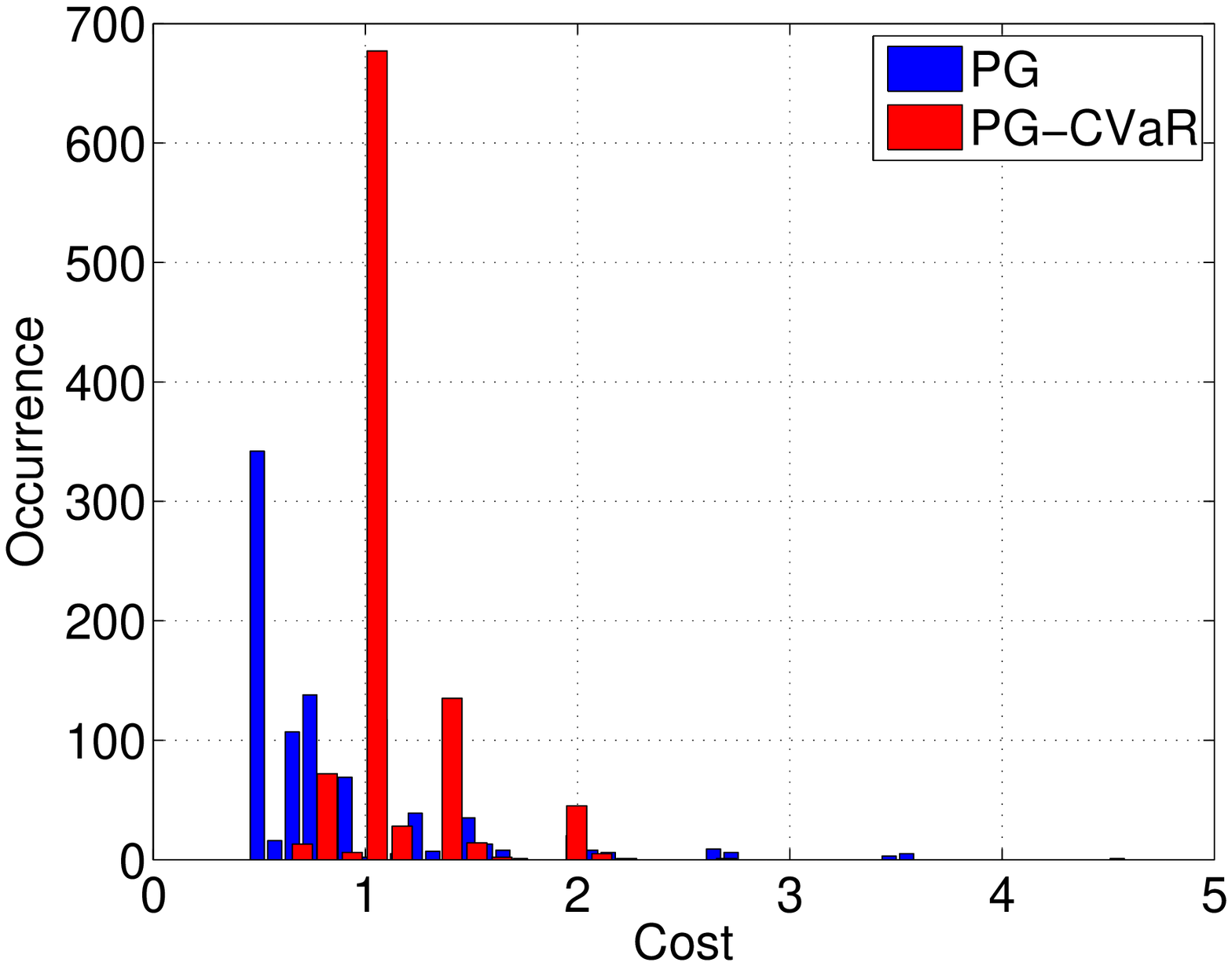}
\includegraphics[width=1.5in,angle=0]{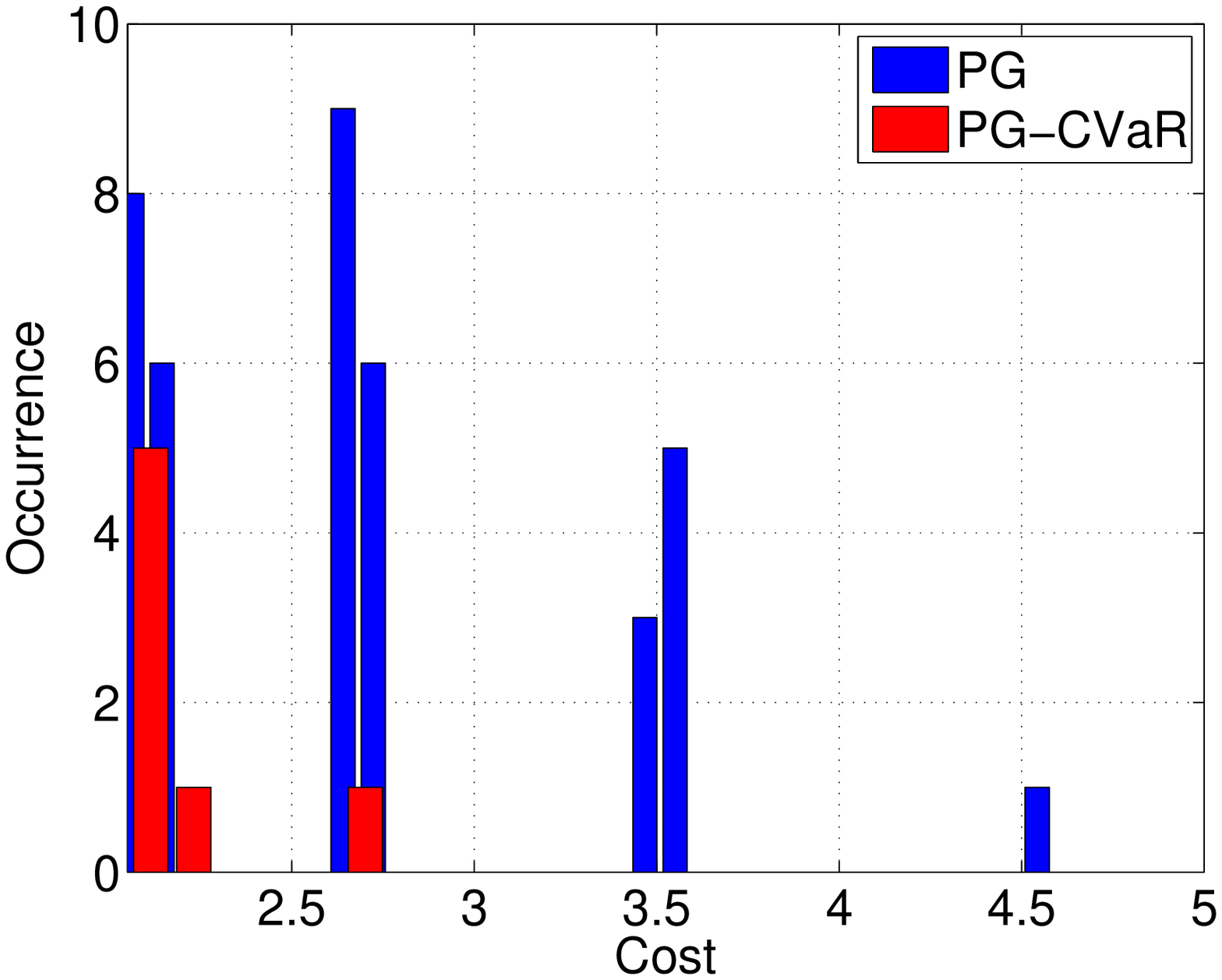}\\
\includegraphics[width=1.5in,angle=0]{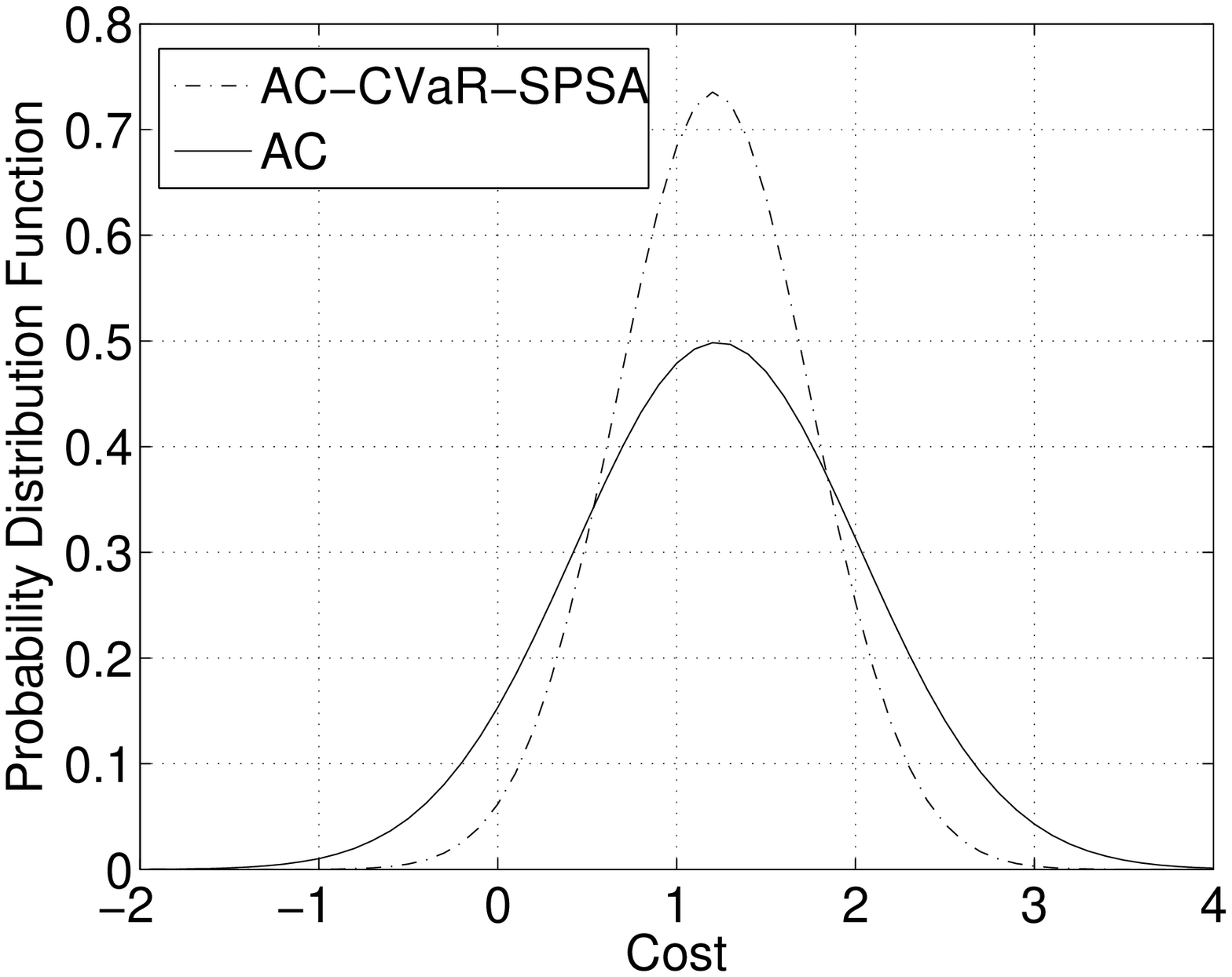}
\includegraphics[width=1.5in,angle=0]{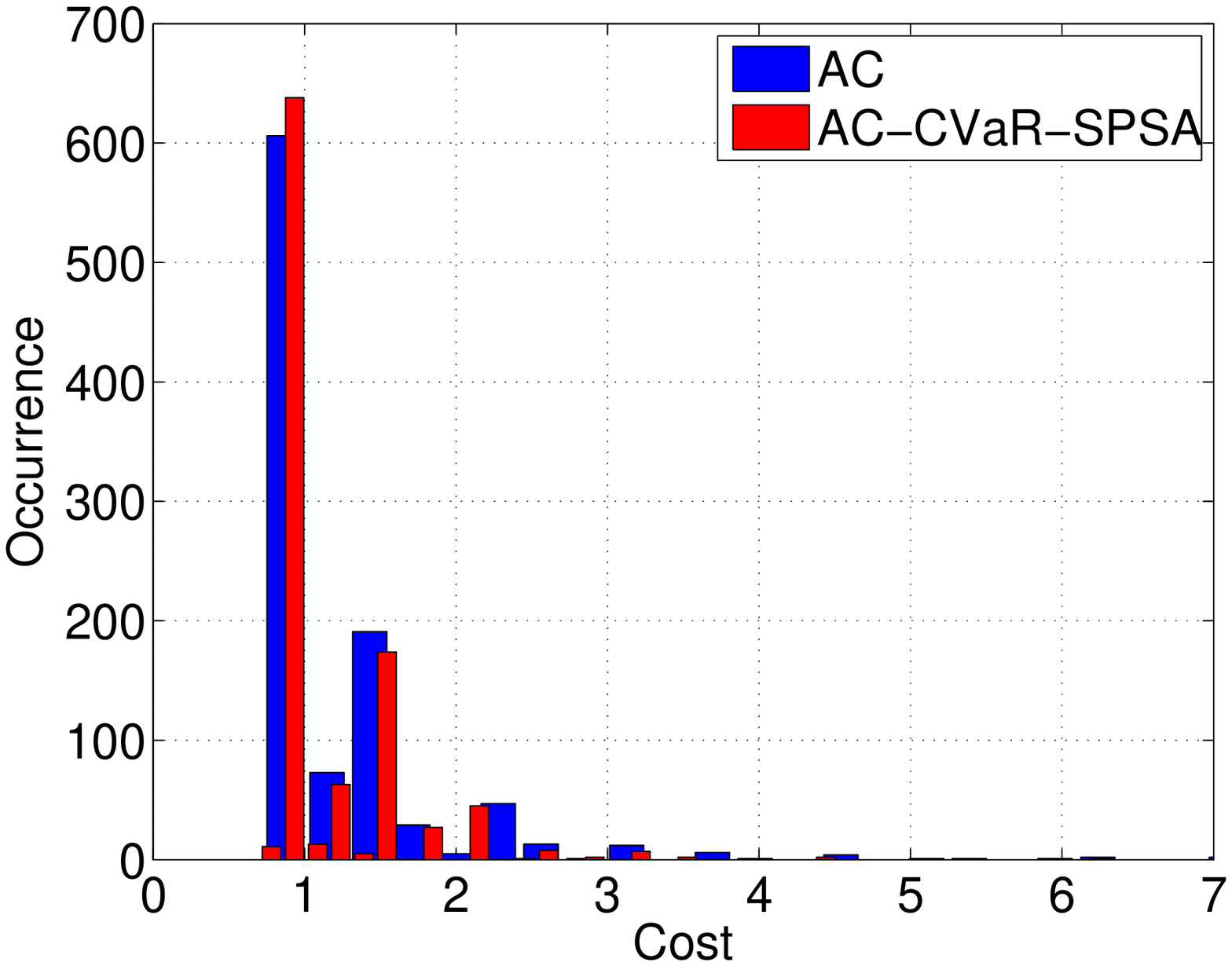}
\includegraphics[width=1.5in,angle=0]{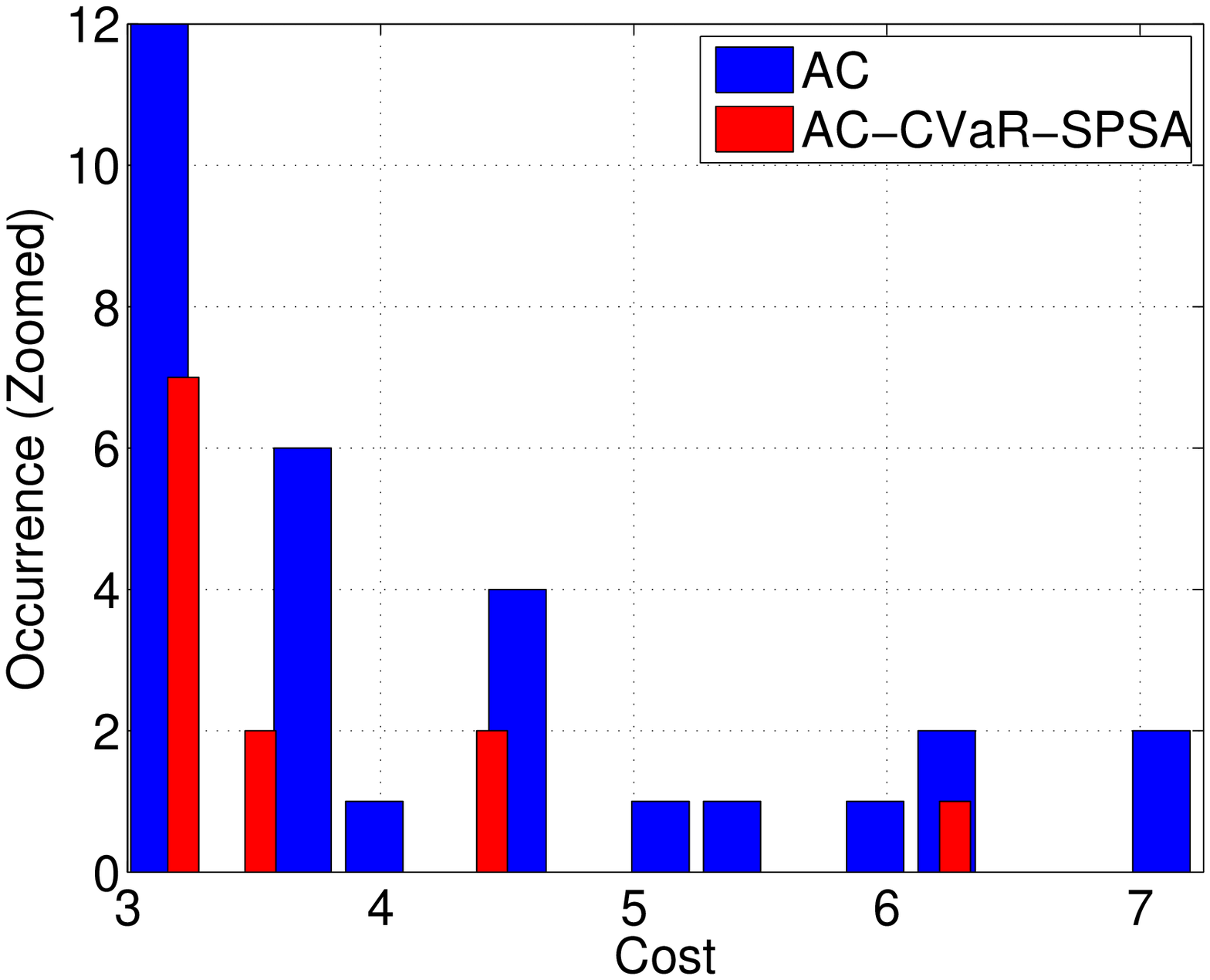}\\
\includegraphics[width=1.5in,angle=0]{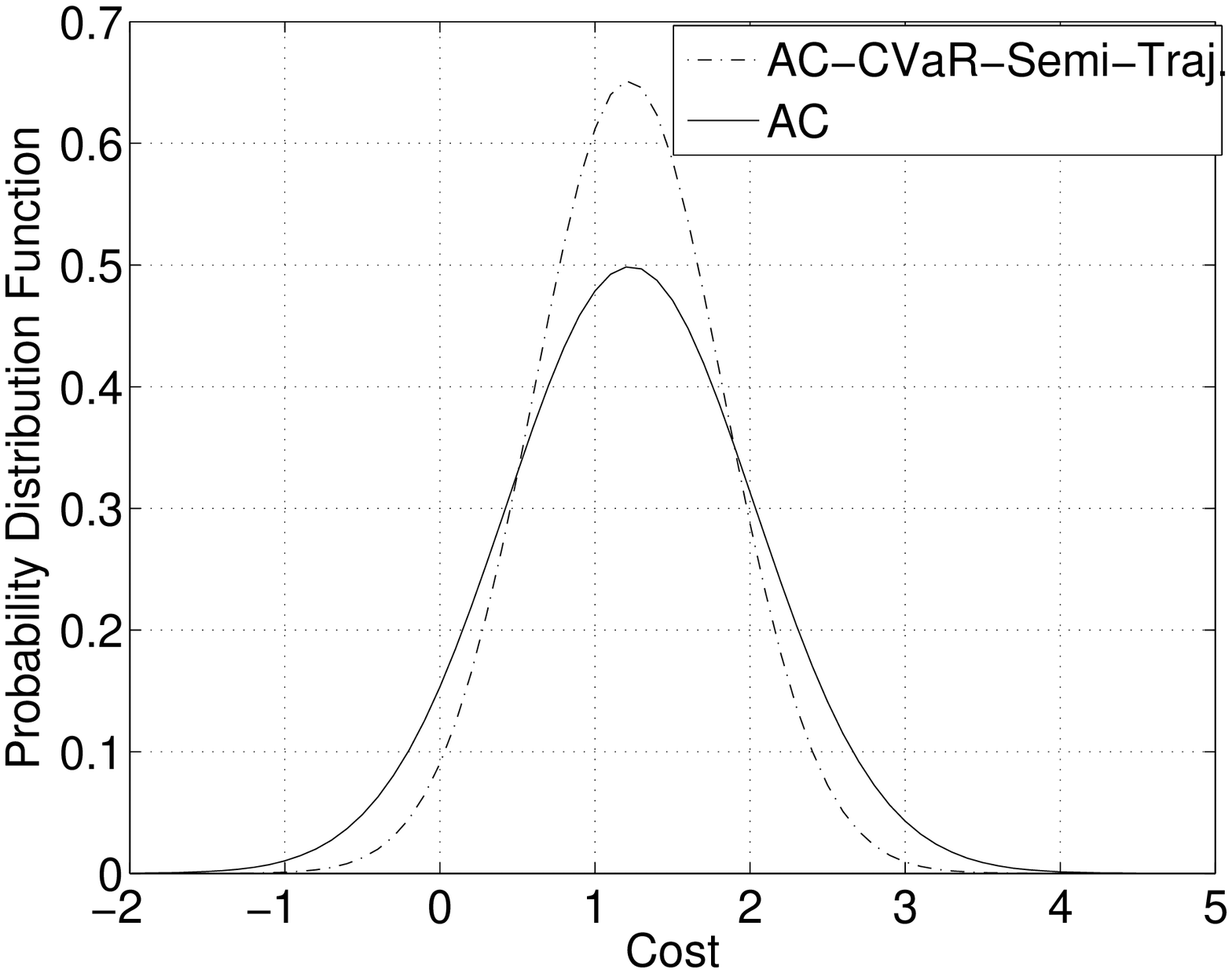}
\includegraphics[width=1.5in,angle=0]{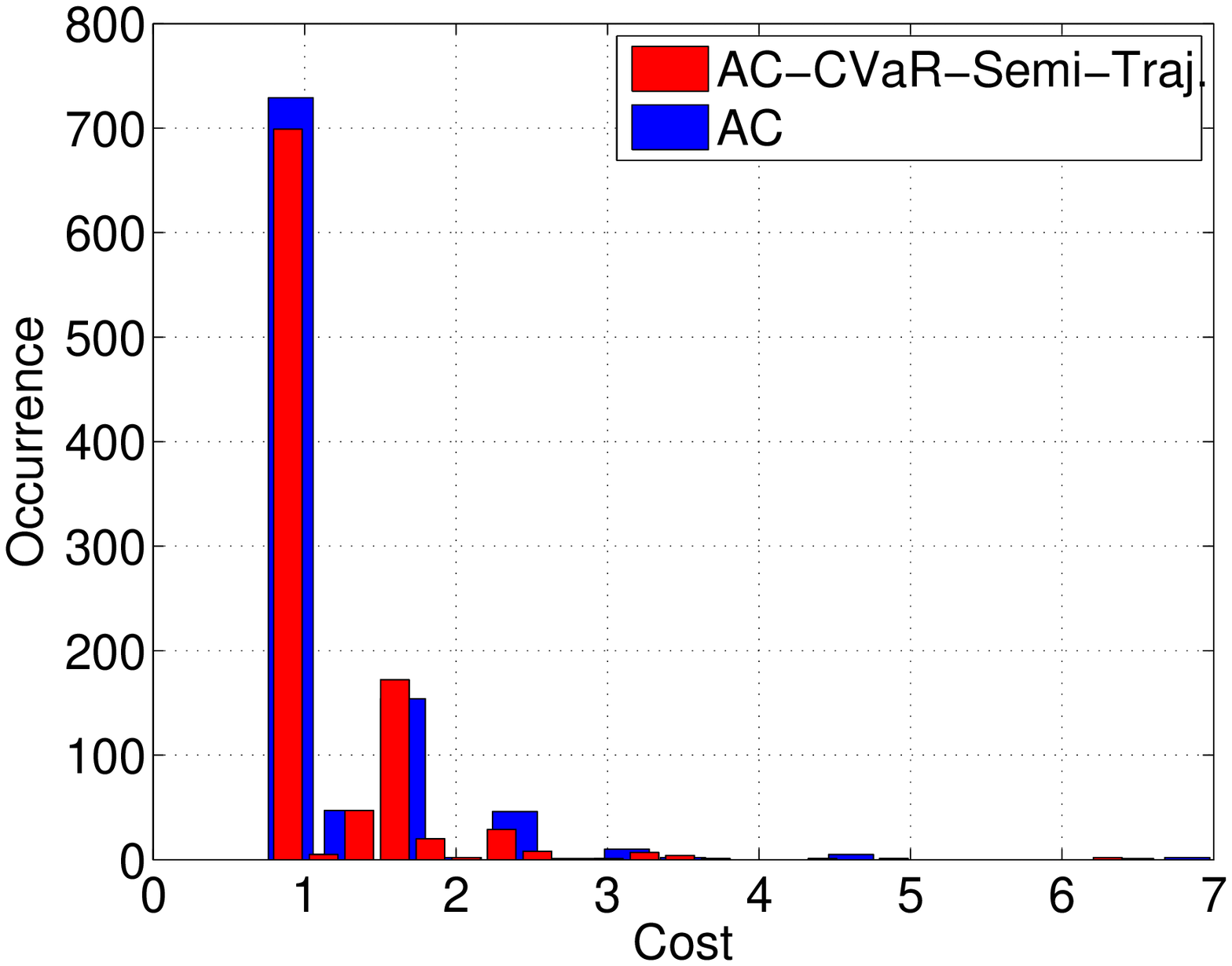}\includegraphics[width=1.5in,angle=0]{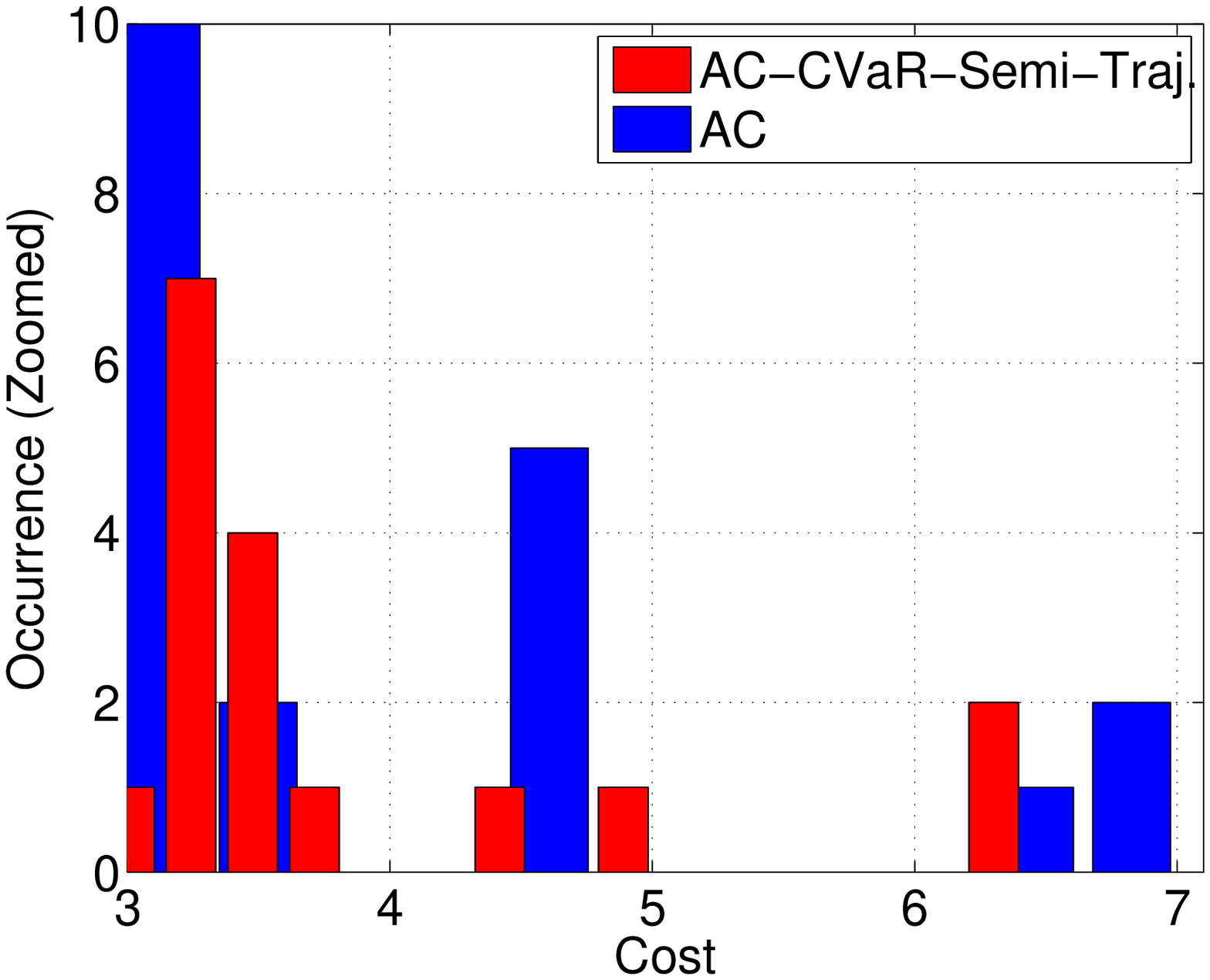}
%     \caption{Performance Comparison of a plain AC algorithm vs. RSAC algorithm using the average reward $\rho$ distribution}
\vspace{-0.1in}
\caption{Loss distributions for the policies learned by the risk-sensitive and risk-neutral algorithms.
%Performance comparison  using the distribution of $D^\theta(x^0)$ for policy gradient algorithms. 
% in both average as well as discounted settings.
}\label{fig:discounted_perf_traj}
\vspace{-0.025in}
\end{figure*}

We compare the performance of our risk-sensitive policy gradient Alg.~\ref{alg_traj} \begin{small}(PG-CVaR)\end{small} and two actor-critic Algs.~\ref{alg:AC} \begin{small}(AC-CVaR-SPSA,AC-CVaR-Semi-Traj)\end{small} with their risk-neutral counterparts \begin{small}(PG and AC)\end{small} (see Appendix~\ref{sec:appendix_experiment} for the details of these experiments). Fig.~\ref{fig:discounted_perf_traj} shows the distribution of the discounted cumulative cost $D^\theta(x^0)$ for the policy $\theta$ learned by each of these algorithms. From left to right, the columns display the first two moments, the whole (distribution), and zoom on the right-tail of these distributions. The results indicate that the risk-sensitive algorithms yield a higher expected loss, but less variance, compared to the risk-neutral methods. More precisely, the loss distributions of the risk-sensitive algorithms have lower right-tail than their risk-neutral counterparts. Table~\ref{tab:discounted_perf} summarizes the performance of these algorithms. The numbers reiterate what we concluded from Fig.~\ref{fig:discounted_perf_traj}.

%From these plots, we notice that the proposed risk-constrained policy gradient and actor critic algorithms, yield a higher expected cost, yet control the CVaR risk metric beyond the threshold. Notice that the policy resulted from the risk-neutral algorithm is infeasible to our problem setup. We observe that the mean performance of our risk-constrained algorithms are not significantly worse than its risk-neutral counterparts. Nevertheless, the proposed algorithms outperform their risk neutral counterparts, in terms of satisfying the CVaR constraint and the ``over-budget" (Value-at-Risk) constraint.

%\subsection{Trajectory Based Algorithms}
%In this section, we have implemented policy gradient algorithms for both the risk neutral (denoted by \textbf{PG}) and CVaR risk constrained (denoted by \textbf{PG-CVaR}) cases. On the other hand, a risk neutral actor critic (denoted by \textbf{AC}) and two risk constrained actor critic algorithms (denoted by \textbf{AC-CVaR-Semi-Traj.} and \textbf{AC-CVaR-SPSA}) are implemented, where details of the risk constrained actor critic algorithms are followed from  Section \ref{sec:AC-alg}. Further details of implementation can be found in Appendix \ref{sec:appendix_experiment}.

%\begin{figure*}[t]
%\centering
%\includegraphics[width=1.5in,angle=0]{figure/incre_based_plot.eps}\includegraphics[width=1.5in,angle=0]{figure/incre_based_histogram.eps}\includegraphics[width=1.5in,angle=0]{figure/incre_based_histogram_zoomed.eps}\\
%\includegraphics[width=1.5in,angle=0]{figure/incre_based_plot_2.eps}\includegraphics[width=1.5in,angle=0]{figure/incre_based_histogram_2.eps}\includegraphics[width=1.5in,angle=0]{figure/incre_based_histogram_zoomed_2.eps}
%%     \caption{Performance Comparison of a plain AC algorithm vs. RSAC algorithm using the average reward $\rho$ distribution}
%\vspace{-0.1in}
%\caption{Performance comparison using the distribution of $D^\theta(x^0)$ for actor critic algorithms. 
%% in both average as well as discounted settings.
%}\label{fig:discounted_perf_incre}
%\end{figure*}

\begin{table*}[t]
\begin{small}
\centering
\begin{tabular}{ |c| c| c| c| c| }
  \hline
  &$\mathbb E(D^\theta(x^0))$ & $\sigma^2(D^\theta(x^0))$ & $\text{CVaR}(D^\theta(x^0))$ & $\mathbb P(D^\theta(x^0)\geq \beta)$\\
  \hline
    PG & 0.8780 & 0.2647 & 2.0855 & 0.058\\
  \hline
  PG-CVaR &1.1128  &0.1109&1.7620 &0.012  \\
  \hline
  AC & 1.1963 & 0.6399 & 2.6479 & 0.029\\
  %AC & 1.2921 & 0.5027 & 3.0154 & 0.043\\
  \hline
  AC-CVaR-SPSA & 1.2031 &0.2942& 2.3865& 0.031 \\
  %AC-CVaR-SPSA & 1.3377 &0.4420& 2.7881& 0.044 \\
  \hline
   AC-CVaR-Semi-Traj. & 1.2169 &0.3747& 2.3889 &0.026  \\
  %AC-CVaR-Semi-Traj. & 1.3073 &0.4407& 2.7345& 0.028 \\
  \hline
\end{tabular}
\vspace{-0.025in}
\caption{\small{Performance comparison for the policies learned by the risk-sensitive and risk-neutral algorithms.}}
%Performance comparison  using the empirical risk metrics of $D^\theta(x^0)$. 
% in both average as well as discounted settings.
\label{tab:discounted_perf}
\end{small}  
\vspace{-0.15in}
\end{table*}

%%%%%%%%%%%%%%%%%%%%%%%%%%%%%%%%%%%%%%%%%%%%%%%%%%%%%%%%%%%%%%
%%%%%%%%%%%%%%%%%%%%%%%%%%%%%%%%%%%%%%%%%%%%%%%%%%%%%%%%%%%%%%
%%%%%%%%%%%%%%%%%%%%%%%%%%%%%%%%%%%%%%%%%%%%%%%%%%%%%%%%%%%%%%
%%%%%%%%%%%%%%%%%%%%%%%%%%%%%%%%%%%%%%%%%%%%%%%%%%%%%%%%%%%%%%
%%%%%%%%%%%%%%%%%%%%%%%%%%%%%%%%%%%%%%%%%%%%%%%%%%%%%%%%%%%%%%

\vspace{-0.125in}
\section{Conclusions and Future Work}
\label{sec:conclusions}
\vspace{-0.1in}

We proposed novel policy gradient and actor critic (AC) algorithms for CVaR optimization in MDPs. We provided proofs of convergence (in the appendix) to locally risk-sensitive optimal policies for the proposed algorithms. Further, using an optimal stopping problem, we observed that our algorithms resulted in policies whose loss distributions have lower right-tail compared to their risk-neutral counterparts. This is extremely important for a risk averse decision-maker, especially if the right-tail contains catastrophic losses. Future work includes: {\bf 1)} Providing convergence proofs for our AC algorithms when the samples are generated by following the policy and not from its discounted visiting distribution (this can be wasteful in terms of samples), {\bf 2)} Here we established asymptotic limits for our algorithms. To the best of our knowledge, there are no convergence rate results available for multi-timescale stochastic approximation schemes, and hence, for AC algorithms. This is true even for the AC algorithms that do not incorporate any risk criterion. It would be an interesting research direction to obtain finite-time bounds on the quality of the solution obtained by these algorithms, {\bf 3)} Since interesting losses in the CVaR optimization problems are those that exceed the VaR, in order to compute more accurate estimates of the gradients, it is necessary to generate more samples in the right-tail of the loss distribution (events that are observed with a very low probability). Although importance sampling methods have been used to address this problem~\cite{Bardou09CV,Tamar14PG}, several issues, particularly related to the choice of the sampling distribution, have remained unsolved that are needed to be investigated, and finally, {\bf 4)} Evaluating our algorithms in more challenging problems.  

%%%%%%%%%%%%%%%%%%%%%%%%%%%%%%%%%%%%%%%%%%%%%%%%%%%%%%%%%%%%%%
%%%%%%%%%%%%%%%%%%%%%%%%%%%%%%%%%%%%%%%%%%%%%%%%%%%%%%%%%%%%%%
%%%%%%%%%%%%%%%%%%%%%%%%%%%%%%%%%%%%%%%%%%%%%%%%%%%%%%%%%%%%%%
%%%%%%%%%%%%%%%%%%%%%%%%%%%%%%%%%%%%%%%%%%%%%%%%%%%%%%%%%%%%%%
%%%%%%%%%%%%%%%%%%%%%%%%%%%%%%%%%%%%%%%%%%%%%%%%%%%%%%%%%%%%%%

\newpage 

\begin{small}
\bibliography{cvar-rl}
\bibliographystyle{plainnat}
\end{small}

%%%%%%%%%%%%%%%%%%%%%%%%%%%%%%%%%%%%%%%%%%%%%%%%%%%%%%%%%%%%%%
%%%%%%%%%%%%%%%%%%%%%%%%%%%%%%%%%%%%%%%%%%%%%%%%%%%%%%%%%%%%%%
%%%%%%%%%%%%%%%%%%%%%%%%%%%%%%%%%%%%%%%%%%%%%%%%%%%%%%%%%%%%%%
%%%%%%%%%%%%%%%%%%%%%%%%%%%%%%%%%%%%%%%%%%%%%%%%%%%%%%%%%%%%%%
%%%%%%%%%%%%%%%%%%%%%%%%%%%%%%%%%%%%%%%%%%%%%%%%%%%%%%%%%%%%%%

\newpage
\appendix
\section{Technical Details of the Trajectory-based Policy Gradient Algorithm}

%%%%%%%%%%%%%%%%%%%%%%%%%%%%%%%%%%%%%%%%%%%%%%%%%%%%%%%%%%%%%%
%%%%%%%%%%%%%%%%%%%%%%%%%%%%%%%%%%%%%%%%%%%%%%%%%%%%%%%%%%%%%%
%%%%%%%%%%%%%%%%%%%%%%%%%%%%%%%%%%%%%%%%%%%%%%%%%%%%%%%%%%%%%%

\subsection{Assumptions}
\label{subsec:ass}

We make the following assumptions for the step-size schedules in our algorithms: \\

\noindent
{\bf (A1)} {\em For any state-action pair $(x,a)$, $\mu(a|x;\theta)$ is continuously differentiable in $\theta$ and $\nabla_\theta\mu(a|x;\theta)$ is a Lipschitz function in $\theta$ for every $a\in\A$ and $x\in\X$.} \\

\noindent
{\bf (A2)} {\em The Markov chain induced by any policy $\theta$ is irreducible and aperiodic.} \\

\noindent
{\bf (A3)} {\em The step size schedules $\{\zeta_3(i)\}$, $\{\zeta_2(i)\}$, and $\{\zeta_1(i)\}$ satisfy} %($k$ is some positive constant)}

\begin{align}
\label{eq:step1}
&\sum_i \zeta_1(i) = \sum_i \zeta_2(i) = \sum_i \zeta_3(i) = \infty, \\
\label{eq:step2}
&\sum_i \zeta_1(i)^2,\;\;\;\sum_i \zeta_2(i)^2,\;\;\;\sum_i \zeta_3(i)^2<\infty, \\
\label{eq:step3}
&\zeta_1(i) = o\big(\zeta_2(i)\big), \;\;\; \zeta_2(i) = o\big(\zeta_3(i)\big).%, \;\;\; \zeta_4(k) = k\zeta_3(k).
\end{align}

\eqref{eq:step1} and~\eqref{eq:step2} are standard step-size conditions in stochastic approximation algorithms, and \eqref{eq:step3} indicates that the update corresponds to $\{\zeta_3(i)\}$ is on the fastest time-scale, the update corresponds to $\{\zeta_2(i)\}$ is on the intermediate time-scale, and the update corresponds to $\{\zeta_1(i)\}$ is on the slowest time-scale.

\subsection{Computing the Gradients}
\label{subsec:grad-comp}

{\bf i) $\nabla_\theta L(\theta,\nu,\lambda)$: Gradient of $L(\theta,\nu,\lambda)$ w.r.t.~$\theta$} 

By expanding the expectations in the definition of the objective function $L(\theta,\nu,\lambda)$ in \eqref{eq:unconstrained-discounted-risk-measure}, we obtain
\begin{equation*}
L(\theta,\nu,\lambda)=\sum_{\xi} \mathbb{P}_\theta(\xi)D(\xi) + \lambda\nu + \frac{\lambda}{1-\alpha}\sum_\xi \mathbb{P}_\theta(\xi) \big(D(\xi)-\nu\big)^+-\lambda\beta.
\end{equation*}
By taking gradient with respect to $\theta$, we have
\begin{equation*}
\nabla_\theta L(\theta,\nu,\lambda)=\sum_{\xi} \nabla_\theta\mathbb{P}_\theta(\xi)D(\xi) + \frac{\lambda}{1-\alpha}\sum_\xi \nabla_\theta\mathbb{P}_\theta(\xi) \big(D(\xi)-\nu\big)^+.
\end{equation*}
This gradient can rewritten as 
\begin{equation}\label{eq:L_theta}
\nabla_\theta L(\theta,\nu,\lambda)=\sum_{\xi} \mathbb{P}_\theta(\xi)\cdot\nabla_\theta\log\mathbb{P}_\theta(\xi)\left(D(\xi) + \frac{\lambda}{1-\alpha}\big(D(\xi)-\nu\big)\mathbf{1}\big\{D(\xi)\geq\nu\big\}\right),
\end{equation}
where 
\begin{equation*}
\begin{split}
\nabla_\theta\log\mathbb{P}_\theta(\xi)=&\nabla_\theta\left\{\sum_{k=0}^{T-1} \log P(x_{k+1}|x_k,a_k)+\log\mu(a_k|x_k;\theta)+\log \mathbf 1\{x_0=x^0\}\right\}\\
=& \sum_{k=0}^{T-1}\frac{1}{\mu(a_k|x_k;\theta)}\nabla_\theta\mu(a_k|x_k;\theta)\\
=&\sum_{k=0}^{T-1}\nabla_\theta\log\mu(a_k|x_k;\theta).
\end{split}
\end{equation*}

{\bf ii) $\partial_\nu L(\theta,\nu,\lambda)$: Sub-differential of $L(\theta,\nu,\lambda)$ w.r.t.~$\nu$}

From the definition of $L(\theta,\nu,\lambda)$, we can easily see that $L(\theta,\nu,\lambda)$ is a convex function in $\nu$ for any fixed $\theta\in \Theta$. Note that for every fixed $\nu$ and any $\nu^\prime$, we have
\begin{equation*}
\big(D(\xi)-\nu'\big)^+ - \big(D(\xi)-\nu\big)^+ \geq g\cdot(\nu'-\nu),
\end{equation*}
where $g$ is any element in the set of sub-derivatives: 
\begin{equation*}
g\in\partial_\nu\big(D(\xi)-\nu\big)^+ \stackrel{\triangle}{=}
\begin{cases}
-1 & \text{if $\nu< D(\xi)$}, \\
-q:q\in[0,1] & \text{if $\nu=D(\xi)$}, \\
0 & \text{otherwise}.
\end{cases}
\end{equation*} 
Since $L(\theta,\nu,\lambda)$ is finite-valued for any $\nu\in\reals$, by the additive rule of sub-derivatives, we have % and the results in~\cite{hiriart1995subdifferential}, we have
\begin{equation}\label{eq:L_nu}
\partial_\nu L(\theta,\nu,\lambda) = \left\{-\frac{\lambda}{1-\alpha}\sum_\xi\mathbb{P}_\theta(\xi)\mathbf{1}\big\{D(\xi)>\nu\big\} - \frac{\lambda q}{1-\alpha}\sum_\xi\mathbb{P}_\theta(\xi)\mathbf{1}\big\{D(\xi)=\nu\big\} + \lambda\mid q\in[0,1]\right\}.% \quad\quad q\in[0,1].
\end{equation}
In particular for $q=1$, we may write the sub-gradient of $L(\theta,\nu,\lambda)$ w.r.t.~$\nu$ as

\vspace{-0.1in}
\begin{small}
\begin{equation*}
\partial_\nu L(\theta,\nu,\lambda)\vert_{q=0}= \lambda - \frac{\lambda}{1-\alpha}\sum_\xi\mathbb{P}_\theta(\xi)\cdot\mathbf{1}\big\{D(\xi)\geq\nu\big\} \quad \text{or} \quad \lambda - \frac{\lambda}{1-\alpha}\sum_\xi\mathbb{P}_\theta(\xi)\cdot\mathbf{1}\big\{D(\xi)\geq\nu\big\} \in \partial_\nu L(\theta,\nu,\lambda).
\end{equation*}
\end{small}
\vspace{-0.15in}

{\bf iii) $\nabla_\lambda L(\theta,\nu,\lambda)$: Gradient of $L(\theta,\nu,\lambda)$ w.r.t.~$\lambda$}

Since $L(\theta,\nu,\lambda)$ is a linear function in $\lambda$, obviously one can express the gradient of $L(\theta,\nu,\lambda)$ w.r.t.~$\lambda$ as follows:
\begin{equation}\label{eq:L_lambda}
\nabla_\lambda L(\theta,\nu,\lambda) = \nu - \beta +\frac{1}{1-\alpha}\sum_{\xi}\mathbb{P}_\theta(\xi)\cdot\big(D(\xi)- \nu\big)\mathbf{1}\big\{D(\xi)\geq\nu\big\}.
\end{equation}

%%%%%%%%%%%%%%%%%%%%%%%%%%%%%%%%%%%%%%%%%%%%%%%%%%%%%%%%%%%%%%
%%%%%%%%%%%%%%%%%%%%%%%%%%%%%%%%%%%%%%%%%%%%%%%%%%%%%%%%%%%%%%
%%%%%%%%%%%%%%%%%%%%%%%%%%%%%%%%%%%%%%%%%%%%%%%%%%%%%%%%%%%%%%

\subsection{Proof of Convergence of the Policy Gradient Algorithm}
\label{subsec:conv-proof}

In this section, we prove the convergence of our policy gradient algorithm (Algorithm~\ref{alg_traj}).

\begin{theorem}\label{thm:converge_h}
Suppose $\lambda^\ast\in[0,\lambda_{\max})$. Then the sequence of $(\theta,\lambda)-$updates in Algorithm \ref{alg_traj} converges to a (local) saddle point $(\theta^*,\nu^\ast,\lambda^*)$ of our objective function $L(\theta,\nu,\lambda)$ almost surely, i.e., it satisfies $L(\theta,\nu, \lambda^*) \ge L(\theta^*,\nu^\ast, \lambda^*) \ge L(\theta^*,\nu^\ast, \lambda),\forall(\theta,\nu)\in\Theta\times [-\frac{C_{\max}}{1-\gamma},\frac{C_{\max}}{1-\gamma}]\cap B_{(\theta^*,\nu^*)}(r)$ for some $r>0$ and $\forall \lambda\in [0,\lambda_{\max}]$. Note that $B_{(\theta^*,\nu^*)}(r)$ represents a hyper-dimensional ball centered at $(\theta^*,\nu^*)$ with radius $r$.
\end{theorem}

Since $\nu$ converges on the faster timescale than $\theta$ and $\lambda$, the $\nu$-update can be rewritten by assuming $(\theta,\lambda)$ as invariant quantities, i.e., 
\begin{equation}\label{nu_up_h_conv}
\nu_{i+1} = \Gamma_N\bigg[\nu_i - \zeta_3(i)\bigg(\lambda - \frac{\lambda}{(1-\alpha)N}\sum_{j=1}^N\mathbf{1}\big\{D(\xi_{j,i})\geq\nu_i\big\}\bigg)\bigg].
\end{equation}
Consider the continuous time dynamics of $\nu$ defined using differential inclusion
\begin{equation}\label{dyn_sys_nu_h}
\dot{\nu}\in \Upsilon_{\nu}\left[-g(\nu)\right], \quad\quad \forall g(\nu)\in\partial_\nu L(\theta,\nu,\lambda),
\end{equation}
where 
\[
\Upsilon_\nu[K(\nu)]:=\lim_{0<\eta\rightarrow 0} \frac{\Gamma_N(\nu+\eta K(\nu))-\Gamma_N(\nu)}{\eta}.
\]
and $\Gamma_N$ is the Euclidean projection operator of $\nu$ to $[-\frac{C_{\max}}{1-\gamma},\frac{C_{\max}}{1-\gamma}]$, i.e., $\Gamma_N(\nu)=\arg\min_{\hat\nu\in[-\frac{C_{\max}}{1-\gamma},\frac{C_{\max}}{1-\gamma}]}\frac{1}{2}\|\nu-\hat\nu\|^2_2$. In general $\Gamma_N(\nu)$ is not necessarily differentiable. $\Upsilon_\nu[K(\nu)]$ is the left directional derivative of the function $\Gamma_N(\nu)$ in the direction of $K(\nu)$. By using the left directional derivative $\Upsilon_\nu\left[ -g(\nu) \right]$ in the sub-gradient descent algorithm for $\nu$, the gradient will point at the descent direction along the boundary of $\nu$ whenever the $\nu-$update hits its boundary.

Furthermore, since $\nu$ converges on the faster timescale than $\theta$, and $\lambda$ is on the slowest time-scale, the $\theta$-update can be rewritten using the converged $\nu^*(\theta)$ and assuming $\lambda$ as an invariant quantity, i.e.,

\begin{equation}\label{theta_up_h_conv}
\begin{split}
\theta_{i+1}=&\Gamma_{\Theta}
\bigg[\theta_{i}-\zeta_2(i)\bigg(\frac{1}{N}\sum_{j=1}^N\nabla_\theta\log\mathbb{P}_\theta(\xi_{j,i})\vert_{\theta=\theta_i}D(\xi_{j,i}) \nonumber \\ 
&\hspace{0.725in}+ \frac{\lambda}{(1-\alpha)N}\sum_{j=1}^N\nabla_\theta\log\mathbb{P}_\theta(\xi_{j,i})\vert_{\theta=\theta_i}\big(D(\xi_{j,i})-\nu\big)\mathbf{1}\big\{D(\xi_{j,i})\geq\nu^*(\theta_i)\big\}\bigg)\bigg].
\end{split}\end{equation}
Consider the continuous time dynamics of $\theta\in \Theta$:
\begin{equation}\label{dyn_sys_kheta_h}
\dot{\theta}=\Upsilon_\theta\left[ -\nabla_\theta L(\theta,\nu,\lambda) \right]\vert_{\nu=\nu^*(\theta)},
\end{equation}
where
\[
\Upsilon_\theta[K(\theta)]:=\lim_{0<\eta\rightarrow 0} \frac{\Gamma_{\Theta}(\theta+\eta K(\theta))-\Gamma_\Theta(\theta)}{\eta}.
\]
and $\Gamma_\Theta$ is the Euclidean projection operator of $\theta$ to $\Theta$, i.e., $\Gamma_\Theta(\theta)=\arg\min_{\hat\theta\in\Theta}\frac{1}{2}\|\theta-\hat\theta\|^2_2$. Similar to the analysis of $\nu$, $\Upsilon_\theta[K(\theta)]$ is the left directional derivative of the function $\Gamma_\Theta(\theta)$ in the direction of $K(\theta)$. By using the left directional derivative $\Upsilon_\theta\left[ - \nabla_\theta L(\theta,\nu,\lambda)\right]$ in the gradient descent algorithm for $\theta$, the gradient will point at the descent direction along the boundary of $\Theta$ whenever the $\theta-$update hits its boundary.

Finally, since $\lambda$-update converges in a slowest time-scale, the $\lambda$-update can be rewritten using the converged $\theta^*(\lambda)$ and $\nu^*(\lambda)$, i.e.,
\begin{equation}\label{lamda_up_h_conv}
\lambda_{i+1}=\Gamma_\Lambda\left(\lambda_i+\zeta_1(i) \bigg(\nu^*(\lambda_i)+\frac{1}{1-\alpha} \frac{1}{N}\sum_{j=1}^N\big(D(\xi_{j,i})- \nu^*(\lambda_i)\big)^+ -\beta\bigg)\right).
\end{equation}
Consider the continuous time system
\begin{equation}\label{dyn_sys_lambda_h}
\dot{\lambda}(t)=\Upsilon_\lambda\left[ \nabla_\lambda L(\theta,\nu,\lambda)\bigg\vert_{\theta=\theta^*(\lambda),\nu=\nu^*(\lambda)}\right], \quad\quad \lambda(t)\geq 0,
\end{equation}
where
\[
\Upsilon_\lambda[K(\lambda)]:=\lim_{0<\eta\rightarrow 0} \frac{\Gamma_\Lambda\big(\lambda+\eta K(\lambda)\big)-\Gamma_\Lambda(\lambda)}{\eta}.
\]
and $\Gamma_\Lambda$ is the Euclidean projection operator of $\lambda$ to $[0,\lambda_{\max}]$, i.e., $\Gamma_\Lambda(\lambda)=\arg\min_{\hat\lambda\in[0,\lambda_{\max}]}\frac{1}{2}\|\lambda-\hat\lambda\|^2_2$. Similar to the analysis of $(\nu,\theta)$, $\Upsilon_\lambda[K(\lambda)]$ is the left directional derivative of the function $\Gamma_\Lambda(\lambda)$ in the direction of $K(\lambda)$. By using the left directional derivative $\Upsilon_\lambda\left[ \nabla_\lambda L(\theta,\nu,\lambda)\right]$ in the gradient ascent algorithm for $\lambda$, the gradient will point at the ascent direction along the boundary of $[0,\lambda_{\max}]$ whenever the $\lambda-$update hits its boundary.

Define
\[
L^*(\lambda)=L(\theta^\ast(\lambda),\nu^\ast(\lambda),\lambda),
\]
for $\lambda\geq 0$ where $(\theta^\ast(\lambda),\nu^\ast(\lambda))\in\Theta\times[-\frac{C_{\max}}{1-\gamma},\frac{C_{\max}}{1-\gamma}]$ is a local minimum of $L(\theta,\nu,\lambda)$ for fixed $\lambda\geq 0$, i.e., $L(\theta,\nu,\lambda)\geq L(\theta^\ast(\lambda),\nu^\ast(\lambda),\lambda)$ for any $(\theta,\nu)\in\Theta\times[-\frac{C_{\max}}{1-\gamma},\frac{C_{\max}}{1-\gamma}]\cap B_{(\theta^\ast(\lambda),\nu^\ast(\lambda))}(r)$ for some $r>0$.

Next, we want to show that the ODE \eqref{dyn_sys_lambda_h} is actually a gradient ascent of the Lagrangian function using the envelope theorem in mathematical economics \cite{milgrom2002envelope}. The envelope theorem describes sufficient conditions for the derivative of $L^*$ with respect to $\lambda$ where it equals to the partial derivative of the objective function $L$ with respect to $\lambda$, holding $(\theta,\nu)$ at its local optimum $(\theta,\nu)=(\theta^\ast(\lambda),\nu^\ast(\lambda))$.  We will show that  $ \nabla_\lambda L^\ast(\lambda)$ coincides with with $\nabla_\lambda L(\theta,\nu,\lambda)\vert_{\theta=\theta^\ast(\lambda),\nu=\nu^\ast(\lambda)}$ as follows.

\begin{theorem}\label{thm:envelop}
The value function $L^*$ is absolutely continuous. Furthermore,
\begin{equation}\label{envel_eq}
 L^*(\lambda)=L^*(0)+\int_{0}^{\lambda}\nabla_{\lambda'} L(\theta,\nu,\lambda')\Big\vert_{\theta=\theta^*(s),\nu=\nu^*(s),\lambda'=s}ds,\,\, \lambda\geq 0.  
 \end{equation}
 \end{theorem}
 \begin{prooff}
The proof follows from analogous arguments of Lemma 4.3 in \cite{borkar2005actor}.
From the definition of $L^*$, observe that for any $\lambda^{\prime
},\lambda^{\prime \prime }\geq 0$ with $\lambda^{\prime }<\lambda^{\prime \prime }$,
\[\small
\begin{split}
 |L^*(\lambda^{\prime \prime })-L^*(\lambda^{\prime })| \leq & \sup_{\theta\in\Theta,\nu\in[-\frac{C_{\max}}{1-\gamma},\frac{C_{\max}}{1-\gamma}]}|L(\theta,\nu,\lambda^{\prime\prime })-L(\theta,\nu,\lambda^{\prime })|\\ 
=&\sup_{\theta\in\Theta,\nu\in[-\frac{C_{\max}}{1-\gamma},\frac{C_{\max}}{1-\gamma}]}\left\vert \int_{\lambda^{\prime }}^{\lambda^{\prime \prime
}}\nabla_\lambda L(\theta,\nu,s)ds\right\vert \\
\leq& \int_{\lambda^{\prime }}^{\lambda^{\prime \prime
}}\sup_{\theta\in\Theta,\nu\in [\frac{-C_{\max}}{1-\gamma},\frac{C_{\max}}{1-\gamma}]}\left|\nabla_\lambda L(\theta,\nu,s)\right|ds\leq \frac{3 C_{\max}}{(1-\alpha)(1-\gamma)}(\lambda^{\prime\prime}-\lambda^\prime).
\end{split}
\]
This implies that $L^*$ is absolutely continuous. Therefore, $L^*$ is continuous everywhere and differentiable almost everywhere.

By the Milgrom-Segal envelope theorem of mathematical economics (Theorem 1 of \cite{milgrom2002envelope}), one can conclude that the derivative of $L^*(\lambda)$ coincides with the derivative of $L(\theta,\nu,\lambda)$ at the point of differentiability $\lambda$ and $\theta=\theta^*(\lambda)$, $\nu=\nu^*(\lambda)$. Also since $L^*$ is absolutely continuous, the limit of $ (L^*(\lambda)-L^*(\lambda^\prime))/(\lambda-\lambda^\prime)$ at $\lambda\uparrow \lambda^\prime$ (or $\lambda\downarrow \lambda^\prime$) coincides with the lower/upper directional derivatives if $\lambda^\prime$ is  a point of non-differentiability. Thus, there is only a countable number of non-differentiable points in $L^*$ and each point of non-differentiability has the same directional derivatives as the point slightly beneath (in the case of $\lambda\downarrow \lambda^\prime$) or above (in the case of $\lambda\uparrow \lambda^\prime$) it. As the set of non-differentiable points of $L^*$ has measure zero, it can then be interpreted that $\nabla_\lambda L^\ast(\lambda)$ coincides with $\nabla_\lambda L(\theta,\nu,\lambda)\vert_{\theta=\theta^*(\lambda),\nu=\nu^*(\lambda)}$, i.e., expression \eqref{envel_eq} holds. 
\end{prooff}
\begin{remark}
It can be easily shown that $L^*(\lambda)$ is a concave function. Since for given $\theta$ and $\nu$, $L(\theta,\nu,\lambda)$ is a linear function in $\lambda$. Therefore, for any $\alpha'\in[0,1]$, $\alpha' L^*(\lambda_1)+(1-\alpha') L^*(\lambda_2)\leq L^*(\alpha'\lambda_1+(1-\alpha')\lambda_2)$, i.e., $L^*(\lambda)$ is a concave function. Concavity of $L^*$ implies that it is continuous and directionally  (both left hand and right hand) differentiable in $\text{int dom}(L^*)$. Furthermore at any $\lambda=\tilde\lambda$ such that the derivative of $ L(\theta,\nu,\lambda)$ with respect of $\lambda$ at $\theta=\theta^*(\lambda), \nu=\nu^*(\lambda)$ exists, by Theorem 1 of \cite{milgrom2002envelope}, $\nabla_\lambda L^*(\lambda)\vert_{\lambda=\tilde\lambda_+}=(L^*(\tilde\lambda_+)-L^*(\tilde\lambda))/(\tilde\lambda_+-\tilde\lambda)\geq \nabla_\lambda L(\theta,\nu,\lambda)\vert_{\theta=\theta^*(\lambda),\nu=\nu^*(\lambda),\lambda=\tilde\lambda} \geq (L^*(\tilde\lambda_-)-L^*(\tilde\lambda))/(\tilde\lambda_--\tilde\lambda)=\nabla_\lambda L^*(\lambda)\vert_{\lambda=\tilde\lambda_-}$. Furthermore concavity of $L^*$ implies $\nabla_\lambda L^*(\lambda)\vert_{\lambda=\tilde\lambda_+}\leq \nabla_\lambda L^*(\lambda)\vert_{\lambda=\tilde\lambda_-}$. Combining these arguments, one obtains $\nabla_\lambda L^*(\lambda)\vert_{\lambda=\tilde\lambda_+}=\nabla_\lambda L(\theta,\nu,\lambda)\vert_{\theta=\theta^*(\lambda),\nu=\nu^*(\lambda),\lambda=\tilde\lambda}=\nabla_\lambda L^*(\lambda)\vert_{\lambda=\tilde\lambda_-}$.
\end{remark}

In order to prove the main convergence result, we need the following standard assumptions and remarks. 

\begin{assumption}\label{assume:semi_cont}
For any given $x^0\in\X$ and $\theta\in \Theta$, the set $\big\{\big(\nu,g(\nu)\big)\mid g(\nu)\in\partial_\nu L(\theta, \nu,\lambda)\big\}$ is closed.
\end{assumption}
\begin{remark}\label{remark:second_d_bounded}
For any given $\theta\in \Theta$, $\lambda\geq 0$, and $g(\nu)\in\partial_\nu L(\theta, \nu,\lambda)$, we have 
\begin{equation}
\label{eq:remark2}
|g(\nu)|\leq 3\lambda(1+|\nu|)/(1-\alpha). 
\end{equation}
To see this, recall from definition that $g$ can be parameterized by $q$ as, for $q\in[0,1]$,
\[
g(\nu)=-\frac{\lambda}{(1-\alpha)}\sum_{\xi}\mathbb P_\theta(\xi)\mathbf 1\left\{D(\xi)>\nu\right\}-\frac{\lambda q}{1-\alpha}\sum_{\xi}\mathbb P_\theta(\xi)\mathbf 1\left\{D(\xi) = \nu\right\}+\lambda.
\]
It is obvious that
$\left|\mathbf 1\left\{D(\xi) = \nu\right\}\right|,\left|\mathbf 1\left\{D(\xi)>\nu\right\}\right|\leq 1+|\nu|$. Thus, 
$\left|\sum_{\xi}\mathbb P_\theta(\xi)\mathbf 1\left\{D(\xi)>\nu\right\}\right|\leq \sup_{\xi}\left|\mathbf 1\left\{D(\xi)> \nu\right\}\right| \leq  1+|\nu|$, and
$\left|\sum_{\xi}\mathbb P_\theta(\xi)\mathbf 1\left\{D(\xi) = \nu\right\}\right| \leq  1+|\nu|$. Recalling $0<(1-q),\,(1-\alpha)<1$, these arguments imply the claim of \eqref{eq:remark2}.
\end{remark}
Before getting into the main result, we need the following technical proposition.
\begin{proposition}\label{L_lips}
$\nabla_\theta L(\theta,\nu,\lambda)$ is Lipschitz in $\theta$.
\end{proposition}
\begin{prooff}
Recall that 
\[
\nabla_\theta L(\theta,\nu,\lambda)=\sum_{\xi} \mathbb{P}_\theta(\xi)\cdot\nabla_\theta\log\mathbb{P}_\theta(\xi)\left(D(\xi) + \frac{\lambda}{1-\alpha}\big(D(\xi)-\nu\big)\mathbf{1}\big\{D(\xi)\geq\nu\big\}\right)
\]
and $\nabla_\theta\log\mathbb{P}_\theta(\xi)=\sum_{k=0}^{T-1}\nabla_\theta\mu(a_k|x_k;\theta)/\mu(a_k|x_k;\theta)$ whenever $\mu(a_k|x_k;\theta)\in ( 0,1]$. Now Assumption (A1) implies that $\nabla_\theta\mu(a_k|x_k;\theta)$ is a Lipschitz function in $\theta$ for any $a\in\A$ and $k\in\{0,\ldots,T-1\}$ and $\mu(a_k|x_k;\theta)$ is differentiable in $\theta$. Therefore, by recalling that $\mathbb{P}_\theta(\xi)=\prod_{k=0}^{T-1} P(x_{k+1}|x_k,a_k)\mu(a_k|x_k;\theta)\mathbf 1\{x_0=x^0\}$ and by combining these arguments and noting that the sum of products of Lipschitz functions is Lipschitz, one concludes that $\nabla_\theta L(\theta,\nu,\lambda)$ is Lipschitz in $\theta$.
\end{prooff}
\begin{remark}\label{remark_lip}
$\nabla_\theta L(\theta,\nu,\lambda)$ is Lipschitz in $\theta$ implies that $\|\nabla_\theta L(\theta,\nu,\lambda)\|^2\leq 2(\|\nabla_\theta L(\theta_0,\nu,\lambda)\|+\|\theta_0\|)^2+2\|\theta\|^2 $ which further implies that
\[
\|\nabla_\theta L(\theta,\nu,\lambda)\|^2\leq K_1(1+\|\theta\|^2).
\]
for $K_1=2\max(1,(\|\nabla_\theta L(\theta_0,\nu,\lambda)\|+\|\theta_0\|)^2)>0$. Similarly, $\nabla_\theta\log\mathbb{P}_{\theta}(\xi)$ is Lipschitz implies that
\[
\|\nabla_\theta\log\mathbb{P}_{\theta}(\xi)\|^2\leq K_2(\xi)(1+\|\theta\|^2).
\]
for a positive random variable $K_2(\xi)$. Furthermore, since $T<\infty $ w.p. $1$, $\mu(a_k|x_k;\theta)\in ( 0,1]$ and $\nabla_\theta\mu(a_k|x_k;\theta)$ is Lipschitz for any $k<T$, $K_2(\xi)<\infty$ w.p. $1$.
\end{remark}

We are now in a position to prove the convergence analysis of Theorem~\ref{thm:converge_h}. \\
\begin{prooff}[{\bf Proof of Theorem \ref{thm:converge_h}}]
We split the proof into the following four steps:

%%%%%%%%%%%%%%%%%%%%%%%%%%%%%%%%%%%%%%%%%%%%%%%%%%%%%%%%%%%%%%

\noindent\paragraph{Step~1 (Convergence of $\nu-$update)} 
Since $\nu$ converges in a faster time scale than $\theta$ and $\lambda$, one can assume both $\theta$ and $\lambda$ as fixed quantities in the $\nu$-update, i.e.,
\begin{equation}\label{update_s_h_2}
\nu_{i+1}=\Gamma_N\left(\nu_{i}+\zeta_3(i)\left(\frac{\lambda}{(1-\alpha)N}\sum_{j=1}^N\mathbf{1}\big\{D(\xi_{j,i})\geq \nu_i\big\}-\lambda+\delta \nu_{i+1}\right)\right),
\end{equation}
and 
\begin{equation}\label{eq:MG_diff_nu_h}
\delta \nu_{i+1}=\frac{\lambda}{1-\alpha}\left(-\frac{1}{N}\sum_{j=1}^N\mathbf{1}\big\{D(\xi_{j,i})\geq \nu_i\big\}+\sum_{\xi}\mathbb P_\theta(\xi) \mathbf 1\{D(\xi)\geq \nu_{i}\}\right).
\end{equation}
First, one can show that $\delta\nu_{i+1}$ is square integrable, i.e. 
\[
\mathbb E[ \|\delta\nu_{i+1}\|^2\mid F_{\nu,i}]\leq 4\left(\frac{\lambda_{\max}}{1-\alpha}\right)^2
\]
where $\mathcal F_{\nu,i}= \sigma\big(\nu_m,\,\delta \nu_m,\,m\leq i\big)$ is the filtration of $\nu_i$ generated by different independent trajectories. 

Second, since the history trajectories are generated based on the sampling probability mass function $\mathbb{P}_{\theta}(\xi)$, expression \eqref{eq:L_nu} implies that $\mathbb E\left[\delta\nu_{i+1}\mid \mathcal F_{\nu,i}\right]=0$.  Therefore, the $\nu$-update is a stochastic approximation of the ODE~\eqref{dyn_sys_nu_h} with a Martingale difference error term, i.e.,
\[
\frac{\lambda}{1-\alpha}\sum_{\xi}\mathbb P_\theta(\xi) \mathbf 1\{D(\xi)\geq \nu_{i}\}-\lambda\in-\partial_\nu L(\theta,\nu,\lambda)\vert_{\nu=\nu_{i}}.
\]

Then one can invoke Corollary~4 in Chapter~5 of~\cite{borkar2008stochastic} (stochastic approximation theory for non-differentiable systems) to show that the sequence $\{\nu_i\},\;\nu_i\in [-\frac{C_{\max}}{1-\gamma},\frac{C_{\max}}{1-\gamma}]$ converges almost surely to a fixed point $\nu^*\in [-\frac{C_{\max}}{1-\gamma},\frac{C_{\max}}{1-\gamma}]$ of differential inclusion~\eqref{dyn_sys_kheta_h}, where $\nu^\ast\in N_{c}:=\{\nu\in [-\frac{C_{\max}}{1-\gamma},\frac{C_{\max}}{1-\gamma}]:\Upsilon_\nu[-g(\nu)]=0,\, g(\nu)\in \partial_\nu L(\theta,\nu,\lambda)\}$. To justify the assumptions of this theorem, 1) from Remark~\ref{remark:second_d_bounded}, the Lipschitz property is satisfied, i.e., $\sup_{g(\nu)\in\partial_\nu L(\theta,\nu,\lambda)}|g(\nu)|\leq 3\lambda(1+|\nu|)/(1-\alpha)$, 2) $\partial_\nu L(\theta,\nu,\lambda)$ is a convex compact set by definition, 3) Assumption \ref{assume:semi_cont} implies that $\{(\nu,g(\nu))\mid g(\nu)\in\partial_\nu L(\theta,\nu,\lambda)\}$ is a closed set. This implies $\partial_\nu L(\theta,\nu,\lambda)$ is an upper semi-continuous set valued mapping 4) the step-size rule follows from~\eqref{subsec:ass}, 5) the Martingale difference assumption follows from~\eqref{eq:MG_diff_nu_h}, and 6) $\nu_i\in[-\frac{C_{\max}}{1-\gamma},\frac{C_{\max}}{1-\gamma}]$, $\forall i$ implies that $\sup_{i}\|\nu_i\|<\infty$ almost surely. 

Consider the ODE of $\nu\in \reals$ in~\eqref{dyn_sys_nu_h}, we define the set-valued derivative of $L$ as follows:
\[
D_t L(\theta,\nu,\lambda)=\big\{g(\nu)\Upsilon_\nu\big[-g(\nu)\big] \mid \forall g(\nu)\in \partial_\nu L(\theta,\nu,\lambda)\big\}.
\]
One may conclude that 
\[
\max_{g(\nu)}D_t L(\theta,\nu,\lambda)= \max\big\{g(\nu)\Upsilon_\nu\big[-g(\nu)\big] \mid g(\nu)\in \partial_\nu L(\theta,\nu,\lambda)\big\}.
\]

We have the following cases:

\noindent \emph{Case 1: When $\nu\in (-\frac{C_{\max}}{1-\gamma},\frac{C_{\max}}{1-\gamma})$.}\\
For every $g(\nu)\in \partial_\nu L(\theta,\nu,\lambda)$, there exists a sufficiently small $\eta_{0}>0$ such that $\nu-\eta_0g(\nu)\in [-\frac{C_{\max}}{1-\gamma},\frac{C_{\max}}{1-\gamma}]$ and 
\[
\Gamma_N\big(\theta-\eta_0g(\nu)\big)-\theta=-\eta_0g(\nu).
\]
Therefore, the definition of $\Upsilon_\theta[-g(\nu)]$ implies 
\begin{equation}\label{scen_1_h}
\max_{g(\nu)}D_t L(\theta,\nu,\lambda)=\max\big\{-g^2(\nu) \mid g(\nu)\in \partial_\nu L(\theta,\nu,\lambda)\big\}\leq 0.
\end{equation}
The maximum is attained because $\partial_\nu L(\theta,\nu,\lambda)$ is a convex compact set and $g(\nu)\Upsilon_\nu\big[-g(\nu)\big]$ is a continuous function.
At the same time, we have $\max_{g(\nu)}D_t L(\theta,\nu,\lambda)<0$ whenever $0\not\in \partial_\nu L(\theta,\nu,\lambda)$.

\noindent \emph{Case 2: When $\nu\in \{-\frac{C_{\max}}{1-\gamma},\frac{C_{\max}}{1-\gamma}\}$ and for any $g(\nu)\in \partial L_\nu(\theta,\nu,\lambda)$ such that $\nu-\eta g(\nu)\in [-\frac{C_{\max}}{1-\gamma},\frac{C_{\max}}{1-\gamma}]$, for any $\eta\in(0,\eta_0]$ and some $\eta_0>0$.}\\
The condition $\nu-\eta g(\nu)\in [-\frac{C_{\max}}{1-\gamma},\frac{C_{\max}}{1-\gamma}]$ implies that
\[
\Upsilon_\nu\big[-g(\nu)\big]=-g(\nu).
\]
Then we obtain
\begin{equation}\label{scen_2_h}
\max_{g(\nu)}D_t L(\theta,\nu,\lambda)=\max\big\{-g^2(\nu) \mid g(\nu)\in \partial_\nu L(\theta,\nu,\lambda)\big\}\leq 0.
\end{equation}
Furthermore, we have $\max_{g(\nu)}D_t L(\theta,\nu,\lambda)<0$ whenever $0\not\in \partial_\nu L(\theta,\nu,\lambda)$.
 
\noindent \emph{Case 3: When $\nu\in \{-\frac{C_{\max}}{1-\gamma},\frac{C_{\max}}{1-\gamma}\}$ and there exists a non-empty set 
$\mathcal G(\nu):=\{g(\nu)\in \partial L_\nu(\theta,\nu,\lambda)\mid \theta-\eta g(\nu)\not\in[-\frac{C_{\max}}{1-\gamma},\frac{C_{\max}}{1-\gamma}],\,\exists \eta\in(0,\eta_0],\,\forall \eta_0>0 \}$}.\\
First, consider any $g(\nu)\in \mathcal G(\nu)$. For any $\eta>0$, define $\nu_\eta:=\nu-\eta g(\nu)$. The above condition implies that when $0<\eta\rightarrow 0$, 
$\Gamma_N\big[\nu_\eta\big]$ is the projection of $\nu_\eta$ to the tangent space of $[-\frac{C_{\max}}{1-\gamma},\frac{C_{\max}}{1-\gamma}]$. For any elements $\hat\nu\in[-\frac{C_{\max}}{1-\gamma},\frac{C_{\max}}{1-\gamma}]$, since the following set
$\{\nu\in[-\frac{C_{\max}}{1-\gamma},\frac{C_{\max}}{1-\gamma}]:\|\nu-\nu_\eta\|_2\leq \|\hat\nu-\nu_\eta\|_2\}$ is compact, the projection of $\nu_\eta$ on $[-\frac{C_{\max}}{1-\gamma},\frac{C_{\max}}{1-\gamma}]$ exists. Furthermore, since $f(\nu):=\frac{1}{2}(\nu-\nu_\eta)^2$ is a strongly convex function and $\nabla f(\nu)=\nu-\nu_\eta$, by first order optimality condition, one obtains
\[
\nabla f(\nu_\eta^\ast)(\nu- \nu_\eta^\ast)=(\nu_\eta^\ast-\nu_\eta)(\nu- \nu_\eta^\ast) \geq 0, \quad \forall \nu \in \left[-\frac{C_{\max}}{1-\gamma},\frac{C_{\max}}{1-\gamma}\right]
\]
where $\nu_\eta^\ast$ is an unique projection of $\nu_\eta$ (the projection is unique because $f(\nu)$ is strongly convex and $[-\frac{C_{\max}}{1-\gamma},\frac{C_{\max}}{1-\gamma}]$ is a convex compact set). Since the projection (minimizer) is unique, the above equality holds if and only if $\nu=\nu_\eta^\ast$.

Therefore, for any $\nu\in[-\frac{C_{\max}}{1-\gamma},\frac{C_{\max}}{1-\gamma}]$ and $\eta>0$,
\[
\begin{split}
&g(\nu)\Upsilon_\nu\big[-g(\nu)\big]=g(\nu)\left(\lim_{0<\eta\rightarrow 0}\frac{\nu_\eta^\ast-\nu}{\eta}\right)\\
=&\left(\lim_{0<\eta\rightarrow 0}\frac{\nu-\nu_\eta}{\eta}\right)\left(\lim_{0<\eta\rightarrow 0}\frac{\nu_\eta^\ast-\nu}{\eta}\right)=\lim_{0<\eta\rightarrow 0}\frac{-\|\nu_\eta^\ast-\nu\|^2}{\eta^2}+
\lim_{0<\eta\rightarrow 0}\big(\nu_\eta^\ast-\nu_\eta\big)\left(\frac{\nu_\eta^\ast-\nu}{\eta^2}\right)\leq 0.
\end{split}
\]
Second, for any $g(\nu)\in \partial_\nu L(\theta,\nu,\lambda)\cap \mathcal G(\nu)^c$, one obtains $\nu-\eta g(\nu)\in [-\frac{C_{\max}}{1-\gamma},\frac{C_{\max}}{1-\gamma}]$, for any $\eta\in(0,\eta_0]$ and some $\eta_0>0$. In this case, the arguments follow from case 2 and the following expression holds, $\Upsilon_\nu\big[-g(\nu)\big]=-g(\nu)$.

Combining these arguments, one concludes that 
\begin{equation}\label{scen_3_h}
\small
\begin{split}
&\max_{g(\nu)}D_t L(\theta,\nu,\lambda)\\
\leq &\max\left\{\max\big\{g(\nu)\;\Upsilon_\nu\big[-g(\nu)\big] \mid g(\nu)\in \mathcal G(\nu)\big\},\max\big\{-g^2(\nu) \mid g(\nu)\in \partial_\nu L(\theta,\nu,\lambda)\cap \mathcal G(\nu)^c\big\}\right\}\leq 0.
\end{split}
\end{equation}
This quantity is non-zero whenever $0\not\in\{g(\nu)\;\Upsilon_\nu\big[-g(\nu)\big]\mid \forall g(\nu)\in \partial_\nu L(\theta,\nu,\lambda)\}$ (this is because, for any $g(\nu)\in \partial_\nu L(\theta,\nu,\lambda)\cap \mathcal G(\nu)^c$, one obtains $g(\nu)\;\Upsilon_\nu\big[-g(\nu)\big]=-g(\nu)^2$).

 Thus, by similar arguments one may conclude that $\max_{g(\nu)}D_t L(\theta,\nu,\lambda)\leq 0$ and it is non-zero if $\Upsilon_\nu\big[-g(\nu)\big]\neq 0$ for every $g(\nu)\in \partial_\nu L(\theta,\nu,\lambda)$. Therefore, by Lasalle's invariance principle for differential inclusion (see Theorem 2.11~\cite{ryan1998integral}), the above arguments imply that with any initial condition $\theta(0)$, the state trajectory $\nu(t)$ of \eqref{dyn_sys_kheta_h} converges to a stable stationary point $\nu^\ast$ in the positive invariant set $N_c$. Since $\max_{g(\nu)}D_t L(\theta,\nu,\lambda)\leq 0$, $\dot\nu$ is a descent direction of $L(\theta,\nu,\lambda)$ for fixed $\theta$ and $\lambda$, i.e., $L(\theta,\nu^\ast,\lambda)\leq L(\theta,\nu(t),\lambda)\leq L(\theta,\nu(0),\lambda)$ for any $t\geq 0$.

%%%%%%%%%%%%%%%%%%%%%%%%%%%%%%%%%%%%%%%%%%%%%%%%%%%%%%%%%%%%%%

\noindent\paragraph{Step~2 (Convergence of $\theta-$update)} 
Since $\theta$ converges in a faster time scale than $\lambda$ and $\nu$ converges faster than $\theta$, one can assume $\lambda$ as a fixed quantity and $\nu$ as a converged quantity  $\nu^*(\theta)$ in the $\theta$-update. The $\theta$-update can be rewritten as a stochastic approximation, i.e., 
\begin{equation}\label{update_theta_h_2}
\theta_{i+1}=\Gamma_{\Theta}\left(\theta_{i}+\zeta_2(i)\bigg(-\nabla_\theta L(\theta,\nu,\lambda)\vert_{\theta=\theta_i,\nu=\nu^*(\theta_i)}+\delta\theta_{i+1}\bigg)\right),
\end{equation}
where 
\begin{equation}\label{eq:MG_diff_theta_h}
\begin{split}
\delta\theta_{i+1}=&\nabla_\theta L(\theta,\nu,\lambda)\vert_{\theta=\theta_i,\nu=\nu^*(\theta_i)}\!-\!\frac{1}{N}\sum_{j=1}^N\nabla_\theta\log\mathbb{P}_{\theta}(\xi_{j,i})\mid_{\theta=\theta_i} D(\xi_{j,i})\\
&- \frac{\lambda}{(1-\alpha)N}\sum_{j=1}^N\nabla_\theta\log\mathbb{P}_\theta(\xi_{j,i})\vert_{\theta=\theta_i}\big(D(\xi_{j,i})-\nu^*(\theta_i)\big)\mathbf{1}\big\{D(\xi_{j,i})\geq\nu^*(\theta_i)\big\}.
\end{split}
\end{equation}

First, one can show that $\delta\theta_{i+1}$ is square integrable, i.e., $\mathbb E[ \|\delta\theta_{i+1}\|^2\mid F_{\theta,i}]\leq K_i(1+\|\theta_i\|^2)$ for some $K_i>0$, where $\mathcal F_{\theta,i}= \sigma\big(\theta_m,\,\delta \theta_m,\,m\leq i\big)$ is the filtration of $\theta_i$ generated by different independent trajectories. To see this, notice that
\[\small
\begin{split}
 \|\delta\theta_{i+1}\|^2\leq&  2 \left(\nabla_\theta L(\theta,\nu,\lambda)\vert_{\theta=\theta_i,\nu=\nu^*(\theta_i)}\right)^2+\frac{2}{N^2}\left(\frac{C_{\max}}{1-\gamma}+\frac{2\lambda C_{\max}}{(1-\alpha)(1-\gamma)}\right)^2\left(\sum_{j=1}^N\nabla_\theta\log\mathbb{P}_{\theta}(\xi_{j,i})\mid_{\theta=\theta_i}\right)^2\\
 \leq & 2 K_{1,i}(1+\|\theta_i\|^2)+\frac{2^{N}}{N^2}\left(\frac{C_{\max}}{1-\gamma}+\frac{2\lambda_{\max} C_{\max}}{(1-\alpha)(1-\gamma)}\right)^2 \left(\sum_{j=1}^N\left\|\nabla_\theta\log\mathbb{P}_{\theta}(\xi_{j,i})\mid_{\theta=\theta_i}\right\|^2\right)\\
 \leq & 2 K_{1,i}(1+\|\theta_i\|^2)+\frac{2^{N}}{N^2}\left(\frac{C_{\max}}{1-\gamma}+\frac{2\lambda_{\max} C_{\max}}{(1-\alpha)(1-\gamma)}\right)^2 \left(\sum_{j=1}^NK_2(\xi_{j,i})(1+\|\theta_i\|^2)\right)\\
 \leq & 2\!\left(K_{1,i}\!+\!\frac{2^{N-1}}{N}\left(\frac{C_{\max}}{1-\gamma}+\frac{2\lambda_{\max} C_{\max}}{(1-\alpha)(1-\gamma)}\right)^2\max_{1\leq j\leq N}K_2(\xi_{j,i})\right)\!(1\!+\!\|\theta_i\|^2)
 \end{split}
\]
The Lipschitz upper bounds are due to results in Remark \ref{remark_lip}. Since $K_2(\xi_{j,i})<\infty$ w.p. $1$, there exists $K_{2,i}<\infty$ such that $\max_{1\leq j\leq N}K_2(\xi_{j,i})\leq K_{2,i}$. Furthermore, $T<\infty$ w.p. $1$ implies $\mathbb E[T^2\mid \mathcal F_{\theta,i}]<\infty$. By combining these results, one concludes that
$\mathbb E[ \|\delta\theta_{i+1}\|^2\mid F_{\theta,i}]\leq K_i(1\!+\!\|\theta_i\|^2)$ where 
\[
K_i=2\left(K_{1,i}\!+\!\frac{2^{N-1}K_{2,i}}{N}\left(\frac{C_{\max}}{1-\gamma}+\frac{2\lambda_{\max} C_{\max}}{(1-\alpha)(1-\gamma)}\right)^2\right)<\infty.
\]

Second, since the history trajectories are generated based on the sampling probability mass function $\mathbb{P}_{\theta_i}(\xi)$, expression \eqref{eq:L_theta} implies that $\mathbb E\left[\delta\theta_{i+1}\mid \mathcal F_{\theta,i}\right]=0$.  Therefore, the $\theta$-update is a stochastic approximation of the ODE~\eqref{dyn_sys_kheta_h} with a Martingale difference error term. In addition, from the convergence analysis of $\nu-$update, $\nu^\ast(\theta)$ is an asymptotically stable equilibrium point of $\{\nu_i\}$. From \eqref{eq:L_nu}, $\partial_{\nu} L(\theta,\nu,\lambda)$ is a Lipschitz set-valued mapping in $\theta$ (since $\mathbb P_\theta(\xi)$ is Lipschitz in $\theta$), it can be easily seen that $\nu^\ast(\theta)$ is a Lipschitz continuous mapping of $\theta$.

Now consider the continuous time system $\theta\in \Theta$ in~\eqref{dyn_sys_kheta_h}. We may write
\begin{equation}\label{eq:lyap_ineq_1}
\frac{d L(\theta,\nu,\lambda)}{dt}\bigg\vert_{\nu=\nu^*(\theta)}=\big(\nabla_\theta L(\theta,\nu,\lambda)\vert_{\nu=\nu^*(\theta)}\big)^\top\;\Upsilon_\theta\big[-\nabla_\theta L(\theta,\nu,\lambda)\vert_{\nu=\nu^*(\theta)}\big].
\end{equation}
We have the following cases:

\noindent \emph{Case 1: When $\theta\in \Theta^\circ$.}\\
 Since $\Theta^\circ$ is the interior of the set $\Theta$ and $\Theta$ is a convex compact set, there exists a sufficiently small $\eta_{0}>0$ such that $\theta-\eta_0\nabla_\theta L(\theta,\nu,\lambda)\vert_{\nu=\nu^*(\theta)}\in \Theta$ and 
\[
\Gamma_\Theta\big(\theta-\eta_0\nabla_\theta L(\theta,\nu,\lambda)\vert_{\nu=\nu^*(\theta)}\big)-\theta=-\eta_0\nabla_\theta L(\theta,\nu,\lambda)\vert_{\nu=\nu^*(\theta)}.
\]
Therefore, the definition of $\Upsilon_\theta\big[-\nabla_\theta L(\theta,\nu,\lambda)\vert_{\nu=\nu^*(\theta)}\big]$ implies 
\begin{equation}\label{scen_1_h}
\frac{d L(\theta,\nu,\lambda)}{dt}\bigg\vert_{\nu=\nu^*(\theta)}=-\left\|\nabla_\theta L(\theta,\nu,\lambda)\vert_{\nu=\nu^*(\theta)}\right\|^2\leq 0.
\end{equation}
At the same time, we have ${d L(\theta,\nu,\lambda)}/{dt}\vert_{\nu=\nu^*(\theta)}<0$ whenever $\|\nabla_\theta L(\theta,\nu,\lambda)\vert_{\nu=\nu^*(\theta)}\|\neq 0$.

\noindent \emph{Case 2: When $\theta\in \partial\Theta$ and $\theta-\eta\nabla_\theta L(\theta,\nu,\lambda)\vert_{\nu=\nu^*(\theta)}\in \Theta$ for any $\eta\in(0,\eta_0]$ and some $\eta_0>0$.}\\
The condition $\theta-\eta\nabla_\theta L(\theta,\nu,\lambda)\vert_{\nu=\nu^*(\theta)}\in \Theta$ implies that
\[
\Upsilon_\theta\big[-\nabla_\theta L(\theta,\nu,\lambda)\vert_{\nu=\nu^*(\theta)}\big]=-\nabla_\theta L(\theta,\nu,\lambda)\vert_{\nu=\nu^*(\theta)}.
\]
Then we obtain
\begin{equation}\label{scen_1_h}
\begin{split}
\frac{d L(\theta,\nu,\lambda)}{dt}\bigg\vert_{\nu=\nu^*(\theta)}=-\left\|\nabla_\theta L(\theta,\nu,\lambda)\vert_{\nu=\nu^*(\theta)}\right\|^2\leq 0.
\end{split}
\end{equation}
Furthermore, ${d L(\theta,\nu,\lambda)}/{dt}\vert_{\nu=\nu^*(\theta)}<0$ when $\|\nabla_\theta L(\theta,\nu,\lambda)\vert_{\nu=\nu^*(\theta)}\|\neq 0$.

\noindent \emph{Case 3: When $\theta\in \partial\Theta$ and $\theta-\eta\nabla_\theta L(\theta,\nu,\lambda)\vert_{\nu=\nu^*(\theta)}\not\in \Theta$ for some $\eta\in(0,\eta_0]$ and any $\eta_0>0$.}\\
For any $\eta>0$, define $\theta_\eta:=\theta-\eta\nabla_\theta L(\theta,\nu,\lambda)\vert_{\nu=\nu^*(\theta)}$. The above condition implies that when $0<\eta\rightarrow 0$, 
$\Gamma_\Theta\big[\theta_\eta\big]$ is the projection of $\theta_\eta$ to the tangent space of $\Theta$. For any elements $\hat\theta\in\Theta$, since the following set
$\{\theta\in\Theta:\|\theta-\theta_\eta\|_2\leq \|\hat\theta-\theta_\eta\|_2\}$ is compact, the projection of $\theta_\eta$ on $\Theta$ exists. Furthermore, since $f(\theta):=\frac{1}{2}\|\theta-\theta_\eta\|_2^2$ is a strongly convex function and $\nabla f(\theta)=\theta-\theta_\eta$, by first order optimality condition, one obtains
\[
\nabla f(\theta_\eta^\ast)^\top(\theta- \theta_\eta^\ast)=(\theta_\eta^\ast-\theta_\eta)^\top(\theta- \theta_\eta^\ast) \geq 0, \quad \forall \theta \in \Theta
\]
where $\theta_\eta^\ast$ is an unique projection of $\theta_\eta$ (the projection is unique because $f(\theta)$ is strongly convex and $\Theta$ is a convex compact set). Since the projection (minimizer) is unique, the above equality holds if and only if $\theta=\theta_\eta^\ast$.

Therefore, for any $\theta\in\Theta$ and $\eta>0$,
\[
\begin{split}
&\big(\nabla_\theta L(\theta,\nu,\lambda)\vert_{\nu=\nu^*(\theta)}\big)^\top\;\Upsilon_\theta\big[-\nabla_\theta L(\theta,\nu,\lambda)\vert_{\nu=\nu^*(\theta)}\big]=\big(\nabla_\theta L(\theta,\nu,\lambda)\vert_{\nu=\nu^*(\theta)}\big)^\top\left(\lim_{0<\eta\rightarrow 0}\frac{\theta_\eta^\ast-\theta}{\eta}\right)\\
=&\left(\lim_{0<\eta\rightarrow 0}\frac{\theta-\theta_\eta}{\eta}\right)^\top\left(\lim_{0<\eta\rightarrow 0}\frac{\theta_\eta^\ast-\theta}{\eta}\right)=\lim_{0<\eta\rightarrow 0}\frac{-\|\theta_\eta^\ast-\theta\|^2}{\eta^2}+
\lim_{0<\eta\rightarrow 0}\big(\theta_\eta^\ast-\theta_\eta\big)^\top\left(\frac{\theta_\eta^\ast-\theta}{\eta^2}\right)\leq 0.
\end{split}
\]
From these arguments, one concludes that ${d L(\theta,\nu,\lambda)}/{dt}\vert_{\nu=\nu^*(\theta)}\leq 0$ and this quantity is non-zero whenever $\left\|\Upsilon_\theta\left[-\nabla_\theta L(\theta,\nu,\lambda)\vert_{\nu=\nu^*(\theta)}\right]\right\|\neq 0$. 

Therefore, by Lasalle's invariance principle \cite{khalil2002nonlinear}, the above arguments imply that with any initial condition $\theta(0)$, the state trajectory $\theta(t)$ of \eqref{dyn_sys_kheta_h} converges to a stable stationary point $\theta^\ast$ in the positive invariant set $\Theta_c$ and $L(\theta^\ast,\nu^\ast(\theta^\ast),\lambda)\leq L(\theta(t),\nu^\ast(\theta(t)),\lambda)\leq L(\theta(0),\nu^\ast(\theta(0)),\lambda)$ for any $t\geq 0$. 

Based on the above properties and noting that 1) from Proposition \ref{L_lips}, $\nabla_\theta L(\theta,\nu,\lambda)$ is a Lipschitz function in $\theta$, 2) the step-size rule follows from Section~\ref{subsec:ass}, 3) expression \eqref{eq:MG_diff_theta_h} implies that $\delta\theta_{i+1}$ is a square integrable Martingale difference, and 4) $\theta_i\in\Theta$, $\forall i$ implies that $\sup_{i}\|\theta_i\|<\infty$ almost surely,
one can invoke Theorem~2 in Chapter~6 of~\cite{borkar2008stochastic} (multi-time scale stochastic approximation theory) to show that the sequence $\{\theta_i\},\;\theta_i\in \Theta$ converges almost surely to a fixed point $\theta^*\in \Theta$ of ODE~\eqref{dyn_sys_kheta_h}, where $\theta^\ast\in\Theta_{c}:=\{\theta\in\Theta:\Upsilon_\theta[-\nabla_\theta L(\theta,\nu,\lambda)\vert_{\nu=\nu^*(\theta)}]=0\}$. Also, it can be easily seen that $\Theta_{c}$ is a closed subset of the compact set $\Theta$, which is a compact set as well.

%
%%%%%%%%%%%%%%%%%%%%%%%%%%%%%%%%%%%%%%%%%%%%%%%%%%%%%%%%%%%%%%
\noindent\paragraph{Step 3 (Local Minimum)}
Now, we want to show that $\{\theta_i,\nu_i\}$ converges to a local minimum of $L(\theta,\nu,\lambda)$ for fixed $\lambda$.
Recall $\{\theta_i,\nu_i\}$ converges to $(\theta^\ast,\nu^\ast):=(\theta^\ast,\nu^\ast(\theta^\ast))$. 
From previous arguments on $(\nu,\theta)$ convergence analysis imply that with any initial condition $(\theta(0),\nu(0))$, the state trajectories $\theta(t)$ and $\nu(t))$ of \eqref{dyn_sys_nu_h} and \eqref{dyn_sys_kheta_h} converge to the set of stationary points $(\theta^\ast,\nu^\ast)$ in the positive invariant set $\Theta_c\times N_c$ and $L(\theta^\ast,\nu^\ast,\lambda)\leq L(\theta(t),\nu^\ast(\theta(t)),\lambda) \leq L(\theta(0),\nu^\ast(\theta(0)),\lambda)\leq L(\theta(0),\nu(t),\lambda)\leq L(\theta(0),\nu(0),\lambda)$ for any $t\geq 0$. 

By contradiction, suppose $(\theta^\ast,\nu^\ast)$ is not a local minimum. Then there exists $(\bar\theta,\bar\nu)\in\Theta\times[-\frac{C_{\max}}{1-\gamma},\frac{C_{\max}}{1-\gamma}]\cap B_{(\theta^\ast,\nu^\ast)}(r)$ such that $L(\bar\theta,\bar\nu,\lambda)=\min_{(\theta,\nu)\in\Theta\times [-\frac{C_{\max}}{1-\gamma},\frac{C_{\max}}{1-\gamma}]\cap B_{(\theta^\ast,\nu^\ast)}(r)}L(\theta,\nu,\lambda)$. The minimum is attained by Weierstrass extreme value theorem. By putting $\theta(0)=\bar\theta$, the above arguments imply that
\[
L(\bar\theta,\bar\nu,\lambda)=\min_{(\theta,\nu)\in\Theta\times [-\frac{C_{\max}}{1-\gamma},\frac{C_{\max}}{1-\gamma}]\cap B_{(\theta^\ast,\nu^\ast)}(r)}L(\theta,\nu,\lambda)<L(\theta^\ast,\nu^\ast,\lambda)\leq L(\bar\theta,\bar\nu,\lambda)
\]
which is clearly a contradiction. Therefore, the stationary point $(\theta^\ast,\nu^\ast)$ is a local minimum of $L(\theta,\nu,\lambda)$ as well.
%%%%%%%%%%%%%%%%%%%%%%%%%%%%%%%%%%%%%%%%%%%%%%%%%%%%%%%%%%%%%%

\noindent\paragraph{Step 4 (Convergence of $\lambda-$update)} Since $\lambda$-update converges in the slowest time scale, it can be rewritten using the converged $\theta^*(\lambda)=\theta^*(\nu^*(\lambda),\lambda)$ and $\nu^*(\lambda)$, i.e.,
\begin{equation}\label{lamda_up_h_conv}
\lambda_{i+1}=\Gamma_\Lambda\left(\lambda_i+\zeta_1(i)\bigg(\nabla_\lambda L(\theta,\nu,\lambda)\bigg\vert_{\theta=\theta^*(\lambda_i),\nu=\nu^*(\lambda_i),\lambda=\lambda_i}+\delta\lambda_{i+1}\bigg)\right)
\end{equation}
where 
\begin{equation}\label{eq:MG_diff_theta_h}
\begin{split}
&\delta\lambda_{i+1}=-\nabla_\lambda L(\theta,\nu,\lambda)\bigg\vert_{\theta=\theta^*(\lambda),\nu=\nu^*(\lambda),\lambda=\lambda_i}+\bigg(\nu^*(\lambda_i)+\frac{1}{1-\alpha} \frac{1}{N}\sum_{j=1}^N\big(D(\xi_{j,i})- \nu^*(\lambda_i)\big)^+ -\beta\bigg).
\end{split}
\end{equation}

From \eqref{eq:L_lambda}, it is obvious that $\nabla_\lambda L(\theta,\nu,\lambda)$ is a constant function of $\lambda$. Similar to $\theta-$update,  one can easily show that $\delta\lambda_{i+1}$ is square integrable, i.e.,
\[
\mathbb E[\|\delta\lambda_{i+1}\|^2\mid \mathcal F_{\lambda,i}]\leq 2\left(\beta+\frac{3C_{\max}}{(1-\gamma)(1-\alpha)}\right)^2,
\]
where $\mathcal F_{\lambda,i}=\sigma\big(\lambda_m,\,\delta \lambda_m,\,m\leq i\big)$ is the filtration of $\lambda$ generated by different independent trajectories.
Furthermore, expression \eqref{eq:L_lambda} implies that $\mathbb E\left[\delta\lambda_{i+1}\mid \mathcal F_{\lambda,i}\right]=0$.  Therefore, the $\lambda$-update is a stochastic approximation of the ODE~\eqref{dyn_sys_lambda_h} with a Martingale difference error term. In addition, from the convergence analysis of $(\theta,\nu)-$update, $(\theta^\ast(\lambda),\nu^\ast(\lambda))$ is an asymptotically stable equilibrium point of $\{\theta_i,\nu_i\}$. From \eqref{eq:L_theta}, $\nabla_{\theta} L(\theta,\nu,\lambda)$ is a linear mapping in $\lambda$, it can be easily seen that $(\theta^\ast(\lambda),\nu^\ast(\lambda))$ is a Lipschitz continuous mapping of $\lambda$.

Consider the ODE of $\lambda\in [0,\lambda_{\max}]$ in~\eqref{dyn_sys_lambda_h}. Analogous to the arguments in the $\theta-$update, we may write
\[
\frac{d L(\theta,\nu,\lambda)}{dt}\bigg\vert_{\theta=\theta^*(\lambda),\nu=\nu^\ast(\lambda)} =\nabla_\lambda L(\theta,\nu,\lambda)\bigg\vert_{\theta=\theta^*(\lambda),\nu=\nu^\ast(\lambda)}\!\!\Upsilon_\lambda\left[\nabla_\lambda L(\theta,\nu,\lambda)\bigg\vert_{\theta=\theta^*(\lambda),\nu=\nu^\ast(\lambda)}\right].
\]
and show that ${d L(\theta,\nu,\lambda)}/{dt}\vert_{\theta=\theta^*(\lambda),\nu=\nu^\ast(\lambda)}\leq 0$, this quantity is non-zero whenever $\left\|\Upsilon_\lambda\left[{ d L(\theta,\nu,\lambda)}/{d\lambda}\vert_{\theta=\theta^*(\lambda),\nu=\nu^\ast(\lambda)}\right]\right\|\neq 0$. Lasalle's invariance principle implies that $\lambda^*\in\Lambda_{c}:=\{\lambda\in[0,\lambda_{\max}]:\Upsilon_\lambda[\nabla_\lambda L(\theta,\nu,\lambda)\mid_{\nu=\nu^\ast(\lambda),\theta=\theta^*(\lambda)}]=0\}$ is a stable equilibrium point.  

Based on the above properties and noting that the step size rule follows from Section~\ref{subsec:ass}, one can apply the multi-time scale stochastic approximation theory (Theorem 2 in Chapter 6 of \cite{borkar2008stochastic}) to show that the sequence $\{\lambda_i\}$ converges almost surely to a fixed point $\lambda^*\in [0,\lambda_{\max}]$ of ODE~\eqref{dyn_sys_lambda_h}, where $\lambda^*\in\Lambda_{c}:=\{\lambda\in[0,\lambda_{\max}]:\Upsilon_\lambda[\nabla_\lambda L(\theta,\nu,\lambda)\mid_{\theta=\theta^*(\lambda),\nu=\nu^\ast(\lambda)}]=0\}$. Since $\Lambda_c$ is a closed set of $[0,\lambda_{\max}]$, it is a compact set as well. Following the same lines of arguments and recalling the envelope theorem (Theorem \ref{thm:envelop}) for local optimum, one further concludes that $\lambda^\ast$ is a local maximum of $L(\theta^\ast(\lambda),\nu^\ast(\lambda),\lambda)=L^\ast(\lambda)$.

%%%%%%%%%%%%%%%%%%%%%%%%%%%%%%%%%%%%%%%%%%%%%%%%%%%%%%%%%%%%%%

\noindent\paragraph{Step 5 (Saddle Point)} 
By letting $\theta^*=\theta^*\big(\nu^*(\lambda^*),\lambda^*\big)$ and $\nu^*=\nu^*(\lambda^*)$, we will show that $(\theta^*,\nu^*,\lambda^*)$ is a (local) saddle point of the objective function $L(\theta,\nu,\lambda)$ if $\lambda^\ast\in[0,\lambda_{\max})$. 

Now suppose the sequence $\{\lambda_i\}$ generated from \eqref{lamda_up_h_conv} converges to a stationary point $\lambda^\ast\in[0,\lambda_{\max})$. Since step~3 implies that $(\theta^*,\nu^*)$ is a local minimum of $L(\theta,\nu,\lambda^\ast)$ over feasible set $(\theta,\nu)\in\Theta\times[-\frac{C_{\max}}{1-\gamma},\frac{C_{\max}}{1-\gamma}]$, there exists a $r>0$ such that
\[
L(\theta^*,\nu^*,\lambda^*)\leq L(\theta,\nu,\lambda^*),\quad\forall (\theta,\nu)\in \Theta\times\left[-\frac{C_{\max}}{1-\gamma},\frac{C_{\max}}{1-\gamma}\right]\cap B_{(\theta^\ast,\nu^\ast)}(r).
\]

In order to complete the proof, we must show
\begin{equation}\label{prop_1}
\nu^*+\frac{1}{1-\alpha}\expec\left[\big(D^{\theta^*}(x^0)- \nu^*\big)^+\right] \leq  \beta,
\end{equation}
and
\begin{equation}\label{prop_2}
\lambda^* \left( \nu^*+\frac{1}{1-\alpha}\expec\left[\big(D^{\theta^*}(x^0)- \nu^*\big)^+\right] - \beta\right)=0.
\end{equation}
These two equations imply 
\[
\begin{split}
L(\theta^*,\nu^*,\lambda^*)=&V^{\theta^*}(x^0) \!+\!\lambda^*\left(\nu^* + \frac{1}{1-\alpha}\expec\left[\big(D^{\theta^*}(x^0)- \nu^*\big)^+\right] - \beta\right) \\
=&V^{\theta^*}(x^0) \\
\geq &V^{\theta^*}(x^0) \!+\!\lambda  \left(\nu^*+\frac{1}{1-\alpha}\expec\left[\big(D^{\theta^*}(x^0)- \nu^*\big)^+\right] - \beta\right)=L(\theta^*,\nu^*,\lambda), 
\end{split}
\]
which further implies that $(\theta^*,\nu^*,\lambda^*)$ is a saddle point of $L(\theta,\nu,\lambda)$. We now show that \eqref{prop_1} and~\eqref{prop_2} hold. 

Recall that $\Upsilon_\lambda\left[\nabla_\lambda L(\theta,\nu,\lambda)\vert_{\theta=\theta^*(\lambda),\nu=\nu^*(\lambda),\lambda=\lambda^*}\right]\vert_{\lambda=\lambda^*}=0$. We show \eqref{prop_1} by contradiction. Suppose $\nu^*+\frac{1}{1-\alpha}\expec\left[\big(D^{\theta^*}(x^0)- \nu^*\big)^+\right] > \beta$. This then implies that for $\lambda^*\in[0,\lambda_{\max})$, we have

\vspace{-0.1in}
\begin{small}
\begin{equation*}
\Gamma_\Lambda\left(\lambda^*-\eta\bigg(\beta-\Big(\nu^*+\frac{1}{1-\alpha}\expec\big[\big(D^{\theta^*}(x^0)-\nu^*\big)^+\big]\Big)\bigg)\right)=\lambda^*-\eta\bigg(\beta-\Big(\nu^*+\frac{1}{1-\alpha}\expec\big[\big(D^{\theta^*}(x^0)- \nu^*\big)^+\big]\Big)\bigg)
\end{equation*}
\end{small}
for any $\eta\in (0,\eta_{\max}]$ for some sufficiently small $\eta_{\max}>0$. Therefore, 
\[
\Upsilon_\lambda\left[\nabla_\lambda L(\theta,\nu,\lambda)\bigg\vert_{\theta=\theta^*(\lambda),\nu=\nu^*(\lambda),\lambda=\lambda^*}\right]\Bigg\vert_{\lambda=\lambda^*}
= \nu^*+\frac{1}{1-\alpha}\expec\left[\big(D^{\theta^*}(x^0)- \nu^*\big)^+\right] -\beta>0.
\]
This contradicts with $\Upsilon_\lambda\left[\nabla_\lambda L(\theta,\nu,\lambda)\vert_{\theta=\theta^*(\lambda),\nu=\nu^*(\lambda),\lambda=\lambda^*}\right]\vert_{\lambda=\lambda^*}=0$. Therefore,~\eqref{prop_1} holds.

To show that \eqref{prop_2} holds, we only need to show that $\lambda^*=0$ if $\nu^*+\frac{1}{1-\alpha}\expec\left[\big(D^{\theta^*}(x^0)- \nu^*\big)^+\right] < \beta$. Suppose $\lambda^*\in(0,\lambda_{\max})$, then there exists a sufficiently small $\eta_0>0$ such that
\[
\begin{split}
&\frac{1}{\eta_0}\left(\Gamma_\Lambda\bigg(\lambda^*-\eta_0\Big(\beta-\big(\nu^*+\frac{1}{1-\alpha}\expec\big[\big(D^{\theta^*}(x^0)- \nu^*\big)^+\big]\big)\Big)\bigg)-\Gamma_\Lambda(\lambda^\ast)\right)\\
=& \nu^*+\frac{1}{1-\alpha}\expec\left[\big(D^{\theta^*}(x^0)- \nu^*\big)^+\right] -\beta<0.
\end{split}
\]
This again contradicts with the assumption $\Upsilon_\lambda\left[\nabla_\lambda L(\theta,\nu,\lambda)\vert_{\theta=\theta^*(\lambda),\nu=\nu^*(\lambda),\lambda=\lambda^*}\right]\vert_{\lambda=\lambda^*}=0$ from \eqref{eq_cond_1}. Therefore \eqref{prop_2} holds. 

Combining the above arguments, we finally conclude that $(\theta^*,\nu^*,\lambda^*)$ is a (local) saddle point of $L(\theta,\nu,\lambda)$ if $\lambda^\ast\in[0,\lambda_{\max})$.
\end{prooff}

\begin{remark}
When $\lambda^\ast=\lambda_{\max}$ and $\nu^*+\frac{1}{1-\alpha}\expec\left[\big(D^{\theta^*}(x^0)- \nu^*\big)^+\right] > \beta$,
\[
\Gamma_\Lambda\left(\lambda^*-\eta\bigg(\beta-\Big(\nu^*+\frac{1}{1-\alpha}\expec\big[\big(D^{\theta^*}(x^0)-\nu^*\big)^+\big]\Big)\bigg)\right)=\lambda_{\max}
\]
for any $\eta>0$ and 
\[
\Upsilon_\lambda\left[\nabla_\lambda L(\theta,\nu,\lambda)\vert_{\theta=\theta^*(\lambda),\nu=\nu^*(\lambda),\lambda=\lambda^*}\right]\mid_{\lambda=\lambda^*}=0.
\]
In this case one cannot guarantee feasibility using the above analysis, and $(\theta^\ast,\nu^\ast,\lambda^\ast)$ is not a local saddle point. Such $\lambda^\ast$ is referred as a spurious fixed point \cite{kushner1997stochastic}. Practically, by incrementally increasing $\lambda_{\max}$ (see Algorithm \ref{alg_traj} for more details), when $\lambda_{\max}$ becomes sufficiently large, one can ensure that the policy gradient algorithm will not get stuck at the spurious fixed point.
\end{remark}

%%%%%%%%%%%%%%%%%%%%%%%%%%%%%%%%%%%%%%%%%%%%%%%%%%%%%%%%%%%%%%
%%%%%%%%%%%%%%%%%%%%%%%%%%%%%%%%%%%%%%%%%%%%%%%%%%%%%%%%%%%%%%
%%%%%%%%%%%%%%%%%%%%%%%%%%%%%%%%%%%%%%%%%%%%%%%%%%%%%%%%%%%%%%
%%%%%%%%%%%%%%%%%%%%%%%%%%%%%%%%%%%%%%%%%%%%%%%%%%%%%%%%%%%%%%
%%%%%%%%%%%%%%%%%%%%%%%%%%%%%%%%%%%%%%%%%%%%%%%%%%%%%%%%%%%%%%

\newpage
\section{Technical Details of the Actor-Critic Algorithms}
\label{subsec:ass_incre}

%%%%%%%%%%%%%%%%%%%%%%%%%%%%%%%%%%%%%%%%%%%%%%%%%%%%%%%%%%%%%%
%%%%%%%%%%%%%%%%%%%%%%%%%%%%%%%%%%%%%%%%%%%%%%%%%%%%%%%%%%%%%%
%%%%%%%%%%%%%%%%%%%%%%%%%%%%%%%%%%%%%%%%%%%%%%%%%%%%%%%%%%%%%%

\subsection{Assumptions}
\label{subsec:ass2}

We make the following assumptions for the proof of our actor-critic algorithms:

\noindent
{\bf (B1)} {\em For any state-action pair $(x,s,a)$ in the augmented MDP $\bar{\mathcal{M}}$, $\mu(a|x,s;\theta)$ is continuously differentiable in  $\theta$ and $\nabla_\theta\mu(a|x,s;\theta)$ is a Lipschitz function in $\theta$ for every $a\in\A$, $x\in\X$ and $s\in\reals$.} \\

\noindent
{\bf (B2)} {\em The augmented Markov chain induced by any policy $\theta$, $\bar{\mathcal{M}}^\theta$, is irreducible and aperiodic.} \\

\noindent
{\bf (B3)} {\em The basis functions $\big\{\phi^{(i)}\big\}_{i=1}^{\kappa_2}$ are linearly independent. In particular, $\kappa_2 \ll n$ and $\Phi$ is full rank.\footnote{We may write this as: In particular, the (row) infinite dimensional matrix $\Phi$ has column rank $\kappa_2$.} Moreover, for every $v\in\reals^{\kappa_2}$, $\Phi v\neq e$, where $e$ is the $n$-dimensional vector with all entries equal to one.} \\

\noindent
{\bf (B4)} {\em For each $(x^\prime,s^\prime,a^\prime)\in\bar{\X}\times\bar{\A}$, there is a positive probability of being visited, i.e.,~$\pi_\gamma^\theta(x^\prime,s^\prime,a^\prime|x,s)>0$. Note that from the definition of the augmented MDP $\bar{\mathcal{M}}$, $\bar{\X}=\X\times\reals$ and $\bar{\A}=\A$.} \\
\noindent
{\bf (B5)} {\em The step size schedules $\{\zeta_4(k)\}$, $\{\zeta_3(k)\}$, $\{\zeta_2(k)\}$, and $\{\zeta_1(k)\}$ satisfy} %($k$ is some positive constant)}

\begin{align}
\label{eq:step1_incre}
&\sum_k \zeta_1(k) = \sum_k \zeta_2(k) = \sum_k \zeta_3(k) = \sum_k \zeta_4(k)=\infty, \\
\label{eq:step2_incre}
&\sum_k \zeta_1(k)^2,\;\;\;\sum_k \zeta_2(k)^2,\;\;\;\sum_k \zeta_3(k)^2,\;\;\;\sum_k \zeta_4(k)^2<\infty, \\
\label{eq:step3_incre}
&\zeta_1(k) = o\big(\zeta_2(k)\big), \;\;\; \zeta_2(k) = o\big(\zeta_3(k)\big),\;\;\; \zeta_3(k) = o\big(\zeta_4(k)\big).%, \;\;\; \zeta_4(k) = k\zeta_3(k).
\end{align}
This indicates that the updates correspond to $\{\zeta_4(k)\}$ is on the fastest time-scale, the update corresponds to $\{\zeta_3(k)\}$, $\{\zeta_2(k)\}$ are on the intermediate time-scale, where $\zeta_3(k)$ converges faster than $\zeta_2(k)$, and the update corresponds to $\{\zeta_1(k)\}$ is on the slowest time-scale.

\noindent {\bf (B6)} {\em The SPSA step size $\{\Delta_k\}$ satisfies $\Delta_k\rightarrow\infty$ as $k\rightarrow\infty$ and $\sum_k (\zeta_2(k)/\Delta_k)^2<\infty$.}

Technical assumptions for the convergence of the actor-critic algorithm will be given in the section for the proof of convergence.

%%%%%%%%%%%%%%%%%%%%%%%%%%%%%%%%%%%%%%%%%%%%%%%%%%%%%%%%%%%%%%
%%%%%%%%%%%%%%%%%%%%%%%%%%%%%%%%%%%%%%%%%%%%%%%%%%%%%%%%%%%%%%
%%%%%%%%%%%%%%%%%%%%%%%%%%%%%%%%%%%%%%%%%%%%%%%%%%%%%%%%%%%%%%

\subsection{Gradient with Respect to $\lambda$ (Proof of Lemma \ref{lem:grad_lambda})}
\label{subsec:grad-lambda-comp}

%From the definition of the Lagrangian function, one obtains
%\[
%\nabla_\lambda L(\theta,\nu,\lambda)=\nabla_\lambda\V^\theta(x,s)\bigg\vert_{x=x^0,s=\nu}+(\nu-K).
%\]
\begin{prooff}
By taking the gradient of $V^\theta(x^0,\nu)$ w.r.t.~$\lambda$ (just a reminder that both $V$ and $Q$ are related to $\lambda$ through the dependence of the cost function $\bar{C}$ of the augmented MDP $\bar{\mathcal{M}}$ on $\lambda$), we obtain

\vspace{-0.1in}
\begin{small}
\begin{align}
\label{eq:grad-lambda0}
\nabla_\lambda V^\theta(x^0,\nu)&=\sum_{a\in\bar{\A}}\mu(a|x^0,\nu;\theta)\nabla_{\lambda}Q^\theta(x^0,\nu,a) \nonumber \\
&=\sum_{a\in\bar{\A}}\mu(a|x^0,\nu;\theta)\nabla_{\lambda}\Big[\bar{C}(x^0,\nu,a)+\sum_{(x',s')\in\bar{\X}}\gamma\bar{P}(x',s'|x^0,\nu,a)V^\theta(x',s')\Big] \nonumber \\
&=\underbrace{\sum_a\mu(a|x^0,\nu;\theta)\nabla_{\lambda}\bar{C}(x^0,\nu,a)}_{h(x^0,\nu)} + \gamma\sum_{a,x',s'}\mu(a|x^0,\nu;\theta)\bar{P}(x',s'|x^0,\nu,a)\nabla_{\lambda}V^\theta(x',s') \nonumber \\
&=h(x^0,\nu) + \gamma\sum_{a,x',s'}\mu(a|x^0,\nu;\theta)\bar{P}(x',s'|x^0,\nu,a)\nabla_{\lambda}V^\theta(x',s') \\
&=h(x^0,\nu) + \gamma\sum_{a,x',s'}\mu(a|x^0,\nu;\theta)\bar{P}(x',s'|x^0,\nu,a)\Big[h(x',s') \nonumber \\ 
&\hspace{0.6in}+\gamma\sum_{a',x'',s''}\mu(a'|x',s';\theta)\bar{P}(x'',s''|x',s',a')\nabla_\lambda V^\theta(x'',s'')\Big] \nonumber
\end{align}
\end{small}
\vspace{-0.15in}

By unrolling the last equation using the definition of $\nabla_\lambda V^\theta(x,s)$ from~\eqref{eq:grad-lambda0}, we obtain

\vspace{-0.1in}
\begin{small}
\begin{align*}
\label{eq:theta_gradient}
\nabla_\lambda V^\theta(x^0,\nu)&=\sum_{k=0}^\infty\gamma^k\sum_{x,s}\mathbb P(x_k=x,s_k=s\mid x_0=x^0,s_0=\nu;\theta)h(x,s) \\ 
&= \frac{1}{1-\gamma}\sum_{x,s}d_\gamma^\theta(x,s|x^0,\nu)h(x,s) = \frac{1}{1-\gamma}\sum_{x,s,a}d_\gamma^\theta(x,s|x^0,\nu)\mu(a|x,s)\nabla_\lambda\bar{C}(x,s,a) \\
&= \frac{1}{1-\gamma}\sum_{x,s,a}\pi_\gamma^\theta(x,s,a|x^0,\nu)\nabla_\lambda\bar{C}(x,s,a) \\
&= \frac{1}{1-\gamma}\sum_{x,s,a}\pi_\gamma^\theta(x,s,a|x^0,\nu)\frac{1}{1-\alpha}\mathbf 1 \{x=x_T\}(-s)^+. 
% &=\int_{s^\prime\in\reals}\mu(a^\prime|x^\prime,s^\prime;\theta)\bar{P}(x_k=x^\prime,s_k=s^\prime|x_0=x,s_0=s)h^\lambda(x^\prime,s^\prime)ds^\prime\\
%=&\frac{1}{1-\gamma}\sum_{x^\prime\in\X} \int_{s^\prime\in\reals}d_{\gamma}^\theta(x^\prime,s^\prime|x_0=x,s_0=s)h^\lambda(x^\prime,s^\prime)ds^\prime\\
%=&\frac{1}{1-\gamma}\sum_{x^\prime\in\X} \int_{s^\prime\in\reals}d_{\gamma}^\theta(x^\prime,s^\prime|x_0=x,s_0=s)\sum_{a^\prime\in \A}\mu(a^\prime|x^\prime,s^\prime;\theta)\frac{1}{1-\alpha}\mathbf 1 \{x^\prime=x_T\}(-s^\prime)^+ds^\prime\\
%=&\frac{1}{1-\gamma}\sum_{x^\prime\in\X,a^\prime\in\A} \int_{s^\prime\in\reals}\pi_{\gamma}^\theta(x^\prime,s^\prime,a^\prime|x_0=x,s_0=s)\frac{1}{1-\alpha}\mathbf 1 \{x^\prime=x_T\}(-s^\prime)^+ds^\prime.
\end{align*}
\end{small}
\end{prooff}

\subsection{Actor-Critic Algorithm with the Alternative Approach to Compute the Gradients}
\label{subsec:alt-alg}

\begin{algorithm}
\begin{small}
\begin{algorithmic}
\WHILE{1}
\STATE {\bf Input:} Parameterized policy $\mu(\cdot|\cdot;\theta)$, value function feature vectors $f(\cdot)$ and $\phi(\cdot)$, confidence level $\alpha$, and loss tolerance $\beta$
\STATE {\bf Initialization:} policy parameters $\theta=\theta_0$; VaR parameter $\nu=\nu_0$; Lagrangian parameter $\lambda=\lambda_0$; value function weight vectors $u=u_0$ and $v=v_0$ 
%\STATE 
%\STATE \textbf{(1) SPSA-based Algorithm:}
\FOR{$k = 0,1,2,\ldots$}
\STATE Draw action $\;a_k\sim\mu(\cdot|x_k,s_k;\theta_k)$
\STATE Observe next state $(x_{k+1},s_{k+1})\sim \bar{P}(\cdot|x_k,s_k,a_k)$; $\;$ \begin{footnotesize}{\em // note that $s_{k+1}=(s_k-C\big(x_k,a_k)\big)/\gamma\;$ (see Sec.~\ref{subsec:grad-theta})}\end{footnotesize}
Observe costs $\;C(x_k,a_k)$ and $\;\bar{C}(x_k,s_k,a_k)$ $\;$ \begin{footnotesize}{\em // $\bar{C}$ and $\bar{P}$ are the cost and transition functions of the}\end{footnotesize}
\STATE \begin{footnotesize}{\em $\quad\;\;$ // augmented MDP $\bar{\mathcal{M}}$ defined in Sec.~\ref{subsec:grad-alter}, while $C$ is the cost function of the original MDP $\mathcal{M}$}\end{footnotesize}
\begin{align}
\textrm{\bf TD Errors:} \quad & \epsilon_k(u_k) = C(x_k,a_k) + \gamma u_k^\top f(x_{k+1}) - u_k^\top f(x_k) \\ %\label{TD-calc2} \\
& \delta_k(v_k) = \bar{C}(x_k,s_k,a_k) + \gamma v_k^\top\phi(x_{k+1},s_{k+1}) - v_k^\top\phi(x_k,s_k) \\
\textrm{\bf Critic Updates:} \quad & u_{k+1}=u_k+\zeta_4(k)\epsilon_k(u_k)f(x_k) \\ %\label{v_up_incre2} \\
& v_{k+1}=v_k+\zeta_4(k)\delta_k(v_k)\phi(x_k,s_k) \\
\textrm{{\bf Actor Updates:}}\quad & \nu_{k+1} = \Gamma_N\left(\nu_k - \zeta_3(k)\lambda_k\left(1 + \frac{v_k^\top\big[\phi\big(x^0,\nu_k+\Delta_k\big)- \phi(x^0,\nu_k-\Delta_k)\big]}{2(1-\alpha)\Delta_k}\right)\right) \\%\label{nu_up_incre_SPSA2} \\
& \theta_{k+1} = \Gamma_\Theta\left(\theta_k-\frac{\zeta_2(k)}{1-\gamma}\nabla_\theta\log\mu(a_k|x_k,s_k;\theta)\vert_{\theta=\theta_k}\cdot\left(\epsilon_k(u_k)+\frac{\lambda_k}{1-\alpha}\delta_k(v_k)\right)\right) \\%\label{theta_up_incre2} \\
&\lambda_{k+1} = \Gamma_\Lambda\left(\lambda_k + \zeta_1(k)\left(\nu_k - \beta + \frac{v^\top\phi(x_k,s_k)}{1-\alpha}\right)\right) %\label{lambda_up_incre2}
\end{align}
\ENDFOR
\IF{$\{\lambda_i\}$ converges to $\lambda_{\max}$}
\STATE{Set $\lambda_{\max}\leftarrow 2\lambda_{\max}$.}
\ELSE
\STATE {\bf return} policy and value function parameters $v,u,\nu,\theta,\lambda$ and {\bf break}
\ENDIF
\ENDWHILE
\end{algorithmic}
\end{small}
\caption{Actor-Critic Algorithm for CVaR Optimization (Alternative Gradient Computation)}
\label{alg:AC-alt}
\end{algorithm}

%%%%%%%%%%%%%%%%%%%%%%%%%%%%%%%%%%%%%%%%%%%%%%%%%%%%%%%%%%%%%%
%%%%%%%%%%%%%%%%%%%%%%%%%%%%%%%%%%%%%%%%%%%%%%%%%%%%%%%%%%%%%%
%%%%%%%%%%%%%%%%%%%%%%%%%%%%%%%%%%%%%%%%%%%%%%%%%%%%%%%%%%%%%%

\subsection{Convergence of the Actor Critic Algorithms}
\label{subsec:convergence-proof-AC}
In this section we want to derive the following convergence results.
\begin{theorem}\label{thm:converge_incre_w}
Suppose $v^*\in\arg\min_{v}\|T_{\theta}[\Phi v]-\Phi v\|_{d^\theta_\gamma}^2$, where 
\[
T_\theta[V](x,s)=\sum_{a}\mu(a|x,s;\theta)\left\{\bar{C}(x,s,a)+\sum_{x^\prime,s^\prime}\bar{P}(x^\prime,s^\prime|x,s,a)V(x^\prime,s^\prime)\right\}
\]
and $\tilde V^*(x,s)=\phi^\top(x,s)v^*$ is the projected Bellman fixed point of $ V^\theta(x,s)$, i.e., $\tilde V^*(x,s)=\Pi T_{\theta}[\tilde V^*](x,s)$.
Also suppose the $\gamma-$stationary distribution $\pi_{\gamma}^\theta$ is used to generate samples of $(x_k,s_k,a_k)$ for any $k\in\{0,1,\ldots,\}$. Then the $v-$updates in the actor critic algorithms converge to $v^*$ almost surely.
\end{theorem}
Next define 
\[
\epsilon_\theta(v_k)=\|T_{\theta}[\Phi v_k]-\Phi v_k\|^2_{d^\theta_\gamma}
\]
as the residue of the value function approximation at step $k$ induced by policy $\mu(\cdot|\cdot,\cdot;\theta)$. By triangular inequality and fixed point theorem $T_{\theta}[V^*]=V^*$, it can be easily seen that $\|V^*-\Phi v_k\|^2_{d^\theta_\gamma}\leq \epsilon_\theta(v_k)+\|T_{\theta}[\Phi v_k]-T_{\theta}[V^*]\|^2_{d^\theta_\gamma}\leq \epsilon_\theta(v_k)+\gamma\|\Phi v_k-V^*\|^2_{d^\theta_\gamma}$. The last inequality follows from the contraction mapping argument. Thus, one concludes that $\|V^*-\Phi v_k\|^2_{d^\theta_\gamma}\leq \epsilon_\theta(v_k)/(1-\gamma)$.
\begin{theorem}\label{thm:converge_incre}
Suppose $\lambda^\ast\in[0,\lambda_{\max})$, $\epsilon_{\theta_k}(v_k)\rightarrow 0$ as $t$ goes to infinity and the $\gamma-$stationary distribution $\pi_{\gamma}^\theta$ is used to generate samples of $(x_k,s_k,a_k)$ for any $k\in\{0,1,\ldots,\}$. For SPSA based algorithm, also suppose the perturbation sequence $\{\Delta_k\}$ satisfies $\epsilon_{\theta_k}(v_k)\mathbb E[1/\Delta_k]\rightarrow 0$. Then the sequence of $(\theta,\nu,\lambda)$-updates in Algorithm \ref{alg:AC} converges to a (local) saddle point $(\theta^*,\nu^*,\lambda^*)$ of our objective function $L(\theta,\nu,\lambda)$ almost surely, it satisfies $L(\theta,\nu, \lambda^*) \ge L(\theta^*,\nu^\ast, \lambda^*) \ge L(\theta^*,\nu^\ast, \lambda),\forall(\theta,\nu)\in\Theta\times [-C_{\max}/(1-\gamma),C_{\max}/(1-\gamma)]\cap B_{(\theta^*,\nu^*)}(r)$ for some $r>0$ and $\forall \lambda\in [0,\lambda_{\max}]$. Note that $B_{(\theta^*,\nu^*)}(r)$ represents a hyper-dimensional ball centered at $(\theta^*,\nu^*)$ with radius $r$.
\end{theorem}

Since the proof of the Multi-loop algorithm and the SPSA based algorithm is almost identical (except the $\nu-$update), we will focus on proving the SPSA based actor critic algorithm.  
\subsubsection{Proof of Theorem \ref{thm:converge_incre_w}: TD(0) Critic Update ($v-$update)} 
By the step length conditions, one notices that $\{v_k\}$ converges in a faster time scale than $\{\theta_k\}$, $\{\nu_{k}\}$ and $\{\lambda_k\}$, one can assume $(\theta,\nu,\lambda)$ in the $v-$update as fixed quantities. The critic update can be re-written as follows:
\begin{equation}\label{eq:TD_0}
v_{k+1}=v_k+\zeta_4(k)\phi(x_k,s_k)\delta_k(v_k)
\end{equation}
where the scaler 
\[
\delta_k\left(v\right)=-\phi^\top(x_k,s_k)v+ \gamma\phi^\top\left(x_{k+1},s_{k+1}\right)v+ \bar{C}(x_k,s_k,a_k).
\]
is known as the temporal difference (TD). Define
\begin{equation}\label{eq:A}
\begin{split}
A=&\sum_{y,a^\prime,s^\prime}\pi_{\gamma}^\theta(y,s^\prime,a^\prime|x,s)\phi(y,s^\prime)\left(\phi^\top(y,s^\prime)- \gamma\sum_{z,s^{\prime\prime}}\bar{P}(z,s^{\prime\prime}|y,s^\prime,a) \phi^\top\left(z,s^{\prime\prime}\right)\right)
\end{split}
\end{equation}
and 
\begin{equation}\label{eq:b}
b=\sum_{yX,a^\prime,s^\prime}\pi_{\gamma}^\theta(y,s^\prime,a^\prime|x,s)\phi(y,s^\prime)\bar{C}(y,s^\prime,a^\prime).
\end{equation}
Based on the definitions of matrices $A$ and $b$, it is easy to see that the TD(0) critic update $v_k$ in \eqref{eq:TD_0} can be re-written as the following stochastic approximation scheme:
\begin{equation}
v_{k+1}=v_k+\zeta_4(k)(b-Av_k+\delta A_{k+1})
\end{equation}
where the noise term $\delta A_{k+1}$ is a square integrable Martingale difference, i.e, $\mathbb E[\delta A_{k+1}\mid \mathcal F_k]=0$ if the $\gamma-$stationary distribution $\pi_{\gamma}^\theta$ used to generate samples of $(x_k,s_k,a_k)$. $\mathcal F_k$ is the filtration generated by different independent trajectories. By writing 
\[
\delta A_{k+1}=-(b-Av_k)+\phi(x_k,s_k)\delta_k(v_k)
\]
and noting $\mathbb E_{\pi_\gamma^\theta}[\phi(x_k,s_k)\delta_k(v_k)\mid \mathcal F_k]=-Av_k+b$,
one can easily check that the stochastic approximation scheme in \eqref{eq:TD_0} is equivalent to the TD(0) iterates in \eqref{eq:TD_0} and $\delta A_{k+1}$ is a Martingale difference, i.e., $\mathbb E_{\pi_\gamma^\theta}[\delta A_{k+1}\mid \mathcal F_k]=0$.
 Let 
\[
h\left(v\right)=-Av+b.
\]
Before getting into the convergence analysis, we have the following technical lemma.
\begin{lemma}
Every eigenvalues of matrix $A$ has positive real part.
\end{lemma}
\begin{prooff}
To complete this proof, we need to show that for any vector $v\in\reals^{\kappa_2}$, $v^\top  Av>0$. Now, for any fixed $v\in\reals^{\kappa_2}$, define $y(x,s)=v^\top\phi^\top(x,s)$. It can be easily seen from the definition of $A$ that 
\[
v^\top  Av=\sum_{x,x^\prime,a,s,s^\prime}\!\!\! y(x,s)\pi_{\gamma}^\theta(x,s,a|x_0=x^0,s_0=\nu)\cdot (\mathbf 1\{x^\prime=x,s^\prime=s\}-\gamma\bar{P}(x^\prime,s^{\prime}|x,s,a) )y(x^\prime,s^{\prime}).
\]
By convexity of quadratic functions and Jensen's inequality, one can derive the following expressions: 
\[\small
\begin{split}
&\sum_{x,x^\prime,a,s,s^\prime}y(x,s)\pi_{\gamma}^\theta(x,s,a|x_0=x^0,s_0=\nu) \gamma\bar{P}(x^\prime,s^{\prime}|x,s,a) y(x^\prime,s^{\prime})\\
\leq& \|y\|_{d_{\gamma}^\theta}\sqrt{\gamma}\sqrt{\sum_{x,x^\prime,a,s,s^\prime}d_{\gamma}^\theta(x,s|x_0=x^0,s_0=\nu) \gamma\mu(a|x,s;\theta)\P(x^\prime,s^{\prime}|x,s,a) (y(x^\prime,s^{\prime}))^2}\\
= &\|y\|_{d_{\gamma}^\theta}\sqrt{\sum_{y,s^\prime}\left(d_{\gamma}^\theta(y,s^\prime|x^0,\nu)-(1-\gamma)\mathbf 1\{x^0=y,\nu=s^\prime\} \right)(y(x^\prime,s^{\prime}))^2}\\
<& \|y\|^2_{d_{\gamma}^\theta}
\end{split}
\]
where $d_{\gamma}^\theta(x,s|x_0=x^0,s_0=\nu)\mu(a|x,s;\theta)=\pi_{\gamma}^\theta(x,s,a|x_0=x^0,s_0=\nu)$ and
\[
\|y\|_{d_{\gamma}^\theta}^2=\sum_{x,s}d_{\gamma}^\theta(x,s|x_0=x^0,s_0=\nu) (y(x,s))^2.
\]
The first inequality is due to the fact that $\mu( a| x, s;\theta),\bar{P}(y,s^\prime| x, s, a)\in[0,1]$ and convexity of quadratic function, the second equality is based on the stationarity property of a $\gamma-$visiting distribution: $d_\gamma^\theta(y,s^{\prime}|x^0,\nu)\geq 0$, $\sum_{y,s^\prime}d_\gamma^\theta(y,s^{\prime}|x^0,\nu)=1$ and
\[
\sum_{x^\prime,s,a} \pi_\gamma^\theta(x^\prime,s,a|x_0=x^0,s_0=\nu) \gamma\bar{P}(y,s^{\prime}|x^\prime,s,a^\prime)=d_\gamma^\theta(y,s^{\prime}|x^0,\nu)\\
-(1-\gamma)1\{x^0=y,\nu=s^\prime\}.
\]

As the above argument holds for any $v\in\reals^{\kappa_2}$ and $y(x,s)=v^\top\phi(x,s)$, one shows that $v^\top  Av>0$ for any $v\in\reals^{\kappa_2}$. This further implies $v^\top  A^\top v>0$ and $v^\top  (A^\top +A)v>0$ for any $v\in\reals^{\kappa_2}$. Therefore, $A+A^\top $ is a symmetric positive  definite matrix, i.e. there exists a $\epsilon>0$ such that $A+A^\top >\epsilon I$. 
To complete the proof, suppose by contradiction that there exists an eigenvalue $\lambda$ of $A$ which has a non-positive real-part. Let $v_\lambda$ be the corresponding eigenvector of $\lambda$. Then, by pre- and post-multiplying $v_\lambda^\ast$ and $v_\lambda$ to $A+A^\top >\epsilon I$ and noting that the hermitian of a real matrix  $A$ is $A^\top $, one obtains $2\text{Re}(\lambda)\|v_\lambda\|^2=v_\lambda^\ast(A+A^\top )v_\lambda=v_\lambda^\ast(A+A^\ast)v_\lambda>\epsilon \|v_\lambda\|^2$. This implies $\text{Re}(\lambda)>0$, i.e., a contradiction. By combining all previous arguments, one concludes that every eigenvalues $A$ has positive real part.
\end{prooff}

We now turn to the analysis of the TD(0) iteration. Note that the following properties hold for the TD(0) update scheme in (\ref{eq:TD_0}):
\begin{enumerate}
\item $h\left(v\right)$ is Lipschitz.\\
\item The step size satisfies the following properties in Appendix \ref{subsec:ass2}.\\
\item The noise term $\delta A_{k+1}$ is a square integrable Martingale difference.\\
\item The function 
\[
h_c\left(v\right):=h\left(cv\right)/c,\,\,c\geq 1
\]
converges uniformly to a continuous function $h_\infty\left(v\right)$ for any $w$ in a compact set, i.e., $h_c\left(v\right)\rightarrow h_\infty\left(v\right)$ as $c\rightarrow\infty$.\\
\item The ordinary differential equation (ODE) 
\[
\dot v= h_\infty\left(v\right)
\] 
has the origin as its unique globally asymptotically stable equilibrium.
\end{enumerate}
The fourth property can be easily verified from the fact that the magnitude of $b$ is finite and $h_\infty\left(v\right)=v$. The fifth property follows directly from the facts that $h_\infty\left(v\right)=-Av$ and all eigenvalues of $A$ have positive real parts. Therefore, by Theorem 3.1 in \cite{borkar2008stochastic}, these five properties imply the following condition:
\[
\text{The TD iterates $\{v_k\}$ is bounded almost surely, i.e.,}\,\,\sup_k\left\|v_k\right\|<\infty\,\,\,\text{almost surely}.
\]
Finally, from the standard stochastic approximation result, from the above conditions,
the convergence of the TD(0) iterates in \eqref{eq:TD_0} can be related to the asymptotic behavior of the ODE
\begin{equation}\label{exp:bdd}
\dot v= h\left(v\right)=b-Av.
\end{equation}
By Theorem 2 in Chapter 2 of \cite{borkar2008stochastic}, when
property (1) to (3) in \eqref{exp:bdd} hold, then $v_k\rightarrow v^*$ with probability $1$ where the limit $v^*$ depends on $(\theta,\nu,\lambda)$ and is the unique solution satisfying $h\left(v^*\right)=0$, i.e., $Av^*=b$. Therefore, the TD(0) iterates converges to the unique fixed point $v^*$ almost surely, at $k\rightarrow\infty$.

\subsubsection{Proof of Theorem \ref{thm:converge_incre}}
\noindent\paragraph{Step 1 (Convergence of $v-$update)}
The proof of the critic parameter convergence follows directly from Theorem \ref{thm:converge_incre_w}.

\noindent\paragraph{Step 2 (Convergence of SPSA based $\nu-$update)}
In this section, we present the $\nu-$update for the incremental actor critic method. This update is based on the SPSA perturbation method. The idea of this method is to estimate the sub-gradient $g(\nu)\in\partial_{\nu}  L(\theta,\nu,\lambda)$ using two simulated value functions corresponding to $\nu^{-}=\nu-\Delta$ and $\nu^{+}=\nu+\Delta$. Here $\Delta\geq 0$ is a positive random perturbation that vanishes asymptotically.

The SPSA-based estimate for a sub-gradient $g(\nu)\in\partial_{\nu}  L(\theta,\nu,\lambda)$ is given by:
\[
g(\nu)\approx \lambda+\frac{1}{2\Delta}\left( \phi^\top\left(x^0,\nu+\Delta\right)- \phi^\top\left(x^0,\nu-\Delta\right)\right)v
\]
where $\Delta\geq 0$ is a ``small" random perturbation of the finite difference sub-gradient approximation. 

Now, we turn to the convergence analysis of sub-gradient estimation and $\nu-$update. 
Since $v$ converges faster than $\nu$, and   $\nu$ converges faster then $\theta$ and $\lambda$, the $\nu-$update in \eqref{nu_up_incre_SPSA} can be rewritten using the converged critic-parameter $v^*(\nu)$ and $(\theta,\lambda$) in this expression is viewed as constant quantities, i.e.,
\begin{equation}\label{eq:s_grad_est}
\nu_{k+1}=\Gamma_N\left(\nu_{k}-\zeta_3(k)\left(
\lambda+\frac{1}{2\Delta_k} \left( \phi^\top\left(x^0,\nu_{k}+\Delta_k\right)- \phi^\top\left(x^0,\nu_{k}-\Delta_k\right)\right)v^*(\nu_{k})\right)\right).
\end{equation}

First, we have the following assumption on the feature functions in order to prove the SPSA approximation is asymptotically unbiased.
\begin{assumption}\label{assume_lip}
For any $v\in\reals^{\kappa_1}$, the feature function satisfies the following conditions
\[
|\phi^\top_V\left(x^0,\nu+\Delta\right)v-\phi^\top_V\left(x^0,\nu-\Delta\right)v|\leq K_1(v)(1+\Delta).
\] 
Furthermore, the Lipschitz constants are uniformly bounded, i.e., $\sup_{v\in\reals^{\kappa_1}}K^2_1(v)<\infty$. 
\end{assumption}
This assumption is mild because the expected utility objective function implies that $L(\theta,\nu,\lambda)$ is Lipschitz in $\nu$, and $\phi^\top_V\left(x^0,\nu\right)v$ is just a linear function approximation of $V^\theta(x^0,\nu)$.
Then, we establish the bias and convergence of stochastic sub-gradient estimates. Let
\[
\overline g(\nu_{k})\in\arg\max\left\{g:g\in\partial_{\nu}  L(\theta,\nu,\lambda)\vert_{\nu=\nu_{k}}\right\} 
\]
and
 \[
 \begin{split}
  \Lambda_{1,k+1}=&\left(
 \frac{\left(\phi^\top\left(x^0,\nu_{k}+\Delta_k\right)- \phi^\top\left(x^0,\nu_{k}-\Delta_k\right)\right)v^*(\nu_{k})}{2\Delta_k}-E_M(k)\right),\\
 \Lambda_{2,k}=&\lambda_k+E^L_M(k)-\overline g(\nu_{k}),\\
 \Lambda_{3,k}=&E_M(k)-E^L_M(k),\\
 \end{split}
 \]
 where
\[
\begin{split}
E_M(k):=&\mathbb E\left[\frac{1}{2\Delta_k}\left(\phi^\top\left(x^0,\nu_{k}+\Delta_k\right)- \phi^\top\left(x^0,\nu_{k}-\Delta_k\right)\right)v^*(\nu_{k})\mid \Delta_k\right]\\
E^L_M(k):=&\mathbb E\left[\frac{1}{2\Delta_k}\left(V^{\theta}\left(x^0,\nu_{k}+\Delta_k\right)- V^{\theta}\left(x^0,\nu_{k}-\Delta_k\right)\right)\mid \Delta_k\right].
\end{split}
\]
Note that \eqref{eq:s_grad_est} is equivalent to
 \begin{equation}\label{eq:s_noisy_grad}
 \nu_{k+1}=\nu_{k}-\zeta_3(k)\left(
 \overline g(\nu_{k})+ \Lambda_{1,k+1}+\Lambda_{2,k}+\Lambda_{3,k}\right)
 \end{equation}
First, it is obvious that $ \Lambda_{1,k+1}$ is a Martingale difference as $\mathbb E[ \Lambda_{1,k+1}\mid \mathcal F_k]=0$, which implies 
\[
M_{k+1}=\sum_{j=0}^k \zeta_{3}(j) \Lambda_{1,j+1}
\]
is a Martingale with respect to filtration $\mathcal F_k$. By Martingale convergence theorem, we can show that if $\sup_{k\geq 0} \mathbb E[M^2_k]<\infty$, when $k\rightarrow\infty$, $M_k$ converges almost surely and $\zeta_3(k) \Lambda_{1,k+1}\rightarrow 0$ almost surely. To show that $\sup_{k\geq 0} \mathbb E[M^2_k]<\infty$, for any $t\geq 0$ one observes that,
\[
\begin{split}
&\mathbb E[M^2_{k+1}]= \sum_{j=0}^k\left(\zeta_{3}(j)\right)^2\mathbb E[\mathbb E[ \Lambda_{1,j+1}^2\mid \Delta_j]]\\
\leq & 2 \sum_{j=0}^k\mathbb E\bigg[\left(\frac{\zeta_{3}(j)}{2\Delta_j}\right)^2\left\{\mathbb E\left[\big(\left(
 {\phi^\top\left(x^0,\nu_j+\Delta_j\right)- \phi^\top\left(x^0,\nu_j-\Delta_j\right)\big)v^*(\nu_j)}\right)^2\mid  \Delta_j\right]\right.\\
 &\quad\quad\quad\quad\quad\quad\quad\quad\quad\left.+\mathbb E\left[\big(\phi^\top\left(x^0,\nu_j+\Delta_j\right)- \phi^\top\left(x^0,\nu_j-\Delta_j\right)\big)v^*(\nu_j)\mid \Delta_j\right]^2\right\}\bigg]
 \end{split}
\]
Now based on Assumption \ref{assume_lip}, the above expression implies
\[\begin{split}
\mathbb E[M^2_{k+1}]\leq & 2 \sum_{j=0}^k\mathbb E\left[\left(\frac{\zeta_{3}(j)}{2\Delta_j}\right)^22K_1^2(1+\Delta_j)^2\right]
\end{split}\]
Combining the above results with the step length conditions, there exists $K=4K_1^2>0$ such that
\[
\sup_{k\geq 0}\mathbb E[M^2_{k+1}] \leq  K \sum_{j=0}^\infty\mathbb E\left[ \left(\frac{\zeta_{3}(j)}{2\Delta_j}\right)^2 \right]+\left(\zeta_{2}(j)\right)^2<\infty.
\]

Second, by the ``Min Common/Max Crossing" theorem, one can show $\partial_{\nu} L(\theta,\nu,\lambda)\vert_{\nu=\nu_{k}}$ is a non-empty, convex and compact set. Therefore, by duality of directional directives and sub-differentials, i.e.,
\[
\max \left\{g:g\in\partial_{\nu}  L(\theta,\nu,\lambda)\vert_{\nu=\nu_{k}}\right\} = \lim_{\xi\downarrow 0}\frac{ L(\theta,\nu_k+\xi,\lambda)-L(\theta,\nu_k-\xi,\lambda)}{2\xi},
\]
one concludes that for $\lambda_k=\lambda$ (converges in a slower time scale),
\[
\lambda+E^L_M(k)= \overline g(\nu_{k})+O(\Delta_k), \text{ almost surely}.
\] 
This further implies that
\[
\Lambda_{2,k}= O(\Delta_k),\,\,\text{i.e., } \Lambda_{2,k}\rightarrow 0 \text{ as } k\rightarrow \infty, \text{ almost surely}.
\]
Third, since $d^\theta_\gamma(x^0,\nu|x^0,\nu)=1$, from definition of $\epsilon_{\theta}(v^*(\nu_{k}))$ it is obvious that $ |\Lambda_{3,k}|\leq 2 \epsilon_{\theta}(v^*(\nu_{k}))\mathbb E[1/\Delta_k]$. When $t$ goes to infinity, $\epsilon_{\theta}(v^*(\nu_{k}))\mathbb E[1/\Delta_k]\rightarrow 0$ by assumption and $\Lambda_{3,k}\rightarrow 0$. 
Finally, as we have just showed that $\zeta_2(k) \Lambda_{1,k+1}\rightarrow 0$, $\Lambda_{2,k}\rightarrow 0$ and $\Lambda_{3,k}\rightarrow 0$ almost surely, the $\nu-$update in \eqref{eq:s_noisy_grad} is a stochastic approximations of an element in the differential inclusion

Now we turn to the convergence analysis of $\nu$. It can be easily seen that the $\nu-$update in \eqref{nu_up_incre_SPSA} is a noisy sub-gradient descent update with vanishing disturbance bias. This update can be viewed as an Euler discretization of the following differential inclusion
 \begin{equation}\label{dyn_sys_s_incre_SPSA}
\dot{\nu}\in \Upsilon_{\nu}\left[-g(\nu)\right], \quad\quad \forall g(\nu)\in\partial_\nu L(\theta,\nu,\lambda),
 \end{equation}  
 Thus, the $\nu-$convergence analysis follows from analogous convergence analysis in step 1 of Theorem \ref{thm:converge_h}'s proof.

\noindent\paragraph{Step 3 (Convergence of $\theta-$update)}
We first analyze the actor update ($\theta-$update).
Since $\theta$ converges in a faster time scale than $\lambda$, one can assume $\lambda$ in the $\theta-$update as a fixed quantity. Furthermore, since $v$ and $\nu$ converge in a faster scale than $\theta$, one can also replace $v$ and $\nu$ with their limits $v^{*}(\theta)$ and $\nu^{*}(\theta)$ in the convergence analysis. In the following analysis, we assume that the initial state $x^0\in\X$ is given. Then the $\theta-$update in \eqref{theta_up_incre} can be re-written as follows:
\begin{equation}\label{theta_up}
\theta_{k+1}=\Gamma_{\Theta}\left(\theta_k-\zeta_2(k) \left(\nabla_\theta \log\mu(a_k|x_k,s_k;\theta)\vert_{\theta=\theta_k}\frac{\delta_k(v^*(\theta_k))}{1-\gamma}\right)\right).
\end{equation}
Similar to the trajectory based algorithm, we need to show that
the approximation of $\nabla_\theta L(\theta,\nu,\lambda)$ is Lipschitz in $\theta$ in order to show the convergence of the $\theta$ parameter. This result is generalized in the following proposition.
\begin{proposition}
The following function is a Lipschitz function in $\theta$:
\[
\begin{split}
\frac{1}{1-\gamma}\sum_{x,a,s}&\pi^\theta_\gamma(x,s,a|x_0=x^0,s_0=\nu)\nabla_{\theta}\log\mu(a|x,s;\theta)\\
&\left(-v^\top\phi(x,s)+\gamma\sum_{x^\prime,s^\prime} \bar{P}(x^\prime,s^\prime|x,s,a) v^\top\phi(x^\prime,s^\prime)+\bar{C}(x,s,a)\right).
\end{split}
\]
\end{proposition}
\begin{prooff}
First consider the feature vector $v$. Recall that the feature vector satisfies the linear equation $Av=b$ where $A$ and $b$ are functions of $\theta$ found from the Hilbert space projection of Bellman operator. It has been shown in Lemma 1 of \cite{bhatnagar2012online} that, by exploiting the inverse of $A$ using Cramer's rule, one can show that $v$ is continuously differentiable of $\theta$. Next, consider the $\gamma-$ visiting distribution $\pi^\theta_\gamma$. From an application of Theorem 2 of \cite{altman2004perturbation} (or Theorem 3.1 of \cite{shardlow2000perturbation}), it can be seen that
the stationary distribution $\pi^\theta_\gamma$ of the process $(x_k,s_k)$ is continuously
differentiable in $\theta$. Recall from Assumption (B1) that $\nabla_\theta\mu(a_k|x_k,s_k;\theta)$ is a Lipschitz function in $\theta$ for any $a\in\A$ and $k\in\{0,\ldots,T-1\}$ and $\mu(a_k|x_k,s_k;\theta)$ is differentiable in $\theta$. Therefore,  by combining these arguments and noting that the sum of products of Lipschitz functions is Lipschitz, one concludes that $\nabla_\theta L(\theta,\nu,\lambda)$ is Lipschitz in $\theta$.
\end{prooff}

Consider the case in which the value function for a fixed policy $\mu$ is approximated by a learned function approximator, $\phi^\top(x,s)v^*$. If the approximation is sufficiently good, we might hope to
use it in place of $V^\theta(x,s)$ and still point roughly in the direction of the
true gradient. Recall the temporal difference error (random variable) for given $(x_k,s_k)\in\X\times\reals$
\[
\delta_k\left(v\right)=-v^\top\phi(x_k,s_k)+ \gamma v^\top\phi\left(x_{k+1},s_{k+1}\right)+ \bar{C}(x_k,s_k,a_k).
\] 

Define the $v-$dependent approximated advantage function 
\[
\tilde A^{\theta,v}(x,s,a)=\tilde \Q^{\theta,v}(x,s,a)-v^\top\phi(x,s),
\]
where
\[
\tilde \Q^{\theta,v}(x,s,a)=\gamma\sum_{x^\prime,s^\prime} \bar{P}(x^\prime,s^\prime|x,s,a) v^\top\phi(x^\prime,s^\prime)+\bar{C}(x,s,a).
\]
The following Lemma first shows that $\delta_k(v)$ is an unbiased estimator of $\tilde A^{\theta,v}$.
\begin{lemma}\label{lem:unbiased_L}
For any given policy $\mu$ and $v\in\reals^{\kappa_2}$, we have 
\[
\tilde A^{\theta,v}(x,s,a)=\mathbb E[\delta_k(v)\mid x_k=x,s_k=s,a_k=a].
\]
\end{lemma}
\begin{prooff}
Note that for any $v\in\reals^{\kappa_2}$,
\[
\mathbb E[\delta_k(v)\mid x_k=x,s_k=s,a_k=a,\mu]=\bar{C}(x,s,a)-v^\top\phi(x,s)+\gamma\mathbb E\left[v^\top\phi(x_{k+1},s_{k+1})\mid x_k=x,s_k=s,a_k=a\right].
\]
where
\[
\begin{split}
\mathbb E\left[v^\top\phi(x_{k+1},s_{k+1})\mid x_k=x,s_k=s,a_k=a\right]=\sum_{x^\prime,s^\prime} \bar{P}(x^\prime,s^\prime|x,s,a) v^\top\phi(x^\prime,s^\prime).
\end{split}
\]
By recalling the definition of $\tilde \Q^{\theta,v}(x,s,a)$, the proof is completed.
\end{prooff}
Now, we turn to the convergence proof of $\theta$. 
\begin{theorem}\label{thm:converge}
Suppose $\theta^*$ is the equilibrium point of the continuous system $\theta$ satisfying 
\begin{equation}\label{eq_cond_1}
\Upsilon_\theta\left[ -\nabla_\theta  L(\theta,\nu,\lambda) \vert_{\nu=\nu^*(\theta)}\right]=0.
\end{equation}
Then the sequence of $\theta-$updates in \eqref{theta_up_incre} converges to $\theta^*$ almost surely.
\end{theorem}
\begin{prooff}
First, the $\theta-$update from (\ref{theta_up}) can be re-written as follows:
\[
\theta_{k+1}=\Gamma_{\Theta}\left(\theta_k+\zeta_2(k)\left(-\nabla_\theta L(\theta,\nu,\lambda)\vert_{\nu=\nu^*(\theta),\theta=\theta_k}+\delta\theta_{k+1}+\delta\theta_\epsilon\right)\right)
\]
where
\begin{equation}\label{eq:MG_diff_theta}
\small
\begin{split}
\delta\theta_{k+1}=&\sum_{x^\prime,a^\prime,s^\prime} \pi_{\gamma}^{\theta_k}(x^\prime,s^\prime,a^\prime|x_0=x^0,s_0=\nu^*(\theta_k))\nabla_\theta\log\mu(a^\prime|x^\prime,s^\prime;\theta)\vert_{\theta=\theta_k}\frac{\tilde A^{\theta_k,v^*(\theta_k)}(x^\prime,s^\prime,a^\prime)}{1-\gamma}\\
&-\nabla_\theta\log\mu(a_k|x_k,s_k;\theta)\vert_{\theta=\theta_k}\frac{\delta_k(v^*(\theta_k))}{1-\gamma}.
\end{split}
\end{equation}
is a square integrable stochastic term of the $\theta-$update and
\[\small
\begin{split}
\delta\theta_\epsilon=&\sum_{x^\prime,a^\prime,s^\prime} \pi_{\gamma}^{\theta_k}(x^\prime,s^\prime,a^\prime|x_0=x^0,s_0=\nu^*(\theta_k))\frac{\nabla_{\theta}\log\mu(a^\prime|x^\prime,s^\prime;\theta)\vert_{\theta=\theta_k}}{1-\gamma}(A^{\theta_k}(x^\prime,s^\prime,a^\prime)-\tilde A^{\theta_k,v^*(\theta_k)}(x^\prime,s^\prime,a^\prime))\\
\leq &\frac{\|\psi_{\theta_k}\|_\infty}{1-\gamma}\sqrt{\left(\frac{1+\gamma}{1-\gamma}\right)\epsilon_{\theta_k}(v^*(\theta_k))}.
\end{split}
\]
where
$\psi_\theta(x,s,a)=\nabla_{\theta}\log\mu(a|x,s;\theta)$ is the ``compatible feature". The last inequality is due to the fact that for  $\pi^\theta_\gamma$ being a probability measure, convexity of quadratic functions implies
\[\small
\begin{split}
&\sum_{x^\prime,a^\prime,s^\prime} \pi_{\gamma}^\theta(x^\prime,s^\prime,a^\prime|x_0=x^0,s_0=\nu^*(\theta))(A^\theta(x^\prime,s^\prime,a^\prime)-\tilde A^{\theta,v}(x^\prime,s^\prime,a^\prime))\\
\leq &\sum_{x^\prime,a^\prime,s^\prime} \pi_{\gamma}^\theta(x^\prime,s^\prime,a^\prime|x_0=x^0,s_0=\nu^*(\theta))( Q^\theta(x^\prime,s^\prime,a^\prime)-\tilde \Q^{\theta,v}(x^\prime,s^\prime,a^\prime))\\
&+\sum_{x^\prime,s^\prime} d_{\gamma}^\theta(x^\prime,s^\prime|x_0=x^0,s_0=\nu^*(\theta))(V^\theta(x^\prime,s^\prime)-\widetilde{V}^{\theta,v}(x^\prime,s^\prime))\\
= &\gamma\sum_{x^\prime,a^\prime,s^\prime} \pi_{\gamma}^\theta(x^\prime,s^\prime,a^\prime|x_0=x^0,s_0=\nu^*(\theta))\sum_{x^{\prime\prime},s^{\prime\prime}}\bar{P}(x^{\prime\prime},s^{\prime\prime}|x^\prime,s^\prime,a^\prime) (V^\theta(x^{\prime\prime},s^{\prime\prime})-\phi^\top(x^{\prime\prime},s^{\prime\prime})v)\\
&+\sqrt{\sum_{x^\prime,s^\prime} d_{\gamma}^\theta(x^\prime,s^\prime|x_0=x^0,s_0=\nu^*(\theta))(V^\theta(x^\prime,s^\prime)-\widetilde{V}^{\theta,v}(x^\prime,s^\prime))^2}\\
\leq &\gamma\sqrt{\sum_{x^\prime,a^\prime,s^\prime} \pi_{\gamma}^\theta(x^\prime,s^\prime,a^\prime|x_0=x^0,s_0=\nu^*(\theta))\sum_{x^{\prime\prime},s^{\prime\prime}} \bar{P}(x^{\prime\prime},s^{\prime\prime}|x^\prime,s^\prime,a^\prime) (V^\theta(x^{\prime\prime},s^{\prime\prime})-\phi^\top(x^{\prime\prime},s^{\prime\prime})v)^2}\\
&+\sqrt{\frac{\epsilon_\theta(v)}{1-\gamma}}\\
\leq &\sqrt{\gamma}\sqrt{\sum_{x^{\prime\prime},s^{\prime\prime}}\left(d_{\gamma}^\theta(x^{\prime\prime},s^{\prime\prime}|x^0,\nu^*(\theta))-(1-\gamma)1\{x^0=x^{\prime\prime},\nu=s^{\prime\prime}\} \right)(V^\theta(x^{\prime\prime},s^{\prime\prime})-\phi^\top(x^{\prime\prime},s^{\prime\prime})v)^2}+\sqrt{\frac{\epsilon_\theta(v)}{1-\gamma}}\\
\leq &\sqrt{\left(\frac{1+\gamma}{1-\gamma}\right)\epsilon_\theta(v)}
\end{split}
\]

Then by Lemma \ref{lem:unbiased_L}, if the $\gamma-$stationary distribution $\pi_\gamma^\theta$ is used to generate samples of $(x_k,s_k,a_k)$, one obtains $\mathbb E\left[\delta\theta_{k+1}\mid \mathcal F_{\theta,k}\right]=0$, where $\mathcal F_{\theta,k}= \sigma(\theta_m,\,\delta \theta_m,\,m\leq k)$ is the filtration generated by different independent trajectories. On the other hand, $|\delta\theta_\epsilon|\rightarrow 0$ as $\epsilon_{\theta_k}(v^*(\theta_k))\rightarrow 0$.
Therefore, the $\theta-$update in \eqref{theta_up} is a stochastic approximation of the ODE 
\[
\dot{\theta}=\Upsilon_\theta\left[ -\nabla_\theta L(\theta,\nu,\lambda)\vert_{\nu=\nu^*(\theta)} \right]
\] 
with an error term that is a sum of a vanishing bias and a Martingale difference. Thus, the convergence analysis of $\theta$ follows analogously from the step 2 of Theorem \ref{thm:converge_h}'s proof.
\end{prooff}
\noindent\paragraph{Step 4 (Local Minimum)}
The proof of local minimum of $(\theta^\ast,\nu^\ast)$ follows directly from the arguments  in Step 3 of Theorem \ref{thm:converge_h}'s proof.

\noindent\paragraph{Step 5 (The $\lambda-$update and Convergence to Saddle Point)}
Notice that $\lambda-$update converges in a slowest time scale, \eqref{nu_up_incre_SPSA} can be rewritten using the converged $v^*(\lambda)$, $\theta^*(\lambda)$ and $\nu^*(\lambda)$, i.e.,
\begin{equation}\label{lamda_up_conv}
\lambda_{k+1}=\Gamma_\Lambda\left(\lambda_k+\zeta_1(k)\left(\nabla_\lambda L(\theta,\nu,\lambda)\bigg\vert_{\theta=\theta^*(\lambda),\nu=\nu^*(\lambda),\lambda=\lambda_k}+\delta\lambda_{k+1}\right)\right)
\end{equation}
where 
\begin{equation}\label{eq:MG_diff_theta}
\delta\lambda_{k+1}=-\nabla_\lambda L(\theta,\nu,\lambda)\bigg\vert_{\theta=\theta^*(\lambda),\nu=\nu^*(\lambda),\lambda=\lambda_k}+
\left(\nu^*(\lambda_k)+\frac{(-s_k)^+}{(1-\alpha)(1-\gamma)}\mathbf 1\{x_k=x_T\} -\beta\right)
\end{equation}
is a square integrable stochastic term of the $\lambda-$update. Similar to the $\theta-$update, by using the $\gamma-$stationary distribution $\pi_{\gamma}^\theta$, one obtains $\mathbb E\left[\delta\lambda_{k+1}\mid \mathcal F_{\lambda,k}\right]=0$ where $\mathcal F_{\lambda,k}= \sigma(\lambda_m,\,\delta \lambda_m,\,m\leq k)$ is the filtration of $\lambda$ generated by different independent trajectories. As above, the $\lambda-$update is a stochastic approximation of the ODE
\[
\dot{\lambda}=\Upsilon_\lambda\left[ \nabla_\lambda L(\theta,\nu,\lambda)\bigg\vert_{\theta=\theta^*(\lambda),\nu=\nu^*(\lambda)}\right]
\]
with an error term that is a Martingale difference. 
Then the $\lambda-$convergence and the (local) saddle point analysis follows from analogous  arguments in step 4 and 5 of Theorem \ref{thm:converge_h}'s proof.

\noindent\paragraph{Step $2^\prime$ (Convergence of Multi-loop $\nu-$update)}
Since $\nu$ converges on a faster timescale than $\theta$ and $\lambda$, the $\nu-$update
in \eqref{nu_up_incre_SPSA_semi} can be rewritten using the fixed $(\theta,\lambda)$, i.e.,
\begin{equation}\label{update_s_multi}
\nu_{i+1}=\Gamma_N\left(\nu_i-\zeta_2(i)\left(\lambda-\frac{\lambda}{1-\alpha}\left(\mathbb P\left(s_{T}\leq 0\mid x_0=x^0,s_0=\nu_i,\mu\right)+\delta \nu_{M,i+1}\right)\right)\right)
\end{equation}
and 
\begin{equation}\label{eq:MG_diff_s_multi}
\delta \nu_{M,i+1}=-\mathbb P\left(s_T\leq 0\mid x_0=x^0,s_0=\nu_i,\mu\right)+\mathbf 1\left\{s_T\leq 0\right\}
\end{equation}
is a square integrable stochastic term of the $\nu-$update. 
It is obvious that $\mathbb E\left[\delta \nu_{M,i+1}\mid \mathcal F_{\nu,i}\right]=0$, where $\mathcal F_{\nu,i}= \sigma(\nu_m,\,\delta\nu_m,\,m\leq i)$ is the corresponding filtration of $\nu$, the $\nu-$update in \eqref{nu_up_incre_SPSA_semi} is a stochastic approximations of an element in the differential inclusion $\partial_{\nu}  L(\theta,\nu,\lambda)\vert_{\nu=\nu_i}$ for any $i$ with an error term that is a Martingale difference, i.e.,
\[
\frac{\lambda}{1-\alpha}\mathbb P\left(s_T\leq 0\mid x_0=x^0,s_0=\nu_i,\mu\right)-\lambda\in-\partial_\nu  L(\theta,\nu,\lambda)\vert_{\nu=\nu_i}.
\]
Thus, the $\nu-$update in \eqref{update_s_multi} can be viewed as an Euler discretization of the differential inclusion in \eqref{dyn_sys_s_incre_SPSA}, and the $\nu-$convergence analysis follows from analogous convergence analysis in step 1 of Theorem \ref{thm:converge_h}'s proof.

%%%%%%%%%%%%%%%%%%%%%%%%%%%%%%%%%%%%%%%%%%%%%%%%%%%%%%%%%%%
%%%%%%%%%%%%%%%%%%%%%%%%%%%%%%%%%%%%%%%%%%%%%%%%%%%%%%%%%%%
%%%%%%%%%%%%%%%%%%%%%%%%%%%%%%%%%%%%%%%%%%%%%%%%%%%%%%%%%%%
%%%%%%%%%%%%%%%%%%%%%%%%%%%%%%%%%%%%%%%%%%%%%%%%%%%%%%%%%%%
%%%%%%%%%%%%%%%%%%%%%%%%%%%%%%%%%%%%%%%%%%%%%%%%%%%%%%%%%%%

\newpage
\section{Experimental Results}
\label{sec:appendix_experiment}

%%%%%%%%%%%%%%%%%%%%%%%%%%%%%%%%%%%%%%%%%%%%%%%%%%%%%%%%%%%
%%%%%%%%%%%%%%%%%%%%%%%%%%%%%%%%%%%%%%%%%%%%%%%%%%%%%%%%%%%
%%%%%%%%%%%%%%%%%%%%%%%%%%%%%%%%%%%%%%%%%%%%%%%%%%%%%%%%%%%

\subsection{Problem Setup and Parameters}
The house purchasing problem can be reformulated as follows
\begin{equation}
\label{eq:norm_reward_eqn_example}
\min_\theta \mathbb E\left[D^\theta(x^0)\right]\quad\quad \text{subject to} \quad\quad \text{CVaR}_\alpha\big(D^\theta(x^0\big)\leq\beta.
\end{equation}
where $D^\theta(x^0)=\sum_{k=0}^T\gamma^k\left(\mathbf 1\{u_k=1\}c_k+\mathbf 1\{u_k=0\}p_h\right)\mid x_0=x,\;\mu$. We will set the parameters of the MDP as follows: $x_0=[1;0]$, $p_h=0.1$, $T = 20$, $\gamma=0.95$, $f_u=1.5$, $f_d=0.8$ and $p=0.65$. For the risk constrained policy gradient algorithm, the step-length sequence is given as follows,
\[
\zeta_1(i)=\frac{0.1}{i},\,\,\zeta_2(i)=\frac{0.05}{i^{0.8}},\,\, \zeta_3(i)=\frac{0.01}{i^{0.55}},\,\,\forall i.
\]
The CVaR parameter and constraint threshold are given by $\alpha=0.9$ and $\beta=1.9$. The number of sample trajectories $N$ is set to $100$. 

For the risk constrained actor critic algorithm, the step-length sequence is given as follows,
\[
\zeta_1(i)=\frac{1}{i},\,\,\zeta_2(i)=\frac{1}{i^{0.85}},\,\, \zeta_3(i)=\frac{0.5}{i^{0.7}},\,\,\zeta_3(i)=\frac{0.5}{i^{0.55}},\,\,\Delta_k=\frac{0.5}{i^{0.1}},\,\,\forall i.
\]
The CVaR parameter and constraint threshold are given by $\alpha=0.9$ and $\beta=2.5$. One can later see that the difference in risk thresholds is due to the different family of parametrized Boltzmann policies. 

The parameter bounds are given as follows: $\lambda_{\max}=1000$, $\Theta=[-60,60]^{\kappa_1}$ and $C_{\max}=4000>x_0\times f_u^{T}$.

%%%%%%%%%%%%%%%%%%%%%%%%%%%%%%%%%%%%%%%%%%%%%%%%%%%%%%%%%%%
%%%%%%%%%%%%%%%%%%%%%%%%%%%%%%%%%%%%%%%%%%%%%%%%%%%%%%%%%%%
%%%%%%%%%%%%%%%%%%%%%%%%%%%%%%%%%%%%%%%%%%%%%%%%%%%%%%%%%%%

\subsection{Trajectory Based Algorithms}

In this section, we have implemented the following trajectory based algorithms.
\begin{enumerate}
\item \textbf{PG:} This is a policy gradient algorithm that minimizes the expected discounted cost function, without considering any risk criteria.
\item \textbf{PG-CVaR:} This is the CVaR constrained simulated trajectory based policy gradient algorithm that is given in Section 4. 
\end{enumerate}
It is well known that a near-optimal policy $\mu$ was obtained using the LSPI algorithm  with 2-dimensional radial basis function
(RBF) features. We will also implement the 2-dimensional RBF feature function $\phi$ and consider the family Boltzmann policies for policy parametrization
\[
\mu(a|x;\theta)=\frac{\exp(\theta^\top\phi(x,a))}{\sum_{a^\prime\in\A}\exp(\theta^\top\phi(x,a^\prime))}.
\] 
The experiments for each algorithm comprised of the following two phases:
\begin{enumerate}
\item \textbf{Tuning phase:} Here each iteration involved the simulation run with the nominal policy parameter $\theta$ where the run length for a particular policy parameter is at most $T$ steps. We run the algorithm for 1000 iterations and stop when the parameter $(\theta,\nu,\lambda)$ converges.
\item \textbf{Converged run:} Followed by the tuning phase, we obtained the converged policy parameter $\theta^*$. In the converged run phase, we perform simulation with this policy parameter for $1000$ runs where each simulation generates a trajectory of at most $T$ steps. The results
reported are averages over these iterations.
\end{enumerate}

\subsection{Incremental Based Algorithm}
On the other hand, we have also implemented the following incremental based algorithms.
\begin{enumerate}
\item \textbf{AC:} This is an actor critic algorithm that minimizes the expected discounted cost function, without considering any risk criteria. This is similar to Algorithm 1 in \cite{Bhatnagar10AC}.
\item \textbf{AC-CVaR-Semi-Traj.:} This is the CVaR constrained multi-loop actor critic algorithm that is given in Section 5. 
\item \textbf{AC-CVaR-SPSA:} This is the CVaR constrained SPSA actor critic algorithm that is given in Section 5. 
\end{enumerate}
Similar to the trajectory based algorithms, we will implement the RBFs as feature functions for $[x;s]$ and consider the family of augmented state Boltzmann policies,
\[
\mu(a|(x,s);\theta)=\frac{\exp(\theta^\top\phi(x,s,a))}{\sum_{a^\prime\in\A}\exp(\theta^\top\phi(x,s,a^\prime))}.
\] 
Similarly, the experiments also comprise of two phases: 1) the tuning phase where the set of parameters $(v,\theta,\nu,\lambda)$ is obtained after the algorithm converges, and 2) the converged run where the policy parameter is simulated for $1000$ runs.

%%%%%%%%%%%%%%%%%%%%%%%%%%%%%%%%%%%%%%%%%%%%%%%%%%%%%%%%%%%
%%%%%%%%%%%%%%%%%%%%%%%%%%%%%%%%%%%%%%%%%%%%%%%%%%%%%%%%%%%
\newpage
\section{Bellman Equation and Projected Bellman Equation for Expected Utility Function}
\subsection{Bellman Operator for Expected Utility Functions}
First, we want find the Bellman equation for the objective function 
\begin{equation}\label{eq:bellman_obj}
\expec\left[D^\theta(x_0)\mid x_0=x^0,s_0=s^0,\mu\right]+\frac{\lambda}{1-\alpha}\expec\left[\left[D^\theta(x_0)- s_0\right]^+\mid x_0=x^0,s_0=s^0,\mu\right]
\end{equation}
where $\lambda$ and $(x^0,s^0)\in\X\times\reals$ are given. 

For any function $V:\X\times\reals\rightarrow\reals$, recall the following Bellman operator on the augmented space $\X\times\reals$:
\[
T_{\theta}[V](x,s):=\sum_{a\in \A}\mu(a|x,s;\theta)\left\{\bar{C}(x,s,a)+\sum_{x^\prime,s^\prime}\gamma\bar{P}(x^\prime,s^\prime|x,s,a)\V\left(x^\prime,s^\prime\right)\right\}.
\]
First, it is easy to show that this Bellman operator satisfies the following properties.
\begin{proposition}
The Bellman operator $T_{\theta}[\V]$ has the following properties:
\begin{itemize}
\item (Monotonicity) If $\V_1(x,s)\geq \V_2(x,s)$, for any $x\in\X$, $s\in\reals$, then $T_{\theta}[\V_1](x,s)\geq T_{\theta}[\V_2](x,s)$.
\item (Constant shift) For $K\in\reals$, $T_{\theta}[\V+K](x,s)=T_{\theta}[\V](x,s)+\gamma K$.
\item (Contraction) 
\[
\|T_{\theta}[\V_1]-T_{\theta}[\V_2]\|_{\infty}\leq \gamma \|\V_1-\V_2\|_{\infty},
\]
where $\|f\|_{\infty}=\max_{x\in\X,s\in\reals} |f(x,s)|$.
\end{itemize}
\end{proposition}
\begin{prooff}
The proof of monotonicity and constant shift properties follow directly from the definitions of the Bellman operator.
Furthermore, denote $c=\|\V_1-\V_2\|_{\infty}$. Since
\[
\V_2(x,s)-\|\V_1-\V_2\|_{\infty}\leq \V_1(x,s)\leq \V_2(x,s)+\|\V_1-\V_2\|_{\infty},\,\, \forall x\in\X, \,s\in\reals,
\]
by monotonicity and constant shift property,
\[
T_{\theta}[\V_2](x,s)-\gamma\|\V_1-\V_2\|_{\infty}\leq T_\theta[\V_1](x,s) \leq T_{\theta}[\V_2](x,s)+\gamma\|\V_1-\V_2\|_{\infty}\,\, \forall x\in\X, \,s\in\reals.
\]
This further implies that 
\[
|T_{\theta}[\V_1](x,s)-T_{\theta}[\V_2](x,s)|\leq \gamma\|\V_1-\V_2\|_{\infty}\,\, \forall x\in\X, \,s\in\reals
\]
and the contraction property follows.
\end{prooff}

The following theorems show there exists a unique fixed point solution to $T_{\theta}[V](x,s)=V(x,s)$, where the solution equals to the value function expected utility. 
\begin{theorem}[Equivalence Condition]\label{thm:opt}
For any bounded function $V_0:\X\times\reals\rightarrow\reals$,  there exists a limit function $\V^\theta$ such that $\V^\theta(x,s)=\lim_{N\rightarrow\infty} T_{\theta}^N[V_{0}](x,s)$. Furthermore, 
\[
 \V^\theta(x^0,s^0)=\expec\left[D^\theta(x_0)\mid x_0=x^0,\mu\right]+\frac{\lambda}{1-\alpha}\expec\left[\left[D^\theta(x_0)- s_0\right]^+\mid x_0=x^0,s_0=s^0,\mu\right].
\] 
\end{theorem}
\begin{prooff}
The first part of the proof is to show that for any $x\in\X$ and $s\in\reals$,
\begin{equation}\label{induction_Vn}
  V_n(x,s):=T_{\theta}^{n}[V_{0}](x^0,s^0)=\expec\left[\sum_{k=0}^{n-1} \gamma^k \bar{C}(x_k,s_k,a_k)+\gamma^nV_0(x_n,s_n)\mid x_0=x,s_0=s,\,\mu\right]
\end{equation}
by induction. For $n=1$, $V_1(x,s)=T_{\theta}[\V_{0}](x,s)=\expec\left[ \bar{C}(x_0,s_0,a_0)+\gamma V_0(x_1,s_1)\mid x_0=x,s_0=s,\,\mu\right]$.
By induction hypothesis, assume \eqref{induction_Vn} holds at $n=k$. For $n=k+1$, 
\[\small
\begin{split}
V_{k+1}(x,s):=&T_{\theta}^{k+1}[V_{0}](x,s)=T_{\theta}[V_k](x,s)\\
=&\sum_{a\in \bar\A}\mu(a|x,s;\theta)\left\{\bar{C}(x,s,a)+\sum_{x^\prime,s^\prime}\gamma\bar{P}(x^\prime,s^\prime|x,s,a)\V_k\left(x^\prime,s^\prime\right)\right\}\\
=& \sum_{a\in \bar\A}\mu(a|x,s;\theta)\left\{\bar{C}(x,s,a)+\sum_{x^\prime,s^\prime}\gamma\bar{P}(x^\prime,s^\prime|x,s,a)\right.\\
&\left.\quad\quad\expec\left[\sum_{k=0}^{k-1} \gamma^k \bar{C}(x_k,s_k,a_k)+\gamma^kV_0(x_k,s_k)\mid x_0=x^\prime,s_0=s^\prime,\,\,\mu\right]\right\}\\
=& \sum_{a\in \bar\A}\mu(a|x,s;\theta)\left\{\bar{C}(x,s,a)+\sum_{x^\prime,s^\prime}\gamma\bar{P}(x^\prime,s^\prime|x,s,a)\right.\\
&\left.\quad\quad\expec\left[\sum_{t=1}^{k} \gamma^k \bar{C}(x_k,s_k,a_k)+\gamma^kV_0(x_{k+1},s_{k+1})\mid x_1=x^\prime,s_1=s^\prime,\,\,\mu\right]\right\}\\
=&\expec\left[\sum_{k=0}^{k} \gamma^k \bar{C}(x_k,s_k,a_k)+\gamma^{k+1}V_0(x_{k+1},s_{k+1})\mid x_0=x,s_0=s,\,\,\mu\right].
\end{split}
\]
Thus, the equality in \eqref{induction_Vn} is proved by induction. 

The second part of the proof is to show that $\V^\theta(x^0,s^0)
:=\lim_{n\rightarrow\infty} V_{n}(x^0,s^0)$ and
\[
\V^\theta(x^0,s^0)
=\expec\left[D^\theta(x_0)\mid x_0=x^0,\mu\right]+\frac{\lambda}{1-\alpha}\expec\left[\left[D^\theta(x_0)- s_0\right]^+\mid x_0=x^0,s_0=s^0,\mu\right].
\]
From the assumption of transient policies, one note that for any $\epsilon>0$ there exists a sufficiently large $k>N(\epsilon)$ such that $\sum_{t=k}^\infty \mathbb P(x_n=z|x_0,\mu)< \epsilon$ for $z\in\X$. This implies $\mathbb P(T<\infty)>1-\epsilon$.
Since $V_0(x,s)$ is bounded for any $x\in\X$ and $s\in\reals$, the above arguments imply
\[\small
\begin{split}
\V^\theta(x^0,s^0)\leq& \mathbb E \left[\sum_{k=0}^{T-1} \gamma^k\bar{C}(x_k,s_k,a_k)\mid x_0=x^0,s_0=s^0,\mu\right](1-\epsilon)+\epsilon\left(\frac{\lambda}{1-\alpha}(|s^0|+C_{\max})+\frac{C_{\max}}{1-\gamma}\right)\\
&+\lim_{n\rightarrow\infty}\mathbb E \left[\sum_{t=T}^{n-1} \gamma^k\bar{C}(x_k,s_k,a_k)+\gamma^nV_0(x_n,s_n)\mid x_0=x^0,s_0=s^0,\mu\right](1-\epsilon)\\
\leq&\lim_{n\rightarrow\infty} \mathbb E \left[\sum_{k=0}^{T-1} \gamma^kC(x_k,a_k)\mid x_0=x^0,s_0=s^0,\mu\right](1-\epsilon)+\epsilon\left(\frac{1-\epsilon}{\epsilon}\gamma^n\|V_0\|_\infty+\frac{\lambda}{1-\alpha}(|s^0|+C_{\max})+\frac{C_{\max}}{1-\gamma}\right) \\
&+\mathbb E \left[\gamma^{T}\bar{C}(x_{T},s_{T},a_{T})\mid x_0=x^0,s_0=s^0,\mu\right](1-\epsilon)\\
=&\mathbb E \left[D^\theta(x_0)\mid x_0=x^0,s_0=s^0,\mu\right](1-\epsilon) \\
&+\frac{\lambda}{1-\alpha}\mathbb E \left[\gamma^{T}(-s_{T})^+\mid x_0=x^0,s_0=s^0,\mu\right](1-\epsilon)+\epsilon\left(\frac{\lambda}{1-\alpha}(|s^0|+C_{\max})+\frac{C_{\max}}{1-\gamma}\right)\\
=&\mathbb E \left[D^\theta(x_0)\mid x_0=x^0,s_0=s^0,\mu\right](1-\epsilon) \\
&+\frac{\lambda}{1-\alpha}\expec\left[\left[D^\theta(x_0)- s_0\right]^+\mid x_0=x^0,s_0=s^0,\mu\right](1-\epsilon)+\epsilon\left(\frac{\lambda}{1-\alpha}(|s^0|+C_{\max})+\frac{C_{\max}}{1-\gamma}\right).
\end{split}
\]
The first inequality is due to the fact for $x_0=x^0$, $s_0=s^0$,
\[
\lim_{n\rightarrow\infty}\sum_{k=0}^{n} \gamma^k\bar{C}(x_k,s_k,a_k)\leq \frac{\lambda}{1-\alpha}|s^0|+\left(1+\frac{\lambda}{1-\alpha}\right)\sum_{k=0}^\infty \gamma^k|c(x_k,a_k)|\leq\frac{\lambda}{1-\alpha}(|s^0|+C_{\max})+\frac{C_{\max}}{1-\gamma},
\]
the second inequality is due to 1) $V_0$ is bounded, $\bar{C}(x,s,a)=C(x,a)$ when $x\neq x_T$ and 2) for sufficiently large $k>N(\epsilon)$ and any $z\in\X$, 
\[
\sum_{t=k}^\infty \sum_{s}\mathbb P(x_k=z,s_k=s|x_0=x^0,s_0=s^0,\mu)ds=\sum_{t=k}^\infty \mathbb P(x_k=z|x_0=x^0,s_0=s^0,\mu)<\epsilon.
\]
The first equality follows from the definition of transient policies and the second equality follows from the definition of stage-wise cost in the $\nu-$augmented MDP.

By similar arguments, one can also show that
\[
\begin{split}
&\V^\theta(x^0,s^0)\geq  \epsilon\left(-\lim_{n\rightarrow\infty}{(1-\epsilon)}\gamma^n\|V_0\|_\infty/{\epsilon}-{ C_{\max}}/{(1-\gamma)}\right)+(1-\epsilon)\\
&\,\,\left(\mathbb E \left[D^\theta(x_0)\mid x_0=x^0,s_0=s^0,\mu\right]+\frac{\lambda}{1-\alpha}\expec\left[\left[D^\theta(x_0)- s_0\right]^+\mid x_0=x^0,s_0=s^0,\mu\right]\right).
\end{split}
\]
Therefore, by taking $\epsilon\rightarrow 0$, we have just shown that for any $(x^0,s^0)\in\X\times\reals$, $\V^\theta(s^0,s^0)= \expec\left[D^\theta(x_0)\mid x_0=x^0,s_0=s^0,\mu\right]+\lambda/(1-\alpha)\expec\left[\left[D^\theta(x_0)- s_0\right]^+\mid x_0=x^0,s_0=s^0,\mu\right]$.
\end{prooff}
Apart from the analysis in \cite{Bauerle11MD} where a fixed point result is defined based on the following specific set of functions $\mathcal   V^\theta$,
we are going to provide the fixed point theorem for general spaces of augmented value functions.

\begin{theorem}[Fixed Point Theorem]\label{thm:fixed_point}
There exists a unique solution to the fixed point equation: $T_{\theta}[V](x,s)=V(x,s)$, $\forall x\in\X$ and $s\in\reals$. Let $  V^*:\X\times\reals\rightarrow\reals$ be such unique fixed point solution. Then,
\[
  V^*(x,s)=\V^\theta(x,s),\,\forall x\in\X ,\,s\in\real.
\]
\end{theorem}
\begin{prooff}
For $V_{k+1}(x,s)=T_{\theta}[V_k](x,s)$ starting at $V_0:\X\times\reals\rightarrow\reals$ one obtains by contraction that $\|V_{k+1}-V_{k}\|_\infty\leq \gamma\|V_{k}-V_{k-1}\|_\infty$.
By the recursive property, this implies 
\[
\|V_{k+1}-V_{k}\|_\infty\leq \gamma^k\|V_{1}-V_{0}\|_\infty.
\]
It follows that for every $k\geq 0$ and $m\geq 1$,
\[\small
\begin{split}
\|V_{k+m}-V_{k}\|_\infty\leq& \sum_{i=1}^m\|V_{k+i}-V_{k+i-1}\|_\infty\leq \gamma^k(1+\gamma+\ldots+\gamma^{m-1})\|V_{1}-V_{0}\|_\infty\\
\leq & \frac{\gamma^k}{1-\gamma}\|V_{1}-V_{0}\|_\infty.
\end{split}
\]
Therefore, $\{V_k\}$ is a Cauchy sequence and must converge to $V^*$ since $(B(\X\times\reals),\|\cdot\|_\infty)$ is a complete space. Thus, we have for $k\geq 1$,
\[
\|T_{\theta}[V^*]-V^*\|_\infty\leq \|T_{\theta}[V^*]-V_k\|_\infty+\|V_k-V^*\|_\infty\leq \gamma\|V_{k-1}-V^*\|_\infty+\|V_k-V^*\|_\infty.
\]
Since $V_k$ converges to $V^*$, the above expression implies  $T_{\theta}[V^*](x,s)=V^*(x,s)$ for any $(x,s)\in\X\times\reals$. Therefore, $V^*$ is a fixed point. Suppose there exists another fixed point $\tilde V^*$. Then,
\[
\|\tilde V^*-V^*\|_\infty=\|T_{\theta}[\tilde V^\theta]-T_{\theta}[V^\theta]\|_\infty\leq \gamma\|\tilde V^\theta-V^\theta\|_\infty
\]
for $\gamma\in(0,1)$. This implies that $\tilde V^*=V^*$.
Furthermore, since $\V^\theta(x,s)=\lim_{n\rightarrow\infty} T_{\theta}^n[V_{0}](x,s)$ with $V_{0}:\X\times\reals\rightarrow\reals$ being an arbitrary initial value function. By the following convergence rate bound inequality
\[
\|T_{\theta}^k[V_{0}]-  V^*\|_\infty=\|T_{\theta}^k[V_{0}]-T_{\theta}^k[  V^*]\|_\infty\leq \gamma^k\|V_{0}-  V^*\|_\infty,\,\,\gamma\in(0,1),
\] 
one concludes that $\V^\theta(x,s)=   V^*(x,s)$ for any $(x,s)\in\X\times\reals$.
\end{prooff}

\subsection{The Projected Bellman Operator}
Consider the $v-$dependent linear value function approximation of $ \V^\theta(x,s)$, in the form of $\phi^\top(x,s)v$, where $\phi(x,s)\in\reals^{\kappa_2}$ represents the state-dependent feature. The feature vectors can also be dependent on $\theta$ as well. But for notational convenience, we drop the indices corresponding to $\theta$. The low dimensional subspace is therefore 
$S_{V}=\left\{\Phi v|v\in\reals^{\kappa_2}\right\}$ 
 where $\phi:\X\times\reals\rightarrow\reals^{\kappa_2}$ is a function mapping such that $\Phi(x,s)=\phi^\top(x,s)$. We also make the following standard assumption on the rank of matrix $\phi$. More information relating to the feature mappings and function approximation $\phi$ can be found in Appendix. Let $v^*\in\reals^{\kappa_2}$ be the best approximation parameter vector. Then 
 $\tilde V^*(x,s)=(v^*)^\top\phi(x,s)$ is the best linear approximation of $ \V^\theta(x,s)$. 
 
Our goal is to estimate $v^*$ from simulated trajectories of the MDP. Thus, it is reasonable to consider the projections from $\reals$ onto $S_{V}$ with respect to a norm that is weighted according to the occupation measure $d_\gamma^\theta(x^\prime,s^\prime|x,s)$, where $(x^0,s^0)=(x,s)$ is the initial condition of the augmented MDP. For a function $y:\X\times\reals\rightarrow\reals$, we introduce the weighted norm:
$\|y\|_d=\sqrt{\sum_{x,s}d(x^\prime,s^\prime|x,s)(y(x^\prime,s^\prime))^2}$
where $d$ is the occupation measure (with non-negative elements). 
We also denote by $\Pi$ the projection from $\X \times\reals$ to $S_{V}$. We are now ready to describe the approximation scheme. Consider the following projected fixed point equation
\[
V(x,s)=\Pi T_{\theta} [V](x,s)
\]
where $T_{\theta}$ is the Bellman operator with respect to policy $\theta$ and let $\tilde V^*$ denote the solution of the above equation. We will show the existence of this unique fixed point by the following 
contraction property of the projected Bellman operator: $\Pi T_{\theta}$.
\begin{lemma}\label{lem:unique_projected_fixed_pt}
There exists $\kappa\in(0,1)$ such that
\[
\|\Pi T_{\theta} [V_1]-\Pi T_{\theta} [V_2]\|_d\leq \kappa\|V_1-V_2\|_d.
\]
\end{lemma}
\begin{prooff}
Note that the projection operator $\Pi$ is non-expansive:
\[
\|\Pi T_{\theta} [V_1]-\Pi T_{\theta} [V_2]\|_d^2\leq \|T_{\theta} [V_1]- T_{\theta} [V_2]\|_d^2.
\]
One further obtains the following expression:
\[\small
\begin{split}
&\|T_{\theta} [V_1]- T_{\theta} [V_2]\|_d^2\\
=&\sum_{\overline x,\overline s}d(\overline x,\overline s|x,s)\left(\sum_{y,\overline a,s^\prime}\gamma\mu(\overline a|\overline x,\overline s;\theta)\bar{P}(y,s^\prime|\overline x,\overline s,\overline a)(V_1(y,s^\prime)-V_2(y,s^\prime))\right)^2\\
\leq &\sum_{\overline x,\overline s}d(\overline x,\overline s|x,s)\left(\sum_{y,\overline a,s^\prime}\gamma^2\mu(\overline a|\overline x,\overline s;\theta)\bar{P}(y,s^\prime|\overline x,\overline s,\overline a)(V_1(y,s^\prime)-V_2(y,s^\prime))^2\right)\\
=&\sum_{y,s^\prime}\left(d(y,s^\prime|x,s)-(1-\gamma)1\{x=y,s=s^\prime\} \right)\gamma (V_1(y,s^\prime)-V_2(y,s^\prime))^2\\
\leq& \gamma\|V_1-V_2\|_d^2.
\end{split}
\]
The first inequality is due to the fact that $\mu(\overline a|\overline x,\overline s;\theta),\bar{P}(y,s^\prime|\overline x,\overline s,\overline a)\in[0,1]$ and convexity of quadratic function, the second equality is based on the property of $\gamma-$visiting distribution. Thus, we have just shown that $\Pi T_{\theta}$ is contractive with $\kappa=\sqrt{\gamma}\in(0,1)$. 
\end{prooff}

Therefore, by Banach fixed point theorem, a unique fixed point solution exists for equation: $\Pi T_{\theta} [ V](x,s)= V(x,s)$ for any $x\in\X$, $s\in\reals$. Denote by $\tilde V^*$ the fixed point solution and $v^*$ be the corresponding weight, which is unique by the full rank assumption. From Lemma \ref{lem:unique_projected_fixed_pt}, one obtains a unique value function estimates  from the following projected Bellman equation:
\begin{equation}\label{eq:unique_fixed_pt}
\Pi T_{\theta} [\tilde V^*](x,s)=\tilde V^*(x,s),\,\, \tilde V^*(x,s,a)=(v^*)^\top\phi(x,s).
\end{equation}
Also we have the following error bound of the value function approximation.
\begin{lemma}\label{lem:error_bdd}
Let $V^*$ be the fixed point solution of $T_{\theta}[V](x,s)=V(x,s)$ and $v^*$ be the unique solution for $\Pi T_{\theta}[\Phi v](x,s)=\phi^\top(x,s)v$. Then, for some $\kappa\in(0,1)$,
\[
\|V^*-\tilde V^*\|_d=\|V^*-\Phi v^*\|_d\leq\frac{1}{\sqrt{1-\gamma}}\|V^*-\Pi V^*\|_d.
\]
\end{lemma}
\begin{prooff}
Note that by the Pythagorean theorem of projection,
\[\small
\begin{split}
\|V^*-\Phi v^*\|_d^2&=\|V^*-\Pi V^*\|_d^2+\|\Pi V^*-\Phi v^*\|_d^2\\
&=\|V^*-\Pi V^*\|_d^2+\|\Pi T_{\theta}[V^*]-\Pi T_{\theta}[\Phi v^*]\|_d^2\\
&\leq \|V^*-\Pi V^*\|_d^2+\kappa^2\|V^*-\Phi v^*\|_d^2
\end{split}
\]
Therefore, by recalling $\kappa=\sqrt{\gamma}$, the proof is completed by rearranging the above inequality. 
\end{prooff}
This implies that if $V^*\in S_V$, $V^*(x,s)=\tilde V^*(x,s)$ for any $(x,s)\in\X\times\reals$. 

Note that we can re-write the projected Bellman equation in explicit form as follows:
\[\small
\begin{split}
&\Pi T_{\theta} [\Phi v^*]=\Phi v^*\\
\iff & \Pi\left[\left\{\sum_{\overline a\in\A}\mu(\overline a|\overline x,\overline s;\theta)\left(\bar{C}(\overline x,\overline s,\overline a)+\gamma \sum_{y,s^\prime}\bar{P}(y,s^\prime|\overline x,\overline s,\overline a) (v^*)^\top\phi\left(y,s^\prime\right) \right)\right\}_{\overline x\in\X,\overline s\in\reals}\right]=\Phi v^*.
\end{split}
\]
By the definition of projection, the unique solution $v^*\in\reals^\ell$ satisfies 
\[\small
\begin{split}
v^*&\in\arg\min_{v}\|T_{\theta}[\Phi v]-\Phi v\|_{d^\theta_\gamma}^2\\
\iff v^*&\in\arg\min_{v} \sum_{y,s^\prime}d^\theta_\gamma(y,s^\prime|x,s)\cdot\\
&\!\!\!\!\!\!\!\!\!\!\!\!\!\!\!\!\!\!\!\!\!\!\!\left(\sum_{ a^\prime\in\A}\mu( a^\prime|y,s^\prime;\theta)\left( \bar{C}(y,s^\prime,a^\prime)+\gamma \sum_{z,s^{\prime\prime}}\bar{P}(z,s^{\prime\prime}|y,s^\prime,a^\prime) \phi^\top\left(z,s^{\prime\prime}\right)v ds^{\prime\prime}\right)-\phi^\top(y,s^\prime)v\right)^2.
\end{split}
\]
By the projection theorem on Hilbert space, the orthogonality condition for $v^*$ becomes:
\[\small
\begin{split}
&\sum_{y,a^\prime,s^\prime}\pi^\theta_\gamma(y,s^\prime,a^\prime|x,s)\phi(y,s^\prime)(v^*)^\top\phi(y,s^\prime)\\
=&\sum_{y,a^\prime,s^\prime}\bigg\{\pi^\theta_\gamma(y,s^\prime,a^\prime|x,s)\phi(y,s^\prime)\bar{C}(y,s^\prime,a^\prime)+\gamma\sum_{z,s^{\prime\prime}}\pi^\theta_\gamma(y,s^\prime,a^\prime|x,s) \bar{P}(z,s^{\prime\prime}|y,s^\prime,a^\prime) \phi(y,s^\prime)\phi^\top\left(z,s^{\prime\prime}\right) \bigg\}v^*.
\end{split}
\]
This condition can be written as $Av^*=b$
where
\begin{equation}\label{eq:A}
A=\sum_{y,a^\prime,s^\prime}\pi^\theta_\gamma(y,s^\prime,a^\prime|x,s)\phi(y,s^\prime)\left(\phi^\top(y,s^\prime)- \gamma\sum_{z,s^{\prime\prime}}\bar{P}(z,s^{\prime\prime}|y,s^\prime,a) \phi^\top\left(z,s^{\prime\prime}\right)ds^{\prime\prime}\right)
\end{equation}
is a finite dimensional matrix in $\reals^{\kappa_2\times\kappa_2}$ and 
\begin{equation}\label{eq:b}
b=\sum_{y,a^\prime,s^\prime}\pi^\theta_\gamma(y,s^\prime,a^\prime|x,s)\phi(y,s^\prime)\bar{C}(y,s^\prime,a^\prime).
\end{equation}
is a finite dimensional vector in $\reals^{\kappa_2}$. The matrix $A$ is invertible since Lemma \ref{lem:unique_projected_fixed_pt} guarantees that \eqref{eq:unique_fixed_pt} has a unique solution $v^*$. 
Note that the projected equation $Av=b$ can be re-written as
\[
v=v-\xi(Av-b)
\]
for any positive scaler $\xi\geq 0$. Specifically, since
\[\small
Av-b=\sum_{y,a^\prime,s^\prime}\pi^\theta_\gamma(y,s^\prime,a^\prime|x,s)\phi(y,s^\prime)\left(v^\top\phi(y,s^\prime)- \sum_{z,s^{\prime\prime}}  \bar{P}(z,s^{\prime\prime}|y,s^\prime,a^\prime) (\gamma v^\top\phi\left(z,s^{\prime\prime}\right)+\bar{C}(y,s^\prime,a^\prime))\right),
\] 
one obtains
\[
Av-b=\mathbb E^{\pi^\theta_\gamma}\left[\phi(x_k,s_k)\left(v^\top\phi(x_k,s_k)- \gamma v^\top\phi\left(x_{k+1},s_{k+1}\right)-\bar{C}(x_k,s_k,a_k)\right)\right]
\]
where the occupation measure $\pi^\theta_\gamma(x,s,a|x^0,\nu)$ is a valid probability measure. Recall from the definitions of $(A,b)$ that
\[
\begin{split}
A=&\mathbb E^{\pi^\theta_\gamma}\left[\phi(x_k,s_k)\left(\phi^\top(x_k,s_k)- \gamma\phi^\top\left(x_{k+1},s_{k+1}\right)\right)\right],\\
b=&\mathbb E^{\pi^{\theta}_\gamma}\left[\phi(x_k,s_k)\bar{C}(x_k,s_k,a_k)\right]
\end{split}
\]
where $\mathbb E^{\pi^{\theta}_\gamma}$ is the expectation induced by the occupation measure (which is a valid probability measure).
%%%%%%%%%%%%%%%%%%%%%%%%%%%%%%%%%%%%%%%%%%%%%%%%%%%%%%%%%%%
%%%%%%%%%%%%%%%%%%%%%%%%%%%%%%%%%%%%%%%%%%%%%%%%%%%%%%%%%%%
%%%%%%%%%%%%%%%%%%%%%%%%%%%%%%%%%%%%%%%%%%%%%%%%%%%%%%%%%%%
\newpage
\section{Supplementary: Gradient with Respect to $\theta$}
By taking gradient of $\V^\theta$ with respect to $\theta$, one obtains
\[\small
\begin{split}
&\nabla_\theta\V^\theta(x^0,\nu)=\sum_{a}\nabla_{\theta}\mu(a|x^0,\nu;\theta) Q^\theta(x^0,\nu,a)+\mu(a|x^0,\nu;\theta)\nabla_{\theta} Q^\theta(x^0,\nu,a)\\
=&\sum_{a}\nabla_{\theta}\mu(a|x^0,\nu;\theta) Q^\theta(x^0,\nu,a)+\mu(a|x^0,\nu;\theta)\nabla_{\theta}\left[\bar{C}(x^0,\nu,a)+\sum_{x^\prime,s^\prime}\gamma\bar{P}(x^\prime,s^\prime|x^0,\nu,a)\V^\theta\left(x^\prime,s^\prime\right)\right]\\
=&\sum_{a}\nabla_{\theta}\mu(a|x^0,\nu;\theta) Q^\theta(x^0,\nu,a)+\gamma\mu(a|x^0,\nu;\theta)\left[\sum_{x^1,s^1}\gamma\bar{P}(x^1,s^1|x^0,\nu,a)\nabla_{\theta}\V^\theta\left(x^1,s^1\right)\right]\\
=&h^\theta(x^0,\nu)+\gamma\sum_{x^1,s^1,a^0}\mu(a^0|x^0,\nu;\theta)\bar{P}(x^1,s^1|x^0,\nu,a^0)\nabla_{\theta}\V^\theta\left(x^1,s^1\right)
\end{split}
\]
where 
\[
h^\theta(x^0,\nu)=\sum_{a}\nabla_{\theta}\mu(a|x^0,\nu;\theta) Q^\theta(x^0,\nu,a).
\]
Since the above expression is a recursion, one further obtains 
\[\small
\begin{split}
\nabla_{\theta}\V^\theta(x^0,\nu)=&h^\theta(x^0,\nu)+\gamma\sum_{a,x^1,s^1}\mu(a|x^0,\nu;\theta)\bar{P}(x^1,s^1|x^0,\nu,a)\\
&\left[h^\theta(x^1,s^1)+\gamma\sum_{a^1,x^2,s^2}\mu(a^1|x^1,s^1;\theta)\bar{P}(x^2,s^2|x^1,s^1,a^1) \nabla_{\theta}\V^\theta\left(x^2,s^2\right)\right].
\end{split}
\]
By the definition of occupation measures, the above expression becomes
\begin{equation}\label{eq:theta_gradient}
\begin{split}
\nabla_\theta\V^\theta(x^0,\nu)=&\sum_{k=0}^\infty\gamma^k\!\!\sum_{x^\prime,a^\prime,s^\prime} \mu(a^\prime|x^\prime,s^\prime;\theta)\bar{P}(x_k=x^\prime,s_k=s^\prime|x_0=x^0,s_0=\nu)h^\theta(x^\prime,s^\prime)\\
=&\frac{1}{1-\gamma}\sum_{x^\prime,s^\prime} d^\theta_\gamma(x^\prime,s^\prime|x_0=x^0,s_0=\nu)h^\theta(x^\prime,s^\prime)\\
=&\frac{1}{1-\gamma}\sum_{x^\prime,s^\prime} d^\theta_\gamma(x^\prime,s^\prime|x_0=x^0,s_0=\nu)\sum_{a^\prime\in \A}\nabla_{\theta}\mu(a^\prime|x^\prime,s^\prime;\theta) Q^\theta(x^\prime,s^\prime,a^\prime)\\
=&\frac{1}{1-\gamma}\sum_{x^\prime,a^\prime,s^\prime} \pi^\theta_\gamma(x^\prime,s^\prime,a^\prime|x_0=x^0,s_0=\nu)\nabla_{\theta}\log\mu(a^\prime|x^\prime,s^\prime;\theta) Q^\theta(x^\prime,s^\prime,a^\prime)\\
=&\frac{1}{1-\gamma}\sum_{x^\prime,a^\prime,s^\prime}\pi^\theta_\gamma(x^\prime,s^\prime,a^\prime|x_0=x^0,s_0=\nu)\nabla_{\theta}\log\mu(a^\prime|x^\prime,s^\prime;\theta)A^\theta(x^\prime,s^\prime,a^\prime)
\end{split}
\end{equation}
where 
\[
A^\theta(x,s,a)= Q^\theta(x,s,a)-\V^\theta(x,s)
\]
is the advantage function. The last equality is due to the fact that 
\[
\begin{split}
\sum_{a}\mu(a|x,s;\theta)\nabla_\theta \log \mu(x|s,a;\theta)\V^\theta(x,s)=&\V^\theta(x,s)\cdot\sum_{a}\nabla_\theta\mu(a|x,s;\theta)\\
=&\V^\theta(x,s)\cdot\nabla_\theta\sum_{a}\mu(a|x,s;\theta)=\nabla_\theta(1)\cdot\V^\theta(x,s)=0.
\end{split}
\]
Thus, the gradient of the Lagrangian function is 
\[
\nabla_\theta L(\theta,\nu,\lambda)=\nabla_\theta\V^\theta(x,s)\bigg\vert_{x=x^0,s=\nu}.
\]

\end{document}